\documentclass[10pt]{article}

\usepackage{float}
\usepackage{drexreport}
\usepackage[utf8]{inputenc}
\usepackage{amsmath,amssymb,amsthm,mathtools}
\usepackage{bm}
\usepackage{dsfont}
\usepackage{booktabs}
\usepackage{multirow}
\usepackage{tabularx}
\usepackage{graphicx}
\graphicspath{{figures/}}
\usepackage{tikz}
\usetikzlibrary{arrows.meta,calc,fit,positioning,backgrounds,decorations.pathreplacing,shapes.geometric,shadows.blur}
\usepackage{subcaption}
\usepackage{enumitem}
\usepackage{microtype}
\usepackage{algorithm}
\usepackage{algpseudocode}
\usepackage{nicefrac}
\usepackage{xspace}
\usepackage{threeparttable}
\usepackage[numbers,sort&compress]{natbib}
\usepackage[colorlinks=true, linkcolor=drexaccent, citecolor=drexaccent, urlcolor=drexaccent]{hyperref}
\definecolor{linkblue}{HTML}{1A56DB}
\newcommand{\bluehref}[2]{\begingroup\hypersetup{urlcolor=drexaccent}\href{#1}{#2}\endgroup}
\usepackage[capitalise,noabbrev]{cleveref}

\definecolor{ink}{HTML}{1B1640}
\definecolor{inkmute}{HTML}{5B5880}
\definecolor{cardline}{HTML}{DCD7F3}
\definecolor{stok}{HTML}{FFFFFF}
\definecolor{sline}{HTML}{C9C2F2}
\definecolor{qtok}{HTML}{EEEBFD}
\definecolor{qline}{HTML}{B9AEF7}
\definecolor{otok}{HTML}{D3CBFA}
\definecolor{etok}{HTML}{6B5CE7}
\definecolor{dtok}{HTML}{F59E0B}
\definecolor{itok}{HTML}{14B8A6}
\definecolor{iline}{HTML}{0F9F8F}
\definecolor{ifill}{HTML}{E6F7F4}
\definecolor{track}{HTML}{ECEAF6}
\definecolor{chipfill}{HTML}{EEEBFD}
\definecolor{arrowc}{HTML}{6B5CE7}
\definecolor{encback}{HTML}{B4ACE8}
\definecolor{encA}{HTML}{241B6B}
\definecolor{encB}{HTML}{4F43D6}
\definecolor{headfill}{HTML}{FAF9FF}
\definecolor{barA}{HTML}{4338CA}
\definecolor{barB}{HTML}{A89CF8}
\colorlet{segS}{encback}
\colorlet{segA}{etok}
\colorlet{segB}{dtok}
\colorlet{tokI}{qtok}
\colorlet{tokO}{otok}
\colorlet{tokE}{etok}
\colorlet{tokD}{dtok}
\colorlet{bbfill}{qtok}
\colorlet{phfill}{headfill}

\newcommand{\CalVtwoN}{118{,}305}
\newcommand{\CalVtwoECE}{1.05}
\newcommand{\CalVtwoECELo}{0.90}
\newcommand{\CalVtwoECEHi}{1.27}
\newcommand{\CalVtwoT}{0.91}

\newcommand{\DILuna}{47.78}
\newcommand{\DIMaxVtwoLo}{3.3}
\newcommand{\DIMaxVtwoHi}{7.8}
\newcommand{\DIMaxWins}{28}
\newcommand{\DIBenchN}{37}
\newcommand{\CalMaxAcc}{87.9}
\newcommand{\CalMaxConf}{91.2}
\newcommand{\CalMaxECE}{3.61}
\newcommand{\CalMaxECET}{1.35}
\newcommand{\CalMaxT}{1.35}
\newcommand{\CalMaxSelHalf}{99.3}
\newcommand{\CalMaxSelEighty}{96.1}
\newcommand{\CalTwoAcc}{86.0}
\newcommand{\CalTwoECE}{1.37}
\newcommand{\CalTwoECET}{0.79}
\newcommand{\CalTwoT}{0.91}
\newcommand{\CalTwoSelHalf}{98.9}
\newcommand{\CalTwoSelEighty}{94.4}
\newcommand{\CalOneECE}{4.54}
\newcommand{\CalOneECET}{1.17}
\newcommand{\CalOneT}{1.48}
\newcommand{\CalLunaECE}{4.87}
\newcommand{\CalLunaECET}{6.57}
\newcommand{\CalLunaT}{1.91}
\newcommand{\PairedMaxVtwo}{1.86}
\newcommand{\PairedMaxVtwoLo}{1.72}
\newcommand{\PairedMaxVtwoHi}{2.00}
\newcommand{\OrdN}{1{,}591}
\newcommand{\OrdReq}{1{,}040}
\newcommand{\OrdMaxFlip}{4.3}
\newcommand{\OrdMaxFlipTwo}{1.0}
\newcommand{\OrdMaxFlipSix}{13.3}
\newcommand{\OrdMaxAccO}{83.6}
\newcommand{\OrdMaxAccP}{83.4}
\newcommand{\OrdVtwoFlip}{5.9}
\newcommand{\OrdVoneFlip}{5.3}
\newcommand{\OrdJevFlip}{4.3}
\newcommand{\OrdNTwo}{915}
\newcommand{\OrdNMid}{503}
\newcommand{\OrdNSix}{173}
\newcommand{\DBMaxPix}{94.0}
\newcommand{\DBMaxPixImg}{98.3}
\newcommand{\DBMaxTxt}{92.1}
\newcommand{\DBMaxTxtImg}{63.3}
\newcommand{\DBMaxPixK}{1{,}007}

\newcommand{\drex}{Drex\xspace}
\newcommand{\kev}{Kev\xspace}
\newcommand{\sysone}{System~One\xspace}
\newcommand{\tok}[1]{\texttt{<#1>}}
\newcommand{\R}{\mathbb{R}}
\newcommand{\E}{\mathbb{E}}
\newcommand{\seg}{\operatorname{seg}}
\newcommand{\softmax}{\operatorname{softmax}}
\newcommand{\KL}{\operatorname{KL}}
\newcommand{\sKL}{\operatorname{\overline{KL}}}

\newcommand{\ECE}{\operatorname{ECE}}
\newcommand{\TV}{\operatorname{TV}}

\newtheorem{definition}{Definition}
\theoremstyle{remark}
\newtheorem{remark}{Remark}
\crefname{equation}{Eq.}{Eqs.}
\Crefname{equation}{Equation}{Equations}
\crefname{proposition}{Proposition}{Propositions}
\crefname{remark}{Remark}{Remarks}

\title{Towards Foundation Decision Models: Drex}
\author{Nischay Dhankhar \quad Alisher Amantay \quad Zhanibek Datbayev \\
      Darkhan Akhmetov \quad  Yash Shah \quad Nurmukhamed Ubaidullayev \\
        Raman Saparkhan \quad Dos Baha \quad Abulhair Saparov}
\affil{Nace.AI Research Team}
\links{\href{https://nace.ai/drex}{nace.ai/drex} \enspace$\cdot$\enspace \href{https://github.com/nace-ai/drex-decision-models}{github.com/nace-ai/drex-decision-models} \enspace$\cdot$\enspace \href{https://huggingface.co/nace-ai}{huggingface.co/nace-ai}}
\date{October 2026}

\begin{document}
\maketitle

\begin{abstract}
Autoregressive approaches with large language models such as long chain-of-thought reasoning have become the default for new AI applications because they reach higher accuracy on a wide range of tasks.
In production, however, such approaches give inconsistent answers, require parsing of outputs, and add cost and latency.
This has driven a resurgence of discriminative inference in the form of decision models, which, given a question and a set of options supplied by the user, return a probability distribution over the options without autoregressive decoding. Decision models match frontier LLMs on well-defined tasks at a fraction of the cost.
We present \drex, a family of four open decision models; the largest, \drex Max, is competitive with frontier models on real-data decisions while remaining fast, calibrated, and stable under reordering of the options.
Each request consists of a \emph{state} (a document, record, or agent context, optionally with images) and questions related to that state, with each asking for a choice among a set of options, a yes/no judgment, or a score on an ordered scale. The four models have 8B (DLM), 9B (v1.5), 35B with 3B active (v2.0), and 27B (Max) parameters.
Training combines supervised fine-tuning with reinforcement learning from adjusted feedback (RLAF), plus vision tuning for Max.
Controlled ablations show that RLAF improves accuracy and consistency.
On Decision Index 0.4.1, Max sets a new state-of-the-art among the decision models we compare (67.52 on the public suite vs 60.70 for Jev 1.14.0).
On the independently run Artificial Analysis Decision Index, it has the second-highest score among decision models (64 vs.\ 65 for the three leading models).
On Decision Bench v4, it beats Jev 1.14 and our other models, is statistically tied with d3-turbo and GPT-6 Sol, and trails Claude Opus 5.5 by 1.6 points; it has the highest accuracy among decision models on DecideBench and is competitive with the top model on JevBench-231.
Max answers a typical request in 183\,ms, against about 2.0\,s for GPT-6 Sol and 3.5\,s for Claude Opus 5.5, and demonstrates output stability under option reordering.
All four models are open-weight\footnote{Max, v2.0, and v1.5 are released under the Nace.AI Open RAIL-M license (Max: modified, version~1.0, October 2026); DLM under the CC BY-NC 4.0 license of its backbone; repository code under Apache-2.0. Max, v2.0, and v1.5 are also served through a hosted API and OpenRouter. Our enterprise models, \drex Max v1.1 and \drex for Finance, are available only through the hosted API at \bluehref{https://console.nace.ai}{\texttt{console.nace.ai}}.}; the weights and inference code are collected at \bluehref{https://github.com/nace-ai/drex-decision-models}{\texttt{github.com/nace-ai/drex-decision-models}}.
\end{abstract}

\begin{center}
\includegraphics[width=\textwidth]{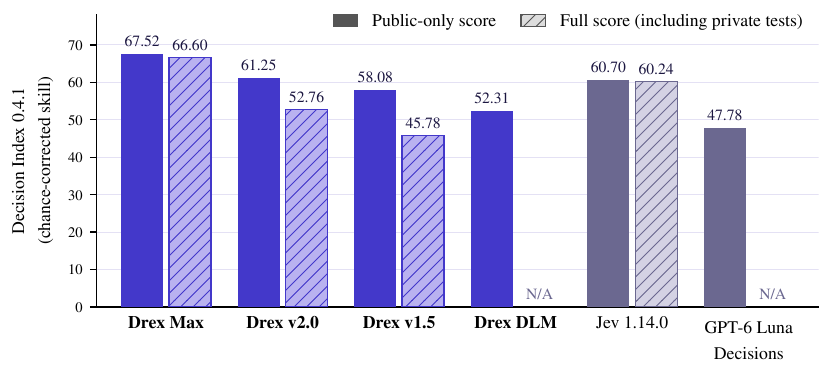}\\[3pt]
\parbox{\textwidth}{\small\textbf{Decision Index 0.4.1 for select decision models.} Public-only score and full score (which includes private tests), both chance-corrected, from the 0.4.1 leaderboard of 22 October 2026 (\cref{tab:di032}); \drex models in blue. \drex DLM and GPT-6 Luna Decisions have no full score.}
\end{center}

\section{Introduction}
\label{sec:intro}

Language models learn their representations through a single objective:
predicting the next token in a sequence, or, in the case of masked language models,
predicting tokens withheld from their context~\citep{devlin2019bert,brown2020gpt3}.
These representations, however, transfer to many tasks in which no text
is generated.
Such tasks can often be categorized into one of three kinds:
\emph{Classification} maps an input to one label from a fixed set~\citep{devlin2019bert,raffel2020t5}.
\emph{Scoring} maps an input to a number, such as the relevance of a passage to a query~\citep{nogueira2019passage,ma2023finetuning} or the similarity of two sentences~\citep{reimers2019sbert}.
\emph{Embedding} maps an input to a vector that is compared, clustered or searched~\citep{reimers2019sbert,wang2024improving,behnamghader2024llm2vec}.
When utilizing autoregressive models to solve these kinds of tasks, the same design question always arises: How to read the answer from a model whose native output is a distribution over each token? To address this, three solutions have been proposed.
Encoder models are fine-tuned with a small head at the last layer---a linear layer to produce a single score or additionally a softmax over a fixed set of output labels~\citep{devlin2019bert,nogueira2019passage}.
T5 recast these tasks as text-to-text, with the label generated as a string~\citep{raffel2020t5}, and for decoder-only models, prompting for a generated label became a common interface~\citep{brown2020gpt3}.
Recent work fine-tunes decoder-only models to return scores and embeddings directly~\citep{ma2023finetuning,wang2024improving,behnamghader2024llm2vec}.

In all three of the above solutions, the shape of the output is fixed before the model is deployed.
A classification head has its labels fixed at training time, a reranker returns one number per candidate, an embedding model returns one vector per input.
In contrast, we are interested in the setting where the set of candidate outputs changes across inputs: the user supplies a context, a question and the set of candidate outputs, and the model produces one probability per option.
A fixed head does not suffice in this setting.
One route is to prompt an autoregressive model for the answer and parse the output text; another is to generate the option and constrain the generation to the allowed strings, a technique with a long history in sequence generation~\citep{anderson2017guided,hokamp2017grid,shin2021constrained,willard2023efficient}.
The second option guarantees a valid output, and the probability of an option is computed as the product of per-token probabilities, but such a model is not trained to predict a distribution over a set of options.
Additionally, computing the output probabilities of constrained generation involves renormalizing the per-token probabilities at every step, and the result can differ from the distribution that the model would assign to the valid options as whole strings, for example when options share prefixes or differ in length~\citep{park2024grammar}.
Scoring every option separately and renormalizing across options recovers a distribution over the valid options, but at the cost of one forward pass per option.
Specialized models that return a probability rather than generated text have been built for single, domain-specific decisions, such as predicting whether a sales conversation will convert~\citep{nandakishor2025salesrl}, but their output is fixed to that one decision.
In contrast, our approach is to develop a new kind of readout: one that performs classification in which the set of options is given at inference time, by the user.
We use the term \emph{decision model} to refer to a model that returns the probabilities directly, without autoregressive decoding, following the \sysone interface introduced by Jev~\citep{typesafe2026jev} and reconstructed in the open \kev stack~\citep{kev2026,nace2026drexrepo} (\cref{sec:problem}).
\emph{Pointer models} (e.g., \drex DLM, v1.5, v2.0) are decision models where the context is encoded once and then shared across multiple questions, reducing cost and latency in appropriate applications.
\drex Max re-encodes the context for each question using a dense multimodal backbone with non-causal full attention, and reads the answer from a \emph{readout head}: each option is labeled with a single-token code (\texttt{A}, \texttt{B}, \dots), and a bias-free linear layer maps the final hidden state to one logit per code, up to 255 codes (\cref{sec:readout}).
This results in higher accuracy in exchange for increased cost per question.
In developing these models, we aim to resolve the question: Can a single model serve \emph{arbitrary} option sets with \emph{calibrated} probabilities, answer \emph{many} questions over one long context, maintain \emph{stability} when the options are reordered, and do so at production latency?
In this report, we investigate the above question by considering the design choices in the development of decision models and their effect on the model's behavior and performance. We organize this investigation into four research questions:
\begin{itemize}[leftmargin=2.6em, itemsep=1pt, topsep=2pt]
  \item[\textbf{RQ1}] \emph{Calibration.} How well do the probabilities returned by decision models match their empirical accuracy, overall and across application areas and numbers of options? Does a single global temperature, fit on held-out decisions, improve this calibration?
  \item[\textbf{RQ2}] \emph{Option order.} How sensitive are the probabilities and decisions of pointer and readout-head models to the order in which the options are presented?
  \item[\textbf{RQ3}] \emph{Training.} How much does each stage of the training pipeline contribute to the final model: supervised fine-tuning (SFT) with consistency losses, reinforcement learning from verifiable rewards adjusted by the output probabilities of an LRM-as-a-judge under a KL penalty relative to the SFT model (RLAF), and vision SFT? Does the vision stage preserve accuracy on text inputs?
  \item[\textbf{RQ4}] \emph{Capacity, attention, and modality.} What are the advantages and disadvantages of using a dense multimodal backbone and bidirectional attention versus a shared-state pointer head?
\end{itemize}
We study these questions by training and studying four models that share one interface (\cref{sec:problem}) and a common training-data base (\cref{sec:universe}), which we extend for training \drex Max.
Most of the evidence is comparative, across models, training stages, and inference-time interventions.
Therefore, outside one ablation of \drex Max's training (\cref{sec:maxablate}), we report associations and trade-offs rather than causal relationships; \cref{sec:limits} lists the comparisons we did not run, and the hypotheses that require further experimentation, such as which specific changes in the design and training of Max drive its gains.
The pointer models are built on top of the open \kev code base.

We evaluate the following properties, which are desirable for decision models: accuracy, as compared against frontier and other decision-specialized models via paired statistical comparisons; calibration, measured on hundreds of thousands of decisions that span diverse applications; robustness to the order of options; and accuracy on image inputs, where we compare the model's performance when given image inputs versus equivalent text inputs.
We additionally measure the contribution of each training stage to these properties.
We quantify robustness to the order of options with the \emph{flip rate}, which is the fraction of decisions for which the option with the highest probability changes when the order of the options changes.
We quantify the usefulness of a model's confidence with \emph{precision@top-$k$\%}, which is the accuracy of the model on the $k$\% of decisions on which it is most confident.

\paragraph{Contributions.}
\begin{itemize}[leftmargin=1.3em, itemsep=1pt, topsep=1pt]
  \item \textbf{Calibration at scale (RQ1).} We measure the calibration of \drex Max, v2.0, v1.5, and our enterprise models on about 230k decisions each. We report bootstrap confidence intervals; breakdowns by domain and by the number of options; precision@top-$k$\%; and the effect of task-specific tuning of the temperature parameter (\cref{sec:calib-results}).
  \item \textbf{Order sensitivity and confidence (RQ2).} We study how the order of the options affects four production models, accessed through their public APIs, on \OrdN{} decisions, each presented in three orders: the original order, the reversed order, and a seeded random permutation. We find that the top option changes with the order mostly on decisions that have many options and low model confidence. On these workloads, declining to answer the decisions on which the top option changes with the order is no more accurate than declining to answer the decisions with the lowest confidence. Averaging the three probability distributions predicted by the model for the respective three orders improves log loss but not accuracy (\cref{sec:order}).
  \item \textbf{A training recipe for highly capable decision models (RQ3).} We provide a three-stage recipe, which we use to train \drex Max (\cref{sec:train}). The first stage is supervised fine-tuning (SFT). The second stage is reinforcement learning from adjusted feedback (RLAF). The third stage is SFT of the image encoder and of its projection into the language model, which teaches the model to make decisions from images. During the first stage, we add two consistency terms to the SFT loss since the architecture of \drex Max does not make its answers independent of the order of the options. The first term penalizes a change in the predicted distribution when the options are permuted; the second term penalizes a change when the sentences or fields of the state are shuffled (\cref{sec:objective}). During RLAF, the model is optimized to maximize a reward that combines the verified correct answer with the probabilities that a judge model assigns to the options. The objective is regularized by the KL divergence between the output distribution of the current model and that of its own SFT checkpoint. \Cref{sec:sdrl} derives the gradient of this objective and the closed-form output distribution that maximizes it. An ablation of the training of \drex Max on its own backbone and training data (\cref{tab:maxablate}) shows that RLAF improves the Decision Index score by 2.0 points, that the addition of the consistency terms in the loss function lowers the flip rate from 6.3\% to 4.2\%, and that the third stage does not reduce accuracy on text inputs. A seven-checkpoint analysis of the training of \drex v2.0 shows the relative contribution of each stage in the training recipe of the pointer models (\cref{sec:dynamics}).
  \item \textbf{\drex Max (RQ4).} We present \drex Max, a dense (i.e., not mixture-of-experts) multimodal decision model trained from Qwen3.8-27B. The model removes the causal attention mask in its full-attention layers, and it produces its answer with a readout head, i.e., a bias-free linear layer that maps the final hidden state to a vector of 255 logits, one per option code (\cref{sec:max}). We describe the cost of running the model, which is higher per question than that of the pointer models since the model re-encodes the context for every question (\cref{sec:maxcost}). We also measure how the accuracy of the model on decisions that involve images changes when it receives the image itself rather than a text description and transcription of the image (\cref{sec:dbench}).
  \item \textbf{One interface on four backbones.} We develop a single interface that seamlessly functions with four distinct backbone models. The pointer head is added to the last layer of each of the following backbone models: (1) a Gated DeltaNet hybrid model, (2) a diffusion language model with bidirectional attention over the context, and (3) a sparse mixture-of-experts model. The readout head is added to the last layer of a dense multimodal model with 27 billion parameters. All four models, one per backbone, accept the same requests and return the same output format (\cref{sec:problem}), and we compare them in \cref{sec:family}.
  \item \textbf{Paired evaluation.} We evaluate the models on Decision Index (versions 0.2.1 and 0.4.1, and version 0.4.1 with its private tests), Decision Bench v4 (with images given as input in image encoder and as text), JevBench, DecideBench, a game arena, a computer-use router, and a vision suite, and we compare them with frontier chat models on the same samples. We report per-benchmark tables, Wilson intervals for accuracies, paired bootstrap intervals for differences between models, and exact McNemar tests throughout (\cref{sec:results}).
  \item \textbf{Deployment and inference optimization.} We describe how our decision models are deployed and the techniques we leverage to improve inference speed (\cref{sec:serving}).
\end{itemize}

\paragraph{Release status.}
All four models are released as open-weight models; the weights and inference code are collected at \bluehref{https://github.com/nace-ai/drex-decision-models}{\texttt{github.com/nace-ai/drex-decision-models}}.
\drex Max, v2.0, and v1.5 are released under the Nace.AI Open RAIL-M license~\citep{rv_naceopenrailm}, with safetensors, GGUF builds, and llama.cpp and Ollama runtimes; \drex DLM is released under the non-commercial license of its backbone, CC BY-NC 4.0~\citep{rv_ccbync40}.
\drex Max, v2.0, and v1.5 are also served through the hosted \sysone API (as \texttt{drex-max}, \texttt{drex-v2.0}, and \texttt{drex-v1.5}) and through OpenRouter~\citep{openrouter2026nace}.

\section{Related work}
\label{sec:related}

Decision models draw on several lines of work: likelihood scoring and multiple-choice prompting with a fixed set of candidate answers~\citep{brown2020gpt3,robinson2023mcqa}, cross-encoders and pointer networks~\citep{nogueira2019passage,vinyals2015pointer}, the option-order sensitivity of language models~\citep{zheng2024mcq,pezeshkpour2024order}, and calibration and proper scoring rules~\citep{guo2017calibration,gneiting2007proper}.
The four \drex models are each built on different language-model architectures: a diffusion language model~\citep{fu2025efficientdlm}, Gated DeltaNet hybrid model~\citep{yang2025gateddeltanet,mimo2026qwen9b,xiaomi2025mimo,qwen38,yang2025qwen3}, and a sparse mixture of experts language model (Qwen3.6-35B-A3B)~\citep{fedus2022switch,rv_jacobs1991moe,shazeer2017moe,qwen36moe}. In \emph{hybrid models}, such as Qwen3.6-35B-A3B, some layers utilize global attention while others utilize linear attention. The training methodology of \drex Max includes an RLAF training stage, which is an instance of KL-regularized learning from feedback~\citep{ouyang2022instructgpt,rafailov2023dpo}. The main benchmarks are Decision Index~\citep{decisionindex} and Decision Bench~\citep{decisionbench2026}.
\Cref{app:related} discusses each line of work in detail.

\paragraph{Relation to Jev and Kev.}
The interface follows Jev's \sysone API; the pointer models are built on the open \kev code base~\citep{kev2026}, and \drex Max's attention pattern and readout adopt the open \texttt{pplx-decider} design~\citep{perplexity2026decider}, but we trained its weights ourselves, starting from the Qwen3.8-27B base model. \drex inherits a number of features from \kev: the packed encoding with a block-causal mask (\cref{sec:arch}), the encoding of each question as a separate causal sequence that is conditioned on a cached state,
the pointer head over tokens that delimit each option, and the loss functions for training (cross-entropy, squared Cram\'er distance, permutation consistency). To the best of our knowledge, everything else is novel, especially in their application to decision models: the additional training stage from adjusted feedback (\cref{sec:rlaf}), the training of the readout head of \drex Max (\cref{sec:train}), the training of its image encoder and the projection of that encoder into the language model (\cref{sec:maxvision}), the evaluation (\cref{sec:results}), and the inference stack (\cref{sec:serving}).

\section{Problem formulation}
\label{sec:problem}
\begin{definition}[Decision request]
A request is a tuple $(S, \mathcal{I}, \{q_k\}_{k=1}^{Q})$ where $S$ is the \emph{state} (text, or JSON rendered as indented text), $\mathcal{I}$ is an optional, possibly empty set of images, and each question $q_k = (\tau_k, u_k, O_k)$ has a type $\tau_k \in \{\texttt{choice},\texttt{noul},\texttt{score}\}$, an instruction $u_k$, and an ordered option list $O_k = (o_{k,1},\dots,o_{k,K_k})$ with $K_k \le 255$.
A decision model returns, for every $k$, a distribution $p_k \in \Delta^{K_k-1}$.
\end{definition}

The three question types all return a probability distribution over a finite list of options; they differ in how the user specifies the options and in how the answer is reported.
A \texttt{choice} question lists named options, with $K \le 255$, and the model returns a probability for each option together with the most probable option.
A \texttt{noul} question is a yes/no question: it is a question with two options, $O=(\text{no},\text{yes})$, for which the user may supply definitions of what ``yes'' and ``no'' mean, and the output is $p(\text{yes})$. We provide it as a separate type so that the user does not have to write out the two options of a binary decision.
A \texttt{score} question is a request for a rating on an integer scale: the options are the integer levels $0,\dots,L-1$, and the model returns the expected level $\hat s = \sum_{i} i\,p_i$ in addition to the probability of each level. We provide this question type since it occurs frequently in many real-world settings, where a single number representing the expected value of the output distribution is desirable,
rather than a probability for each level.
\Cref{tab:types} in the Appendix summarizes what the user supplies and what the model returns for each type.
We define confidence as in the \sysone interface introduced by Jev (\cref{sec:related}), so that our models follow the interface that existing deployments already use and report confidence values with the same meaning~\citep{nace2026drexrepo}:
\begin{equation}
  c_{\texttt{choice}} = \frac{\max_j p_j - 1/K}{1 - 1/K},
  \qquad
  c_{\texttt{score}} = 1 - \frac{\E_{i\sim p}\,|i - \hat\imath|}{L - 1},\quad \hat\imath = \arg\max_i p_i .
  \label{eq:confidence}
\end{equation}

\paragraph{Desirable properties.}
The following are desirable properties for decision models:
\begin{itemize}[leftmargin=1.3em, itemsep=1pt, topsep=2pt]
  \item \emph{Closed-world output:} The distribution $p_k$ is supported on $O_k$, because the output of the model is defined only over the supplied options. It is not supported on $O_k$ merely because the input instructs the model to choose among the options (i.e., prompting) or because generated text is restricted to the options (i.e., constrained decoding).
  \item \emph{Question independence:} The distribution $p_k$ does not depend on which other questions are asked in the same request. Therefore, a question receives the same probabilities, up to floating-point error, whether it is queried alone, processed together with other questions in one batch, or replayed later from a cached request.
  \item \emph{Order robustness:} The predicted probability of an option is invariant with respect to the ordering of the options in the input. That is, for any permutation $\pi$ of $O_k$, $p_k(\cdot \mid \pi O_k) = \pi\, p_k(\cdot \mid O_k)$. A weaker version of this condition is that the top option is unchanged under every permutation.
  \item \emph{Calibration:} $\Pr[\arg\max_j p_{k,j} \text{ is correct} \mid \max_j p_{k,j} = c] = c$. This property is desirable, for example, because a threshold on the confidence can then serve as a rule for routing decisions.
  \item \emph{Grounding in every modality supplied:} When the evidence is an image, the model must process the image and incorporate the information therein when forming its prediction, instead of relying on a text description of the image.
\end{itemize}
The architectures of all \drex models guarantee the first two properties by construction (\cref{app:archdetails}). The last three properties are learned, so we measure how well each of these properties holds rather than assuming that they hold.

We measure the expected calibration error over 15 equal-width bins (ECE). We also measure the log loss (also, negative log-likelihood, abbreviated NLL), the Brier score, and precision@top-$k$\%, which is the accuracy on the fraction of decisions with the highest confidence (\cref{app:metrics} in the Appendix).
We measure order robustness with the top-1 flip rate, which is the fraction of decisions whose top option changes under a permutation of the options (\cref{app:metrics} in the Appendix).

\section{Pointer architecture}
\label{sec:arch}

\begin{figure}[t]
\centering
\resizebox{\textwidth}{!}{\input{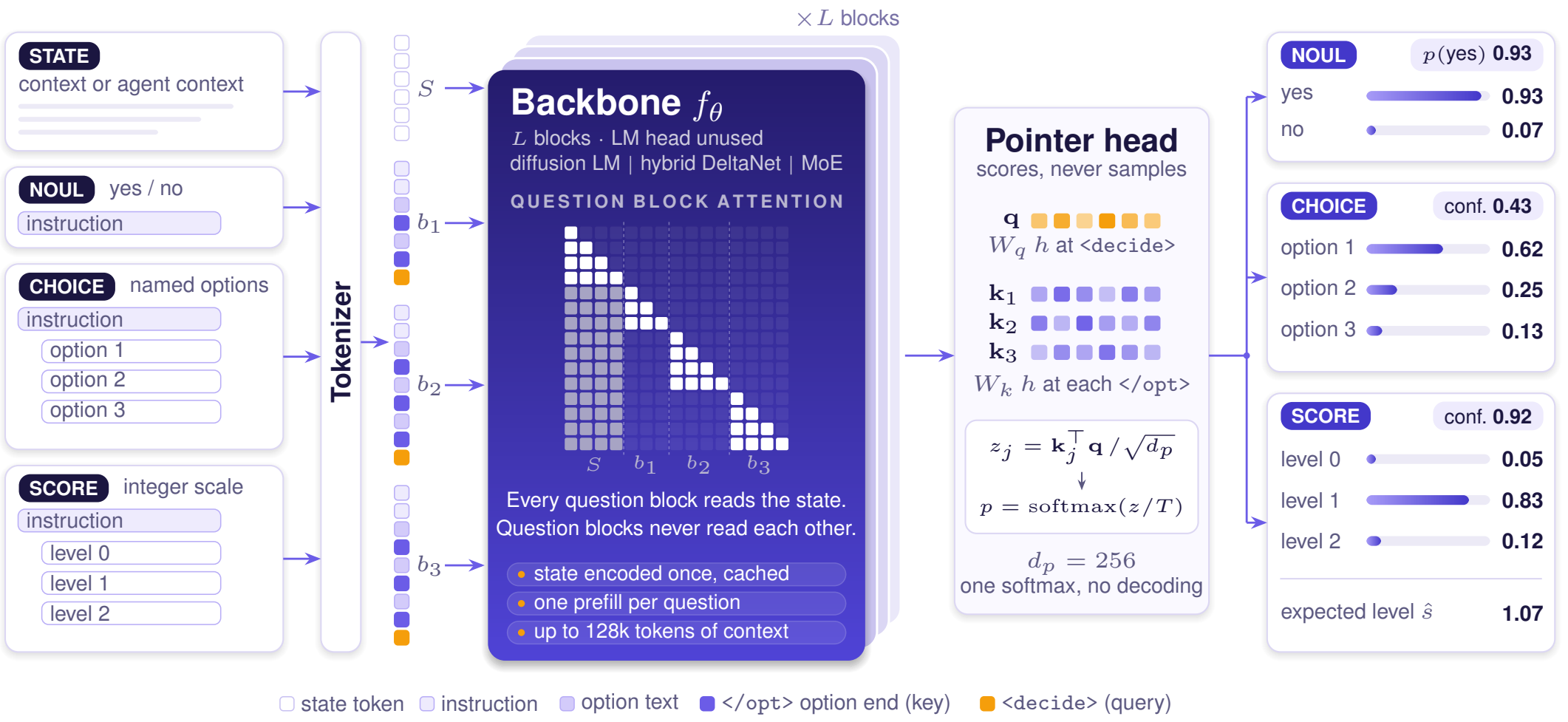}}
\vspace{-0.6em}
\caption{\textbf{\drex pointer architecture} (DLM, v1.5, and v2.0, not including \drex Max, which is shown in \cref{fig:maxarch}).
The state and the questions (left) are tokenized into one sequence that consists of a state segment $S$ followed by one \emph{question block} $b_k$ for each question.
Each question block is a contiguous block of text that contains the instruction, the options, and a final \tok{decide} token, in this order; each option is enclosed between an opening tag \tok{opt} and a closing tag \tok{/opt}.
The backbone $f_\theta$ encodes the sequence under the block-causal pattern shown inside it (\cref{eq:mask}): every token of a question block can attend to all tokens of the state and to the preceding tokens of the same question block, but not to any token of another question block.
The pointer head turns the hidden state at \tok{decide} into a query $\mathbf{q}$ and each \tok{/opt} state into a key $\mathbf{k}_j$, and returns a softmax over exactly the supplied options (right).
The example outputs are consistent with \cref{eq:confidence}: $(0.62 - \nicefrac13)/(1 - \nicefrac13) = 0.43$ and $\hat s = 0.83 + 2 \times 0.12 = 1.07$.}
\label{fig:arch}
\end{figure}

Every \drex pointer model is a pretrained language-model backbone $f_\theta$ plus a pointer head $(W_q, W_k)$ (\cref{fig:arch}).
Nothing is decoded at inference time.
\drex Max uses a different output head and attention pattern and is described in \cref{sec:max}.

\paragraph{Packing and attention.}
A request is serialized, or \emph{packed}, into one token sequence that consists of a state segment followed by one \emph{question block} per question.
Each question block is a contiguous block of text that contains the instruction, the options, and a final \tok{decide} token, in this order; each option is enclosed between an opening tag \tok{opt} and a closing tag \tok{/opt}.
These tags and the other delimiters are special tokens, which do not appear in the text supplied by the user (\cref{eq:packing}).
On attention-only backbones, the sequence is encoded under a block-causal mask (\cref{fig:masks}, Appendix \cref{eq:mask}): every token of a question block can attend to all tokens of the state and to the preceding tokens of the same question block, but not to any token of another question block.
\drex DLM adds bidirectional attention within the state, so that each token in the state can attend to every other token in the state, including later ones.
\drex v1.5 and v2.0 use hybrid models as backbones, which interleave layers with linear attention (Gated DeltaNet layers) with global attention layers.
Hybrid models cannot apply an arbitrary attention mask; for these models, each question is processed as a separate causal sequence that consists of the state followed by that question's block, and the encoding of the state is cached and reused across questions.
Hence, up to floating-point error, $p_k$ depends only on the state and question $k$ (\cref{app:archdetails}).
Adding, removing, or reordering the other questions of a request therefore does not change $p_k$, because the position IDs of the question blocks are set such that, in terms of the position IDs, each question block immediately follows the end of the state.
As a consequence, regardless of whether a question appears alongside other questions for a given input state, the predicted probabilities will be the same for that question's options. Furthermore, the encoding of the state can be cached and reused when answering further questions with the same state.
While the predicted probabilities are invariant to the order of the questions, they are not necessarily invariant to the order of the options within one question, which is the subject of our experiments measuring order robustness (\cref{sec:order}).
\Cref{app:archdetails} gives further details, including an optional option-isolation attention mask that makes the pointer head permutation-equivariant on attention-only backbones.

\subsection{Pointer head}
\label{sec:pointer}

Let $h_{d_k} \in \R^{d}$ be the final-layer hidden state at the \tok{decide} token of question block $k$, and let $h_{o_{k,j}} \in \R^{d}$ be the hidden state at its $j$-th \tok{/opt} token.
The pointer computes
\begin{equation}
  z_{k,j} = \frac{1}{\sqrt{d_p}}\,\bigl(W_k h_{o_{k,j}}\bigr)^{\!\top}\bigl(W_q h_{d_k}\bigr),
  \qquad
  p_k = \softmax\!\left(z_k / T\right),
  \label{eq:pointer}
\end{equation}
with $W_q, W_k \in \R^{d_p \times d}$ and $d_p = 256$ on every released pointer checkpoint, which gives $2 d_p d$ parameters: 2.1M for v1.5 ($d{=}4096$) and 1.0M for v2.0 ($d{=}2048$), under 0.03\% of either backbone.
Because \tok{/opt} sits at the end of its option, causal attention lets its hidden state summarize the whole option in the context of the state, the instruction, and earlier options.
The temperature $T$ is set to $T{=}1$ during training and in the released v1.5 and v2.0 pointer heads.
There is no classification over a global label set.
Options are supplied per request, so a question with three options and a question with 150 options use the same head, and an option never seen in training is scored by the same mechanism as a familiar one.

Inference requires one encoding of the state and one short forward pass over the tokens of each question, and no autoregressive decoding (\cref{alg:serve}, \cref{sec:inference}).

\section{\drex Max: a dense multimodal decision model}
\label{sec:max}
\begin{figure}[!t]
\centering
\resizebox{\textwidth}{!}{\begin{tikzpicture}[x=1cm, y=1cm,
  every node/.style={font=\sffamily\fontsize{6}{7}\selectfont, inner sep=0pt, text=ink},
  lbl/.style={font=\sffamily\fontsize{5.4}{6.4}\selectfont, text=inkmute},
  lbs/.style={font=\sffamily\fontsize{5.0}{6}\selectfont, text=inkmute},
  val/.style={font=\sffamily\bfseries\fontsize{5.6}{6.6}\selectfont, text=ink},
  ttl/.style={font=\sffamily\bfseries\fontsize{7.5}{8.5}\selectfont, text=ink, anchor=west},
  pill/.style={font=\sffamily\bfseries\fontsize{5.1}{6}\selectfont, text=white, rounded corners=3pt,
               inner xsep=2.8pt, inner ysep=1.8pt},
  card/.style={draw=cardline, fill=white, line width=0.4pt, rounded corners=3.5pt,
               blur shadow={shadow blur steps=10, shadow xshift=0pt, shadow yshift=-0.8pt, shadow opacity=16}},
  arr/.style={-{Stealth[length=1.5mm, width=1.15mm]}, line width=0.55pt, draw=arrowc},
  bbt/.style={anchor=west, text=white, font=\sffamily\bfseries\fontsize{8.5}{10}\selectfont},
  bbs/.style={anchor=west, text=white!78!encA, font=\sffamily\fontsize{5.2}{6.2}\selectfont},
]
\def\drexTok#1#2#3{\filldraw[fill=#1, draw=#2, line width=0.3pt, rounded corners=0.9pt] (#3,4.400) rectangle ++(0.135,0.135);}
\def\drexCode#1#2{\filldraw[fill=ink, draw=ink, line width=0.3pt, rounded corners=0.9pt] (#1,4.400) rectangle ++(0.135,0.135);
  \node[text=white, font=\sffamily\bfseries\fontsize{3.6}{4}\selectfont] at (#1+0.0675,4.4675) {#2};}
\def\drexSpan#1#2#3{\draw[inkmute!55, line width=0.35pt] (#1,4.600) -- ++(0,0.05) -- (#2,4.650) -- ++(0,-0.05);
  \node[lbl, anchor=south] at ({(#1+#2)/2},4.700) {#3};}
\def\drexBar#1#2#3#4{%
  \node[lbl, anchor=west] at (#1+0.12,#2) {#3};
  \fill[track, rounded corners=1.5pt] (#1+0.88,#2-0.045) rectangle (#1+1.42,#2+0.045);
  \shade[left color=barB, right color=barA, rounded corners=1.5pt] (#1+0.88,#2-0.045) rectangle ({#1+0.88+0.54*#4},#2+0.045);
  \node[val, anchor=east] at (#1+1.78,#2) {#4};}

\node[ttl] at (0.20,6.30) {(a) Pointer models};
\node[pill, fill=ink, anchor=west] at (2.95,6.30) {DLM / v1.5 / v2.0};

\foreach \x in {0.30,0.47,...,1.33} {\drexTok{stok}{sline}{\x}}
\foreach \x in {1.615,1.785,1.955,2.125} {\drexTok{qtok}{qline}{\x}}
\foreach \x in {2.295,2.465,2.635} {\drexTok{otok}{qline}{\x}}
\drexTok{etok}{etok}{2.805}
\foreach \x in {2.975,3.145,3.315} {\drexTok{otok}{qline}{\x}}
\drexTok{etok}{etok}{3.485}
\drexTok{dtok}{dtok}{3.655}
\begin{scope}[opacity=0.45]
\foreach \x in {3.950,4.120} {\drexTok{qtok}{qline}{\x}}
\drexTok{otok}{qline}{4.290}\drexTok{etok}{etok}{4.460}
\drexTok{otok}{qline}{4.630}\drexTok{etok}{etok}{4.800}
\drexTok{dtok}{dtok}{4.970}
\end{scope}
\node[lbl] at (5.40,4.4675) {$\cdots$};

\drexSpan{0.30}{1.455}{state $S$}
\drexSpan{1.615}{3.790}{question 1 ($b_1$)}
\drexSpan{3.950}{5.105}{question 2 ($b_2$)}

\filldraw[fill=encback!22, draw=white, rounded corners=3.4pt] (0.26,3.09) rectangle (6.26,3.99);
\shade[top color=encA, bottom color=encB, rounded corners=3.2pt,
  blur shadow={shadow blur steps=8, shadow xshift=0pt, shadow yshift=-0.8pt, shadow opacity=22}]
  (0.20,3.15) rectangle (6.20,3.95);
\node[bbt] at (0.38,3.55) {Backbone $f_\theta$};
\node[bbs] at (2.55,3.76) {each question attends to the state,};
\node[bbs] at (2.55,3.56) {never to another question;};
\node[bbs] at (2.55,3.36) {the state is encoded once and cached};
\foreach \x in {0.8775,2.7025,4.5275} \draw[arr] (\x,4.38) -- (\x,3.97);

\filldraw[card, fill=headfill] (0.20,1.30) rectangle (4.00,2.85);
\node[anchor=west, font=\sffamily\bfseries\fontsize{7}{8}\selectfont] at (0.35,2.62) {Pointer head};
\node[lbs, anchor=west] at (0.35,2.36) {query $\mathbf{q}$ at \tok{decide}};
\node[lbs, anchor=west] at (0.35,2.18) {key $\mathbf{k}_j$ at each \tok{/opt}};
\foreach \x/\o in {2.80/60, 2.98/85, 3.16/45, 3.34/100, 3.52/70, 3.70/55} \fill[dtok!\o, rounded corners=0.8pt] (\x,2.49) rectangle ++(0.15,0.13);
\foreach \x/\o in {2.80/55, 2.98/90, 3.16/70, 3.34/40, 3.52/80, 3.70/60} \fill[etok!\o, rounded corners=0.8pt] (\x,2.27) rectangle ++(0.15,0.13);
\foreach \x/\o in {2.80/80, 2.98/45, 3.16/95, 3.34/60, 3.52/50, 3.70/75} \fill[etok!\o, rounded corners=0.8pt] (\x,2.09) rectangle ++(0.15,0.13);
\filldraw[fill=white, draw=cardline, line width=0.4pt, rounded corners=3pt] (0.35,1.38) rectangle (3.85,2.01);
\node[font=\fontsize{6}{7}\selectfont] at (2.10,1.78) {$z_j = \mathbf{k}_j^{\top}\mathbf{q}\,/\sqrt{d_p}$};
\node[font=\fontsize{6}{7}\selectfont] at (2.10,1.51) {$p = \mathrm{softmax}(z/T)$};
\draw[arr, draw=etok] (2.8725,3.13) -- (2.8725,2.64);
\draw[arr, draw=etok] (3.5525,3.13) -- (3.5525,2.64);
\draw[arr, draw=dtok] (3.7225,3.13) -- (3.7225,2.64);

\filldraw[card] (4.30,1.30) rectangle (6.20,2.85);
\node[pill, fill=barA, anchor=west] at (4.42,2.62) {CHOICE};
\drexBar{4.30}{2.22}{option 1}{0.62}
\drexBar{4.30}{1.86}{option 2}{0.38}
\draw[arr] (4.00,1.69) -- (4.28,1.69);

\begin{scope}[shift={(7.00,0)}]
\node[ttl] at (0.20,6.30) {(b) \drex Max};

\filldraw[fill=ifill, draw=iline, line width=0.4pt, rounded corners=1.2pt] (0.02,5.55) rectangle (0.40,5.97);
\fill[dtok] (0.21,5.86) circle (0.04);
\fill[itok, opacity=0.75] (0.05,5.58) -- (0.15,5.74) -- (0.23,5.65) -- (0.29,5.74) -- (0.37,5.58) -- cycle;
\filldraw[fill=white, draw=iline, line width=0.4pt, rounded corners=1.2pt] (0.45,5.55) rectangle (0.79,5.97);
\fill[itok, opacity=0.65] (0.50,5.86) rectangle (0.74,5.91);
\fill[itok, opacity=0.35] (0.50,5.77) rectangle (0.70,5.81);
\fill[itok, opacity=0.35] (0.50,5.68) rectangle (0.72,5.72);
\fill[itok, opacity=0.35] (0.50,5.59) rectangle (0.66,5.63);
\filldraw[fill=ink, rounded corners=1.2pt] (0.84,5.55) rectangle (1.20,5.97);
\fill[itok, opacity=0.55] (0.88,5.64) rectangle (0.98,5.89);
\fill[itok, opacity=0.75] (1.00,5.64) rectangle (1.09,5.89);
\fill[itok] (1.11,5.64) rectangle (1.17,5.89);
\node[lbl, anchor=west] at (1.28,5.76) {input images};
\filldraw[fill=ifill, draw=iline, line width=0.4pt, rounded corners=1.8pt] (-0.24,5.05) rectangle (1.78,5.37);
\node[font=\sffamily\bfseries\fontsize{5.2}{5.8}\selectfont] at (0.77,5.21) {image encoder (ViT)};
\draw[arr] (0.61,5.53) -- (0.61,5.40);
\draw[arr] (0.61,5.03) -- (0.61,4.66);

\foreach \x in {0.30,0.47,0.64,0.81} {\drexTok{itok}{iline}{\x}}
\foreach \x in {1.105,1.275,...,2.13} {\drexTok{stok}{sline}{\x}}
\foreach \x in {2.420,2.590,2.760,2.930,3.100,3.270} {\drexTok{qtok}{qline}{\x}}
\drexCode{3.565}{A}
\foreach \x in {3.735,3.905,4.075,4.245} {\drexTok{otok}{qline}{\x}}
\drexCode{4.540}{B}
\foreach \x in {4.710,4.880,5.050,5.220} {\drexTok{otok}{qline}{\x}}
\drexTok{dtok}{dtok}{5.655}

\drexSpan{1.105}{2.260}{state $S$}
\drexSpan{2.420}{3.405}{instruction}
\drexSpan{3.565}{4.380}{option 1}
\drexSpan{4.540}{5.355}{option 2}
\node[lbl, anchor=south] at (5.7225,4.700) {\tok{ans}};

\filldraw[fill=encback!22, draw=white, rounded corners=3.4pt] (0.26,3.09) rectangle (6.26,3.99);
\shade[top color=encA, bottom color=encB, rounded corners=3.2pt,
  blur shadow={shadow blur steps=8, shadow xshift=0pt, shadow yshift=-0.8pt, shadow opacity=22}]
  (0.20,3.15) rectangle (6.20,3.95);
\node[bbt] at (0.38,3.55) {Backbone $f_\theta$};
\node[bbs] at (2.40,3.76) {48 Gated DeltaNet layers: left to right;};
\node[bbs] at (2.40,3.56) {16 full-attention layers: both directions;};
\node[bbs] at (2.40,3.36) {one sequence per question, no shared cache};
\foreach \x in {0.6175,1.6825,2.9125,3.9725,4.9475,5.7225} \draw[arr] (\x,4.38) -- (\x,3.97);

\filldraw[card, fill=headfill] (2.40,1.30) rectangle (6.20,2.85);
\node[anchor=west, font=\sffamily\bfseries\fontsize{7}{8}\selectfont] at (2.55,2.62) {Readout head};
\node[lbs, anchor=west] at (2.55,2.36) {hidden state $h$ at \tok{ans}};
\node[lbs, anchor=west] at (2.55,2.18) {linear $W_r$, no bias};
\foreach \x/\o in {4.85/55, 5.03/85, 5.21/45, 5.39/70, 5.57/100, 5.75/60} \fill[dtok!\o, rounded corners=0.8pt] (\x,2.49) rectangle ++(0.15,0.13);
\draw[-{Stealth[length=1mm]}, line width=0.4pt, draw=inkmute] (5.285,2.47) -- (5.285,2.25);
\foreach \x/\c/\o in {4.85/A/100, 5.03/B/100, 5.21/C/25, 5.39/D/25, 5.57/E/25} {
  \fill[barA!\o, rounded corners=0.8pt] (\x,2.09) rectangle ++(0.15,0.13);
  \node[text=white, font=\sffamily\bfseries\fontsize{3.6}{4}\selectfont] at (\x+0.075,2.155) {\c};}
\node[lbs] at (5.95,2.155) {$\cdots$};
\filldraw[fill=white, draw=cardline, line width=0.4pt, rounded corners=3pt] (2.55,1.38) rectangle (6.05,2.01);
\node[font=\fontsize{6}{7}\selectfont] at (4.30,1.78) {$\ell = W_r h$, one logit per code};
\node[font=\fontsize{6}{7}\selectfont] at (4.30,1.51) {$p = \mathrm{softmax}(\ell_1, \dots, \ell_K)$};
\draw[arr, draw=dtok] (5.7225,3.13) -- (5.7225,2.64);

\filldraw[card] (0.20,1.30) rectangle (2.10,2.85);
\node[pill, fill=barA, anchor=west] at (0.32,2.62) {CHOICE};
\drexBar{0.20}{2.22}{option 1}{0.71}
\drexBar{0.20}{1.86}{option 2}{0.29}
\draw[arr] (2.40,1.69) -- (2.12,1.69);
\end{scope}

\def\drexKey#1#2#3#4{\filldraw[fill=#2, draw=#3, line width=0.3pt, rounded corners=0.9pt] (#1,0.785) rectangle ++(0.13,0.13);
  \node[lbl, anchor=west] at (#1+0.19,0.85) {#4};}
\drexKey{0.30}{stok}{sline}{state token}
\drexKey{1.60}{qtok}{qline}{instruction}
\drexKey{2.85}{otok}{qline}{option text}
\drexKey{4.15}{etok}{etok}{\tok{/opt} option end (key)}
\drexKey{6.65}{dtok}{dtok}{\tok{decide} / \tok{ans} answer position}
\drexKey{10.05}{itok}{iline}{image token}
\filldraw[fill=ink, draw=ink, line width=0.3pt, rounded corners=0.9pt] (11.40,0.785) rectangle ++(0.13,0.13);
\node[text=white, font=\sffamily\bfseries\fontsize{3.6}{4}\selectfont] at (11.465,0.85) {A};
\node[lbl, anchor=west] at (11.59,0.85) {option code (one token)};
\end{tikzpicture}}
\vspace{-0.6em}
\caption{\textbf{Two ways to score the options of a question.}
Each small square is one token, colored by its type as in \cref{fig:arch}; the legend at the bottom lists the token types.
\textbf{(a)} The pointer models (DLM, v1.5, and v2.0) encode one sequence per request: the state $S$, followed by one question block per question.
Each question block attends to the state but not to other question blocks, so the state is encoded once and shared by all questions.
The pointer head scores each option as the scaled dot product of the hidden state at the option's \tok{/opt} token (the key) and the hidden state at the \tok{decide} token of its question block (the query).
\textbf{(b)} \drex Max encodes each question as its own sequence: the image tokens that an image encoder (ViT) produces from the input images, the state $S$, the instruction, the options, and a final answer marker, the single token \tok{ans}.
Each option is written after its option code (\texttt{A}, \texttt{B}, \dots), which is a single token, and a colon.
The 48 Gated DeltaNet layers of the backbone process the sequence from left to right, and its 16 full-attention layers (every fourth layer) attend in both directions.
A readout head, i.e., a bias-free linear layer, maps the final hidden state at \tok{ans} to one logit per option code (255 in total), and a softmax over the logits of the $K$ supplied options gives their probabilities.
The pointer models trade accuracy for a shared, cacheable state; \drex Max trades the cache for full bidirectional context in every question.}

\label{fig:maxarch}
\end{figure}
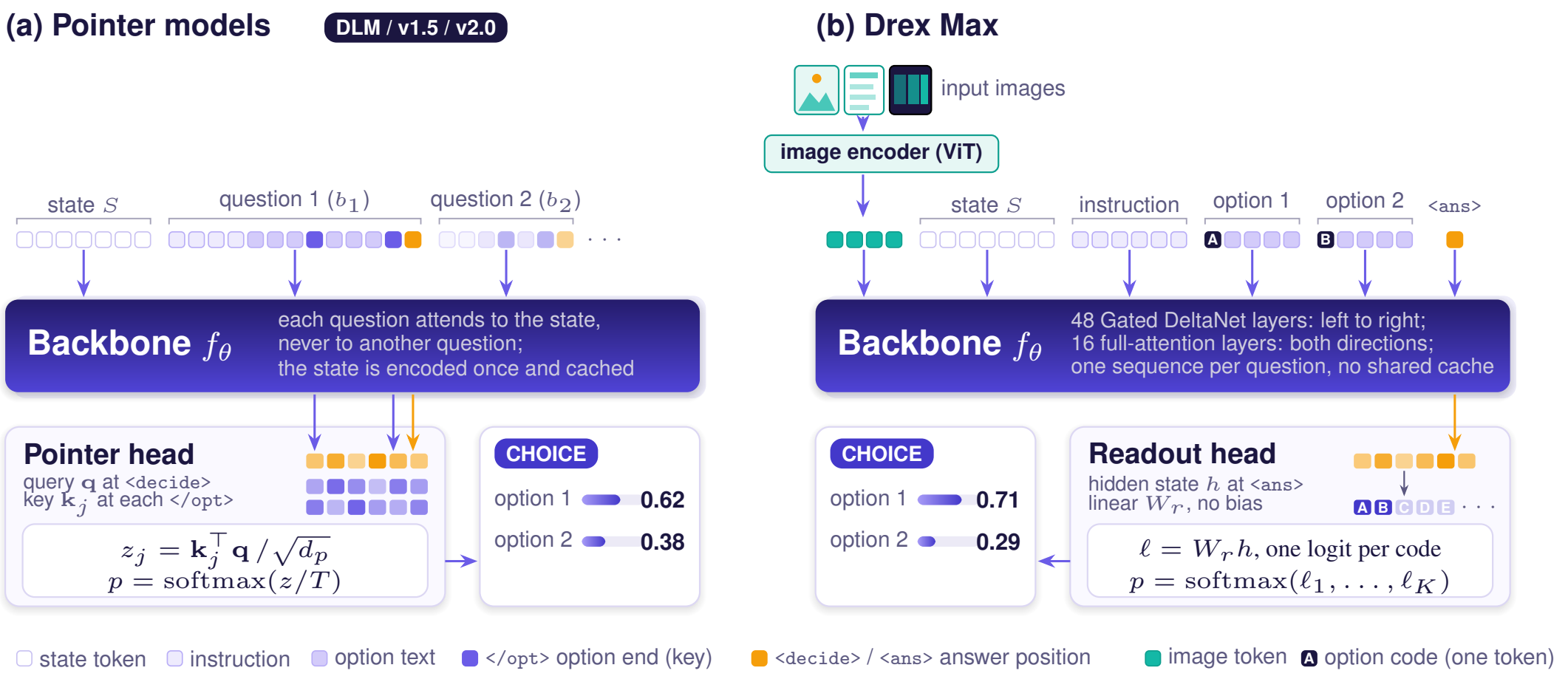

The pointer models answer a question by comparing hidden states \emph{inside} one causal encoding of the context.
\drex Max asks a different question of the architecture: what is the effect on the accuracy and the inference cost of the resulting decision model if every question can attend to the entire request in both directions, and the backbone is a dense 27B model with an image encoder?
The attention pattern and output head of \drex Max follow the design of the publicly released \texttt{pplx-decider} model~\citep{perplexity2026decider}.
This section describes the architecture; we train its weights from the Qwen3.8-27B base model with the recipe of \cref{sec:maxtrain}.

\subsection{Backbone and attention pattern}
\label{sec:maxattn}

The backbone is Qwen3.8-27B~\citep{qwen38,yang2025qwen3}, a dense model with $L{=}64$ layers and hidden size $d{=}5{,}120$: 48 Gated DeltaNet layers and 16 full-attention layers, with every fourth layer being a full-attention layer.
Let $\mathcal{F} \subset \{1,\dots,64\}$ be the indices of the full-attention layers.
\drex Max removes the causal mask in exactly those layers: for a sequence $x$ of length $n$,
\begin{equation}
  M^{(\ell)}_{ij} =
  \begin{cases}
    1 & \text{if } \ell \in \mathcal{F} \quad\text{(non-causal: every token attends to the whole sequence)},\\
    \mathds{1}[j \le i] & \text{if } \ell \notin \mathcal{F} \quad\text{(DeltaNet recurrence, left to right)}.
  \end{cases}
  \label{eq:maxmask}
\end{equation}
Because the DeltaNet layers are untouched, information still flows left to right in three of every four layers, but every fourth layer lets early tokens, such as the context and the start of the option list, attend to any token in the sequence, including all subsequent tokens, such as the options and the instruction.
The trained weights give the intended scores only when the mask of \cref{eq:maxmask} is utilized in the forward pass; performing a forward pass with the same weights and the original causal mask in all layers does not reproduce the intended scores.

\paragraph{Choosing the non-causal ratio.}
How many layers to make non-causal is a design choice.
To make it, we used a meta-learning probe: a separate hypernetwork generates weight edits for the backbone, and for each candidate mask we apply these edits and measure the resulting catastrophic forgetting on held-out decisions, alongside downstream decision accuracy.
We compared non-causal ratios of [0, 1/16, 1/8 and 1/4] of the layers.
Making one in every four layers non-causal, i.e.\ all of $\mathcal{F}$, gave the lowest forgetting and the highest decision accuracy, so \drex Max uses the mask of \cref{eq:maxmask}.
Since the DeltaNet layers are recurrent, ratios above $1/4$ would require replacing DeltaNet layers with full attention, which we leave to future work.

\subsection{Readout head}
\label{sec:readout}

In \drex Max, each question $k$ is encoded as one sequence
\begin{equation}
  x_k \;=\; \bigl[\, \mathcal{I};\; S;\; u_k;\; \gamma_1{:}\,o_{k,1};\; \dots;\; \gamma_{K_k}{:}\,o_{k,K_k};\; \tok{ans} \,\bigr],
  \label{eq:maxprompt}
\end{equation}
where semicolons separate consecutive parts of the sequence: the images $\mathcal{I}$, the state $S$, the instruction $u_k$, the options, and a final answer marker $\tok{ans}$.
The notation $\gamma_j{:}\,o_{k,j}$ means that the text of option $j$ is written after its option code $\gamma_j$ and a colon, for example ``\texttt{A}: refund request''.
The option codes $\gamma_1,\dots,\gamma_{255} = \texttt{A},\dots,\texttt{Z},\texttt{AA},\dots$ are each a single token in the vocabulary, and $\tok{ans}$ is a fixed special token that ends the sequence; the answer is read from the hidden state at its position.
Let $h_k \in \R^{d}$ be the final-layer hidden state at the last position of $x_k$.
The readout head is a bias-free linear layer $W_r \in \R^{255 \times d}$ that maps $h_k$ to a vector of 255 logits, one for each option code; only the first $K_k$ logits, which belong to the $K_k$ options of the question, are kept:
\begin{equation}
  z_k = W_r h_k \in \R^{255},
  \qquad
  p_{k,j} = \frac{\exp(z_{k,j}/T)}{\sum_{i=1}^{K_k} \exp(z_{k,i}/T)}, \qquad j = 1,\dots,K_k,
  \label{eq:readout}
\end{equation}
where $T > 0$ is the temperature parameter.
We set $T{=}1.009$, a value fitted on held-out records after the last training stage (\cref{sec:maxtrain}); it is close to $1$, so it changes the probabilities only slightly, and it does not change the most probable option.
The readout head has $255 \times 5{,}120 \approx 1.3$M parameters, comparable to the pointer head.
The output format described in \cref{sec:problem} is unchanged: $p_k$ is supported on $O_k$ by construction, $K \le 255$, and the confidence statistics of \cref{eq:confidence} are computed from $p_k$ in the same way.

\paragraph{Comparison with the pointer head.}
The two heads differ in how they represent each option.
In the pointer head, option $j$ is represented by its own hidden state $h_{o_{k,j}}$, which summarizes the text of the option, and the score of the option is computed from this hidden state.
In the readout head, option $j$ is identified only by the code $\gamma_j$ that precedes it, so the backbone must learn to associate each code with the text of its option, which is the symbol-binding problem of multiple-choice prompting~\citep{robinson2023mcqa}, and a permutation of the options changes which of the 255 logits corresponds to which option.
Order robustness is therefore entirely learned in \drex Max, which is why we measure it directly (\cref{sec:order}).
In exchange, the readout head lets the full-attention layers compare all options with one another and with the context in both directions before the answer is computed, which the causal pointer cannot.

\subsection{Images}
\label{sec:maxvision}

\drex Max keeps the image encoder of the backbone, the part of the model that converts images into token embeddings.
Each image in $\mathcal{I}$ is split into patches, encoded by the vision transformer~\citep{dosovitskiy2021vit}, and merged into visual tokens that enter the sequence ahead of the state (\cref{eq:maxprompt}), as in the standard encoder-plus-LM layout~\citep{liu2023llava}.
Because each question is encoded as a distinct sequence containing the images of the request, every question in a request can attend to all of the images in that request. A request can contain both questions that depend on the images and questions that do not.
Video is handled by sampling frames and passing them as images, since the model has no temporal encoder.

\paragraph{Cost.}
Because the full-attention layers are non-causal, the representation of the state depends on the question, so it cannot be cached across questions (\cref{sec:maxcost}): \drex Max pays for the state once per question, the pointer models once per request.
For short contexts the difference is small: at launch, the median latency of a single request was 183\,ms for \drex Max, against 151\,ms for \drex v2.0 and 145\,ms for \drex v1.5 (see \cref{tab:efficiency}, which describes how latency was measured).
For many questions over long documents, the pointer models remain the cheaper choice.

\section{Training}
\label{sec:train}
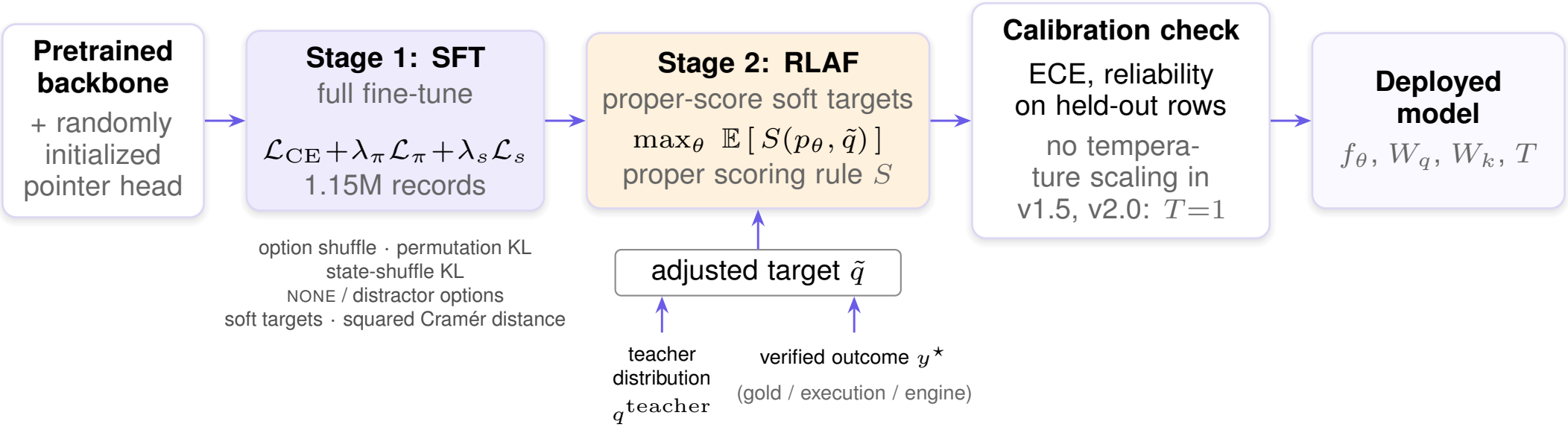
\begin{figure}[t]
\centering
\resizebox{\textwidth}{!}{\begin{tikzpicture}[
  x=1cm, y=1cm,
  stage/.style={draw=cardline, line width=0.5pt, rounded corners=3.5pt, blur shadow={shadow blur steps=10, shadow xshift=0pt, shadow yshift=-0.8pt, shadow opacity=16}, align=center, inner sep=4pt,
                minimum height=1.55cm, text width=2.35cm, font=\sffamily\scriptsize},
  small/.style={draw=black!45, line width=0.4pt, rounded corners=1.5pt, align=center, inner sep=2.5pt,
                font=\sffamily\scriptsize, text width=1.75cm},
  arr/.style={-{Stealth[length=1.5mm, width=1.15mm]}, line width=0.55pt, draw=arrowc},
]
\node[stage, fill=white, text width=1.5cm] (s0) at (0,0)
  {\textbf{Pretrained}\\\textbf{backbone}\\[2pt]{\color{black!60}+ randomly initialized pointer head}};
\node[stage, fill=qtok, right=0.36cm of s0] (s1)
  {\textbf{Stage 1: SFT}\\[1pt]{\color{black!60}full fine-tune}\\[6pt]
   $\mathcal{L}_{\mathrm{CE}}+\lambda_\pi\mathcal{L}_{\pi}+\lambda_s\mathcal{L}_{s}$\\[1pt]
   {\color{black!60}1.15M records}};
\node[stage, fill=dtok!14, right=0.36cm of s1, text width=2.75cm] (s2)
  {\textbf{Stage 2: RLAF}\\[1pt]{\color{black!60}proper-score soft targets}\\[2pt]
   $\max_\theta\ \mathbb{E}\,[\,S(p_\theta,\tilde q)\,]$\\[1pt]
   {\color{black!60}proper scoring rule $S$}};
\node[stage, fill=white, right=0.36cm of s2] (s3)
  {\textbf{Calibration check}\\[3pt]
   ECE, reliability\\on held-out rows\\[2pt]
   {\color{black!60}no temperature scaling in v1.5, v2.0: $T{=}1$}};
\node[stage, fill=phfill, right=0.36cm of s3, text width=1.95cm] (s4)
  {\textbf{Deployed}\\\textbf{model}\\[2pt]{\color{black!60}$f_\theta$, $W_q$, $W_k$, $T$}};
\draw[arr] (s0) -- (s1);
\draw[arr] (s1) -- (s2);
\draw[arr] (s2) -- (s3);
\draw[arr] (s3) -- (s4);

\node[draw=black!45, line width=0.4pt, rounded corners=1.5pt, fill=white, font=\sffamily\scriptsize, align=center,
      text width=2.35cm, inner sep=2.5pt, below=0.35cm of s2] (adj) {adjusted target $\tilde q$};
\draw[arr] (adj.north) -- (s2.south);
\node[font=\sffamily\tiny, align=center, text width=1.6cm, anchor=north] (tq) at ($(adj.south)+(-0.85,-0.32)$) {teacher\\distribution\\[3pt]$q^{\mathrm{teacher}}$};
\node[font=\sffamily\tiny, align=center, text width=2.2cm, anchor=north] (vy) at ($(adj.south)+(0.85,-0.32)$) {verified outcome $y^\star$\\[3pt]{\color{black!60}(gold / execution / engine)}};
\draw[arr] (tq.north) -- (tq.north |- adj.south);
\draw[arr] (vy.north) -- (vy.north |- adj.south);

\node[font=\sffamily\tiny, align=center, anchor=north, black!70] at ($(s1.south)+(0,-0.12)$)
  {option shuffle $\cdot$ permutation KL\\state-shuffle KL\\\textsc{none} / distractor options\\soft targets $\cdot$ squared Cram\'er distance};
\end{tikzpicture}}
\caption{\textbf{Post-training pipeline} of the pointer models (shown for \drex v1.5).
Stage~1 fine-tunes every weight in the backbone and a randomly initialized pointer head by optimizing the supervised objective of \cref{eq:sft}.
Stage~2 (RLAF) replaces one-hot labels with an adjusted target $\tilde q$, which is constructed from a teacher distribution and the verified outcome. RLAF then maximizes the expected reward $R$ of \cref{eq:proper}, which is largest when the predicted distribution equals $\tilde q$.
\drex Max is trained with full-parameter SFT (with the same consistency losses), then the low-rank variant of RLAF as described in \cref{sec:sdrl}, then SFT of the image encoder (\cref{sec:maxtrain}).}
\label{fig:train}
\end{figure}

Each training example is a labeled \sysone record, i.e., a state with one or more questions, not a string of tokens.
All \drex models share the training data of \cref{sec:universe} and the supervised objective of \cref{sec:objective}, and are then trained with reinforcement learning from adjusted feedback (RLAF).
The pointer models use the RLAF formulation of \cref{sec:rlaf} (\cref{fig:train}); \drex v2.0 additionally includes a task-specific specialization stage between supervised fine-tuning and RLAF to improve its performance on a particular family of tasks (\cref{sec:s2e}).
\drex Max uses a KL-regularized variant of RLAF (\cref{sec:sdrl}) and adds a stage that trains its image encoder; \cref{sec:maxtrain} gives its full recipe.
\Cref{sec:family} lists the backbones, heads, and inference software of the four models.

\subsection{Training data}
\label{sec:universe}

\begin{figure}[t]
\centering
$\vcenter{\hbox{\rotatebox{90}{\small Data domains}}}\;\vcenter{\hbox{\includegraphics[width=0.96\textwidth]{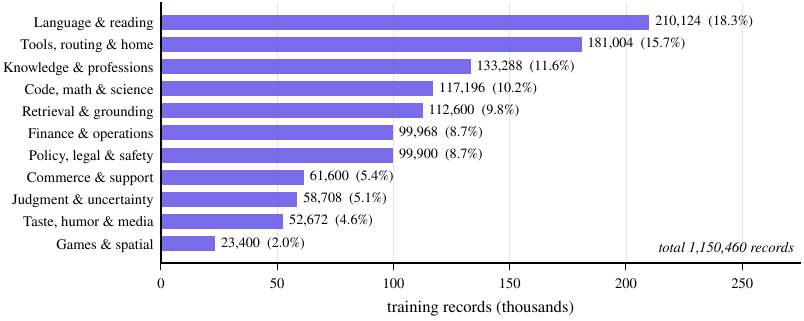}}}$
\caption{\textbf{Training data} of \drex v1.5: 1{,}150{,}460 verified text decision records over eleven domains. Each bar shows the number of records in the respective domain.}
\label{fig:mix}
\end{figure}

We assembled the training data of \drex v1.5, consisting of a mix of exactly \textbf{1{,}150{,}460} textual decision records across eleven domains (\cref{fig:mix}; exact counts in \cref{tab:mix}).
Most records are distilled from stronger teachers; the remainder come from open-source datasets and from verifiable procedurally-generated data (executed programs, outputs from game engines).
Every label is checked against a gold answer, an execution result, or an engine.
Decision Index and JevBench evaluation items are excluded; the training splits of open-source datasets that are used as public benchmarks are included in the mix.
Visual question-answering data for the same API is collected separately and is not part of this text mix.

\subsection{Supervised fine-tuning}
\label{sec:objective}

For an example with predicted logits $z \in \R^K$ (pointer logits, or the first $K$ readout logits for \drex Max), target $t \in \Delta^{K-1}$ (one-hot, or a teacher distribution when one is available), and index $y$ of the correct level (if the question is a \texttt{score} question), the per-question loss is
\begin{equation}
  \ell_{\mathrm{sup}}(z) = -\sum_{j} t_j \log p_j
  \;+\; \lambda_{\mathrm{ord}}\,\mathds{1}[\tau{=}\texttt{score}]\,\mathrm{D}_{\mathrm{C}}^{2}(p, e_y),
  \label{eq:sup}
\end{equation}
with $p = \softmax(z)$ and levels indexed from $1$ and $\mathrm{D}_{\mathrm{C}}^{2}(p, e_y)$ is the squared Cram\'er distance between $p$ and $e_y$, where $e_y$ is the one-hot distribution on level $y$, whose cumulative distribution is $\mathds{1}[i \ge y]$. Given two distributions $p, q \in \Delta^{K-1}$ over integers, the squared \emph{Cram\'er distance}~\citep{rv_bellemare2017cramer} between $p$ and $q$ is
\begin{equation}
  \mathrm{D}_{\mathrm{C}}^{2}(p, q) = \frac{1}{K-1}\sum_{i=1}^{K-1}\Bigl(\textstyle\sum_{j\le i} p_j - \sum_{j\le i} q_j\Bigr)^{2}.
  \label{eq:cramer}
\end{equation}
This second term penalizes a prediction whose cumulative distribution differs from that of the correct level.
In the forecast-verification literature, this quantity is known as the ranked probability score~\citep{gneiting2007proper}.
The options of every \texttt{choice} question with $K \ge 3$ are shuffled each time a record is seen.
Two consistency regularizers are meant to incentivize robustness to the order of options ($\ell_\pi$) and to the presentation of the state ($\ell_s$), for backbones that cannot isolate options architecturally.
Let $\pi$ be a random permutation of the options and let $\tilde{S}$ be the state with its sentence or fields shuffled uniformly; with $\sKL(a,b) = \tfrac12[\KL(a\|b) + \KL(b\|a)]$,
\begin{equation}
  \ell_{\pi} = \sKL\bigl(p(\cdot \mid S, b),\; \pi^{-1} p(\cdot \mid S, \pi b)\bigr),
  \qquad
  \ell_{s} = \KL\bigl(\operatorname{sg}[p(\cdot \mid S, b)] \,\|\, p(\cdot \mid \tilde S, b)\bigr),
  \label{eq:consistency}
\end{equation}
where $\operatorname{sg}$ stops gradients.
The Stage~1 objective is
\begin{equation}
  \mathcal{L}_{\mathrm{SFT}} = \E\bigl[\ell_{\mathrm{sup}} + \lambda_\pi \ell_\pi + \lambda_s \ell_s\bigr].
  \label{eq:sft}
\end{equation}
In a random fraction of the training records, we insert an extra \textsc{none} option or a distractor option, so that the model learns not to force a choice when none of the options applies.
In v1.5, Stage~1 updates every weight in the backbone and the pointer head.

\subsection{Reinforcement learning from adjusted feedback}%
\label{sec:rlaf}

In supervised training, labels are one-hot even when the honest answer is uncertain, which trains over-confidence~\citep{muller2019labelsmoothing}.
RLAF instead builds, for each question, an \emph{adjusted target} $\tilde q \in \Delta^{K-1}$: a teacher proposes a distribution $q^{\mathrm{teacher}}$ over the options of the question, which is then adjusted by the verified outcome $y^\star$ (a gold label, an execution result, or an output from a game engine), so that $\tilde q$ is consistent with what is known to be true while retaining the teacher's uncertainty among the remaining options.
The reward is a \emph{proper} scoring rule applied to the prediction and $\tilde q$.
A scoring rule is called \emph{proper} if, when outcomes are drawn from the distribution $\tilde q$, its expected value is maximized by the prediction $p = \tilde q$~\citep{gneiting2007proper}.
Our training pipeline for pointer models uses the reward
\begin{equation}
  R(p, \tilde q) = \sum_j \tilde q_j \log p_j \;+\; w_{\mathrm{sph}}\,\frac{\tilde q^{\top} p}{\lVert p \rVert_2}
  \;-\; w_{\mathrm{cr}}\,\mathds{1}[\tau{=}\texttt{score}]\,\mathrm{D}_{\mathrm{C}}^{2}(p, \tilde q),
  \label{eq:proper}
\end{equation}
with $w_{\mathrm{sph}} = 0.5$ and $w_{\mathrm{cr}} = 1$, where $\mathrm{D}_{\mathrm{C}}^{2}$ is the squared Cram\'er distance of \cref{eq:cramer}.
The goal of this training stage is to find the model parameters $\theta$ that maximize the expected reward,
\begin{equation}
  \max_\theta \; \E\bigl[R(p_\theta, \tilde q)\bigr].
  \label{eq:rlafobj}
\end{equation}
Because each term is a proper scoring rule~\citep{gneiting2007proper}, the expected reward is maximized at $p = \tilde q$, so the objective incentivizes high accuracy and calibration together rather than trading one for the other.
We optimize \cref{eq:proper} by direct differentiation, which makes it a deterministic loss on soft targets (cross-entropy with respect to $\tilde q$ plus spherical and squared Cram\'er distance terms); the reinforcement lies in how $\tilde q$ is built.
We also implemented a REINFORCE estimator~\citep{williams1992reinforce} of the gradient, but we found that it does not outperform direct differentiation (\cref{sec:limits}).
For \drex Max, a variant of this stage replaces the one described here; \cref{sec:sdrl} formulates it as reinforcement learning with a KL-regularized reward.
For \drex v2.0, \cref{sec:dynamics} reports benchmark scores at checkpoints before and after this stage.

\subsection{Specialization stages}
\label{sec:specialist}

For some task families on which the model remains weak after supervised fine-tuning, we perform an additional stage of training, which we call a \emph{specialization stage}.
It is performed after supervised fine-tuning and before RLAF, and continues training from the final SFT checkpoint, resulting in a model that is specialized for the task family of interest.
Each specialization stage targets a single task family; and we sometimes apply multiple specialization stages to a model, in sequence, with the aim of each stage being to improve the model's performance in one task family. However, among the released models, the specialization stage is only used in the training of \drex v2.0~(\cref{sec:dynamics}).
The specialization stage is not used in the training of \drex v1.5 and \drex Max.

The specialization stage trains low-rank adapters~\citep{hu2022lora} rather than all weights.
Its training data consist of: (1) decisions in the target task family whose correct answers are computed by a solver, (2) difficult examples from that family which are mined from existing data, and (3) replayed examples from earlier training stages, which prevent the model from forgetting other tasks.
For the same purpose, we add a KL penalty $\KL(p_{\text{ref}} \,\|\, p_\theta)$ to the loss, where $p_{\text{ref}}$ is the distribution predicted by the checkpoint at the start of this training stage; this term penalizes divergence from that checkpoint.
The final checkpoint is selected on a held-out split of the stage's own training data, not on any benchmark.

\paragraph{Contamination filtering.}
No evaluation example from Decision Index or JevBench is used for training, and we drop training records whose state exactly matches, or shares more than two 8-word sequences with, any benchmark example. These filters miss two kinds of contamination: paraphrased benchmark items, and knowledge of benchmarks that may be distilled from teacher models that were themselves trained on benchmark data.

\subsection{RLAF for \drex Max}
\label{sec:sdrl}

The RLAF stage of \drex Max refines the SFT model with feedback that is more informative than a one-hot label. This feedback consists of a verified outcome (where one exists) and the opinion of a judge.
For a question with input $x$ and options $O$, $|O| = K$, let $\pi_\theta(a \mid x) = p_{\theta,a}$ be the policy, which is the model's predicted distribution as described in \cref{eq:readout}; let $\pi_0$ be the frozen SFT model; let $y^\star$ be the verified outcome when one is available; and let $q^{J} \in \Delta^{K-1}$ be the predicted distribution over the same options produced by a judge, here the generative Qwen3.8-27B model at high reasoning effort, given the same input request.
The judge is the generative form of Max's own base model, not a stronger model, and it is not expected to be more accurate than Max.
It contributes a second alternate opinion, because it is given the same request as a text prompt and generates its answer token by token, and can perform long chain-of-thought reasoning before producing an answer, which a single-pass decision model cannot. The judge outputs a distribution for every record, including records that have no verifiable outcome.
The reward for producing the answer $a$ is
\begin{equation}
  r(a) \;=\; \rho\,\mathds{1}[a = y^\star] \;+\; \eta \log q^{J}_a,
  \label{eq:reward}
\end{equation}
with $\rho = 0$ when no verified outcome exists. RLAF maximizes
\begin{equation}
  J(\theta) \;=\; \E_x\Bigl[\, \sum_{a \in O} \pi_\theta(a \mid x)\, r(a) \;-\; \beta\, \KL\bigl(\pi_\theta(\cdot \mid x) \,\big\|\, \pi_0(\cdot \mid x)\bigr) \Bigr] \;-\; \lambda_\pi\, \E[\ell_\pi] \;-\; \lambda_s\, \E[\ell_s].
  \label{eq:sdrl}
\end{equation}
where $\ell_\pi$ is the permutation-consistency loss and $\ell_s$ is the state-shuffle consistency loss, both from \cref{eq:consistency}, and $\lambda_\pi$ and $\lambda_s$ are their weights.

The coefficient $\beta$ controls how far RLAF may move the model from the SFT model: a larger $\beta$ keeps the optimal policy closer to $\pi_0$, and a smaller $\beta$ increases the weight of the judge and the verified outcome.
The procedure here differs from the RLAF stage of the pointer models in two ways: it keeps the model close to the SFT model $\pi_0$ through the KL term, and the verified outcome and the judge enter as rewards rather than being combined directly into a target such as $\tilde q$.

\paragraph{When the judge overrides the verified answer.}
The optimal policy does not always have the verified answer as the most probable option, since the reward function, in \cref{eq:reward}, is the sum of two terms: the first computed from the verified answer and the second from the judge.
In theory, the contribution from the judge can override the contribution from the verified answer.
A judge that places sufficient probability on the wrong option will cause the optimal policy to likewise have the same wrong option as the most probable.
A larger $\rho$ or smoothing $q^J$ would prevent this.
We have not measured how often this occurs in our training.
For \drex Max we use $\rho = 0.2$ and $\eta = 0.05$ (\cref{tab:maxhp}).

Two consequences are testable but untested. The first is that if the judge's predictions align with the verified answer, it can effectively sharpen the optimal policy's distribution, leading to model overconfidence. This may partly explain the observed over-confidence of \drex Max (\cref{sec:calib-results}). The second is that the optimal policy may learn to imitate any bias from the judge.

\subsection{Calibration}
\label{sec:calibration}

We apply a temperature to the logits of every model before the final softmax operation (\cref{eq:pointer,eq:readout}).
We call the temperature value that ships with a model its \emph{deployed temperature}, $T_{\mathrm{dep}}$.
The released v1.5 and v2.0 pointer heads use $T_{\mathrm{dep}}{=}1$, i.e., no temperature scaling.
\drex Max uses $T_{\mathrm{dep}}{=}1.009$, fitted after the last training stage on held-out records from our training distribution by minimizing the negative log-likelihood~\citep{guo2017calibration}:
\begin{equation}
  T = \arg\min_{T > 0} \; \sum_{(z, y) \in \mathcal{D}} -\log \softmax(z / T)_y,
  \label{eq:temp}
\end{equation}
where $\mathcal{D}$ is a set of held-out decisions with each decision having logits $z$ and correct option $y$.
The temperature changes the confidence of every prediction but does not change which option is the most probable, since dividing all logits by the same $T > 0$ factor preserves their order.

In \cref{sec:calib-results}, we compare each model at its deployed temperature with the same model at a temperature $T^\star$ fitted by \cref{eq:temp} on half of the decisions in Decision Index and evaluated on the other half.
The difference in performance between the two resulting models reveals the effect of task-specific temperature tuning on the calibration error on the Decision Index. It also gives an indication into the disparity between the deployed temperature and the temperature value obtained from task-specific tuning.

\section{Experimental setup}
\label{sec:setup}
\paragraph{Decision Index.}
The Decision Index is a recognized public leaderboard consisting of a set of decision benchmarks, with each benchmark being categorized into one of 5 domains (i.e., areas)~\citep{decisionindex}.
For each benchmark, we convert the official metric (e.g., accuracy, macro-F1, nDCG, Brier skill, \ldots) into chance-corrected \emph{skill} metric, where 0 is uniform guessing and 100 is perfect, so that a binary task and a 150-option task are comparable (\cref{app:di}). In our results, we report the non-corrected metric as ``\emph{raw}'', and unanswered or unsupported requests count as wrong.
Versions of the Decision Index differ in their set of decision benchmarks and the weight of each benchmark in the set (\cref{app:evaldetails}).
The Decision Index validation set is partitioned into two subsets: a the public subset and a private subset, which is hidden.
The private subset is further subdivided into a \emph{same-skill} subset, which contains examples for tasks that are similar to those in the public subset, and a \emph{new-domain} subset, which contains examples for domains that do not exist in the public subset.
Two scores can be reported on Decision Index: (1) a \emph{public-only} score which is measured using examples from the subset, and (2) a \emph{full} score which is computed using a formula that depends on the model's performance on both the public and private subsets.
In our experiments, we compute the public scores and we submit our models to the maintainers of Decision Index to compute the full score.
Our evaluations of \drex Max, v2.0, v1.5, and GPT-6 Luna Decisions on Decision Index 0.4.1 each cover all 140{,}620 requests from the public subset (140{,}178 of which were not discarded by the Decision Index maintainers), where we record all per-request predicted probabilities.

\paragraph{Decision Bench v4.}
Decision Bench v4 is a benchmark with 1{,}071 single-question decisions across 35 tasks in 11 professional categories, sampled from 36 public datasets of real records~\citep{decisionbench2026}; 949 are text-only, 62 combine text and an image, and 60 are image questions.
Every question with an image also comes with a text rendering of the image (a description and, where applicable, text extracted via OCR).
We evaluate \drex Max twice: once with only the text rendering of each image, and once with the image itself provided to the model through its image encoder.
All other decision models receive the text rendering; GPT-6 Sol, Claude Opus 5.5, and d3-turbo, which is a commercial decision model, additionally receive the images whenever available.
Native probabilities are available for every decision model, so we report the expected calibration error (ECE) and the log loss for them; chat models return verbalized probabilities, which we do not score for calibration.

\paragraph{Other evaluations.}
We also evaluate on JevBench-231~\citep{standhartinger2026jevbench}, DecideBench v1.1~\citep{yong2026decidebench}, an OpenSpiel game arena~\citep{openspiel}, an OSWorld judging probe~\citep{xie2024osworld}, the Decision Index vision suites~\citep{decisionindex}, and long-context documents (\cref{app:evaldetails}).

\paragraph{Baselines.}
We compare against a handful of baseline systems.The decision-specialized systems are Jev 1.13.0~\citep{typesafe2026jev} and 1.14.0~\citep{typesafe2026jev114}, GPT-6 Luna Decisions~\citep{openai2026decisions}, d3-turbo~\citep{vllmsr2026d3turbo}, and other models with published entries on the Decision Index leaderboard.
We also compare against two frontier chat models, GPT-6 Sol and Claude Opus 5.5, which we prompt with lettered options and ask for JSON-formatted probabilities.
Leaderboard numbers are taken from the official leaderboards~\citep{decisionindex}; all other quantities reported in \cref{sec:results} come from our own experiments unless otherwise marked as reported. Whenever a model card reports a separate run on the same benchmark, we report only the accuracy that we measured.

For accuracy on a fixed set of questions, such as Decision Bench v4 and JevBench-231, we report 95\% Wilson confidence intervals~\citep{wilson1927}.
When we compare two models on the same questions, we use two methods: First, we compute a percentile bootstrap over questions (5{,}000 resamples) which gives a confidence interval on the difference in accuracy of the two models. Second, the exact two-sided McNemar test determines whether the two models are equally accurate, using only the questions on which exactly one of the two models answers correctly~\citep{mcnemar1947,dietterich1998tests,efron1993bootstrap}. On the Decision Index, a request can contain several decisions about the same state, and these decisions are correlated, so we resample whole requests rather than decisions.
For the confidence interval on the Decision Index score itself, we resample benchmarks within each domain. This interval shows how much the score would change if a different set of benchmarks had been chosen, rather than if a different set of questions were sampled within each benchmark.
We run many statistical tests in this paper and do not adjust their $p$-values for multiple comparisons (e.g., Bonferroni adjustments). An isolated $p$-value near 0.05 should therefore be interpreted as suggestive, not conclusive.

\section{Results}
\label{sec:results}
We first report accuracy on the Decision Index and on Decision Bench in \cref{sec:di032} and \cref{sec:dbench}, respectively. We then report calibration (\cref{sec:calib-results}) and stability under reordering of the options (\cref{sec:order}).
The remaining subsections contain results on multimodal decisions (\cref{sec:vision}), the contribution of each training stage (\cref{sec:dynamics}), and controlled comparisons on \drex Max (\cref{sec:maxablate}).

\subsection{Decision Index 0.4.1}
\label{sec:di032}

\begin{table}[t]
\centering
\begin{threeparttable}
\caption{\textbf{Decision Index 0.4.1} (leaderboard of 22 October 2026). \emph{Same}: subscore computed on the same-skill portion of the private subset; \emph{New}: subscore computed on the new-domain portion of the private subset; \emph{Raw}: the unweighted average accuracy on the public subset of 37 benchmarks, which is shared with version 0.3; \emph{Index}: the average accuracy on the public subset weighted as per weights provided by Decision Index; the last five columns give the chance-corrected skill on the public subset in each of the five domains of benchmarks (Knowledge, Language, Retrieval, Tools, Arts). Best per column in bold.}
\label{tab:di032}
\footnotesize
\setlength{\tabcolsep}{3.2pt}
\begin{tabular}{@{}lcrrrrrrrrrr@{}}
\toprule
 & & & \multicolumn{2}{c}{Private} & \multicolumn{2}{c}{Public} & \multicolumn{5}{c}{Public per-domain skill} \\
\cmidrule(lr){4-5}\cmidrule(lr){6-7}\cmidrule(lr){8-12}
Model & Active / total & Full & Same & New & Index & Raw & Know. & Lang. & Retr. & Tools & Arts \\
\midrule
\textbf{\drex Max} & 27B / 27B & \textbf{66.60} & \textbf{63.08} & \textbf{56.45} & \textbf{67.52} & \textbf{75.40} & \textbf{57.1} & \textbf{72.2} & \textbf{68.8} & \textbf{81.9} & 53.4 \\
\textbf{\drex v2.0} & 3B / 35B & 52.76 & 54.30 & 39.22 & 61.25 & 70.55 & 47.9 & 65.3 & 63.0 & 76.5 & \textbf{53.9} \\
\textbf{\drex v1.5} & 9B / 9B & 45.78 & 49.65 & 30.82 & 58.08 & 68.08 & 44.6 & 60.4 & 61.7 & 75.0 & 48.8 \\
\midrule
Jev 1.14.0 & N/A & 60.24 & 59.28 & 48.26 & 60.70 & 69.98 & \textbf{57.1} & 61.2 & 63.5 & 73.8 & 39.2 \\
Jev 1.13.0 & N/A & 59.22 & 58.04 & 47.72 & 57.96 & 68.55 & 53.9 & 59.2 & 55.4 & 75.1 & 39.1 \\
GPT-6 Luna Decisions$^{a}$ & N/A & --- & --- & --- & 47.78 & 61.15 & 32.2 & 51.9 & 49.5 & 68.0 & 37.1 \\
\bottomrule
\end{tabular}
\begin{tablenotes}\footnotesize
\item All rows except GPT-6 Luna Decisions are from the 0.4.1 leaderboard. All analyses in the below sections use the per-request outputs of our own evaluations of \drex Max, v2.0, and v1.5 on the public subset.
\item[$a$] Not on the leaderboard; the scores are from our own evaluation on the public subset, so the full score is not available.
\end{tablenotes}
\end{threeparttable}
\end{table}

\begin{figure}[t]
\centering
\includegraphics[width=\textwidth]{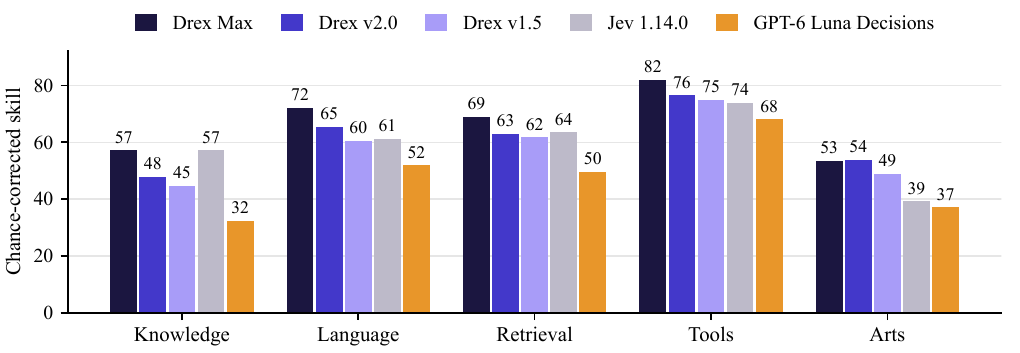}
\caption{\textbf{Per-domain skill on Decision Index 0.4.1} (public subset, scores from the leaderboard, except for GPT-6 Luna Decisions, which we evaluated locally). \drex Max leads in every domain except Arts, where \drex v2.0 is 0.5 points higher, and Knowledge, where it ties with Jev 1.14.0 at 57.1.}
\label{fig:di032areas}
\end{figure}

On the public subset, \drex Max scores 67.52 vs.\ 61.25 for v2.0, 58.08 for v1.5, 60.70 for Jev 1.14.0, and \DILuna{} for GPT-6 Luna Decisions (\cref{tab:di032}, \cref{fig:di032areas}).
There is a larger discrepancy in the full score, which includes results on private subset on the same skills and on new domains. The full score of Max (66.60) is 0.9 points lower than its public-only score (67.52), whereas the full score of v2.0 (52.76) is 8.5 points lower and that of v1.5 (45.78) is 12.3 points lower. On the new-domain tests in particular, Max scores 56.45 vs.\ 39.22 for v2.0 and 30.82 for v1.5 (\cref{tab:di032}).
Therefore, on the private tests, Max's lead holds, and over v2.0 widens. On the full score, Max is also 6.4 points ahead of Jev 1.14.0 (60.24), and they tie in skill on the public Knowledge domain (57.1).
On the leaderboard's public domains, Max's largest gains over v2.0 are in Knowledge (+9.2 skill), Language (+6.9), Retrieval (+5.8), and Tools (+5.4).
All per-benchmark results and confidence intervals below come from our own evaluations on the public subset. In these evaluations, the public-only score of Max is 5.5 points higher than that of v2.0, while the leaderboard reports a lead of 6.3 points (67.52 vs.\ 61.25). Our scores for Max differ from the leaderboard's by up to 0.8 points, which is our estimate of the variation between repeated evaluations (\cref{sec:limits}). Computing the percentile bootstrap confidence interval for this difference between the scores of Max and v2.0, we obtain a 95\% confidence interval of [\DIMaxVtwoLo, \DIMaxVtwoHi] points. Max also scores higher than v2.0 on \DIMaxWins{} of the \DIBenchN{} benchmarks, so the gain is not driven by a few benchmarks.
At the level of individual decisions, on the 230k \texttt{choice} decisions with gold labels that both models answered in these experiments, Max is more accurate by \PairedMaxVtwo{} percentage points (95\% confidence interval [\PairedMaxVtwoLo, \PairedMaxVtwoHi]).
The largest per-benchmark gains in our experiments are on GSM8K ($66.8{\to}83.2$ accuracy; \citealp{rv_gsm8k}), BBH ($78.3{\to}93.7$; \citealp{suzgun2023bbh}), ACOS ($23.9{\to}39.8$ F1; \citealp{rv_acos}), WinoGrande ($86.1{\to}94.8$; \citealp{rv_winogrande}), MuSR ($59.6{\to}70.3$; \citealp{rv_musr}), and the home-appliance simulator ($54.5{\to}70.5$) (\cref{fig:deltamax} in the Appendix, \cref{app:bench032}).

On two benchmarks, Max scores are substantially lower than v2.0.
On the POP909-CL benchmark, a next-chord prediction task from symbolic music, the score falls from 79.5 to 55.6, and on the Habermas Machine benchmark, a preference-aggregation task, from 62.4 to 50.5; both are Arts benchmarks that the pointer models handle well and that depend on reading long structured sequences, whereas the fact that the readout head must bind each option code to a long option may hurt its performance.
The scores on When2Call ($86.5{\to}85.2$) and VAST ($78.4{\to}75.6$) also decline slightly.
The gap in performance between Max and frontier chat models on knowledge-heavy benchmarks is not closed: on a sample of the same 100 requests per benchmark (\cref{tab:frontier} in the Appendix), Max correctly answers 21\% of HLE~\citep{rv_hle} and 55\% of GPQA Diamond~\citep{rein2024gpqa}, vs.\ 39\% and 90\% for GPT-6 Sol and 56\% and 89\% for Claude Opus 5.5, respectively.

\subsection{Decision Bench v4}
\label{sec:dbench}

\begin{table}[t]
\centering
\begin{threeparttable}
\caption{\textbf{Decision Bench v4} (1{,}071 decisions; consisting of 949 text, 62 text+image, and 60 image examples). Accuracy with 95\% Wilson confidence intervals; expected calibration error (ECE) and log loss from the model's probabilities; median end-to-end wall time per request from our client. \drex Max (text rendering) receives only the text rendering of each image.}
\label{tab:dbench}
\small
\setlength{\tabcolsep}{2.6pt}
\begin{tabular}{@{}lccccccc@{}}
\toprule
 & & \multicolumn{3}{c}{Accuracy by modality (\%)} & & & \\
\cmidrule(lr){3-5}
Model & Accuracy (\%) & Text & Text+Img & Image & ECE (\%) & Log loss & \shortstack{Median time\\per request (s)} \\
\midrule
\textbf{\drex Max} (image encoder) & 94.0 [92.4, 95.3] & 93.4 & 100.0 & 98.3 & 1.40 & 0.159 & 0.66 \\
\drex Max (text rendering) & 92.1 [90.3, 93.5] & 93.4 & 100.0 & 63.3 & 1.51 & 0.194 & 0.66 \\
\midrule
\drex v2.0$^{\dagger}$ & 87.9 [85.8, 89.7] & 88.3 & 100.0 & 68.3 & 1.73 & 0.287 & 0.38 \\
\drex v1.5$^{\dagger}$ & 86.8 [84.7, 88.7] & 86.9 & 98.4 & 73.3 & 4.77 & 0.339 & 0.26 \\
Jev 1.14$^{\dagger}$ & 92.3 [90.5, 93.7] & 93.4 & 100.0 & 66.7 & 1.60 & 0.212 & 0.51 \\
\midrule
d3-turbo & 93.7 [92.1, 95.0] & 93.0 & 100.0 & 98.3 & 1.18 & 0.166 & 0.72 \\
GPT-6 Sol & 94.7 [93.2, 95.9] & 94.0 & 100.0 & 100.0 & --- & --- & 2.05 \\
Claude Opus 5.5 & 95.6 [94.2, 96.7] & 95.0 & 100.0 & 100.0 & --- & --- & 4.33 \\
\bottomrule
\end{tabular}
\begin{tablenotes}\footnotesize
\item[$\dagger$] Text-only model; image questions are answered from the text rendering. The images are provided to d3-turbo, GPT-6 Sol and Claude Opus 5.5 via their image encoders.
\end{tablenotes}
\end{threeparttable}
\end{table}

\paragraph{Image inputs.}
On the 60 image-only questions, every text-only model receives only a caption and text extracted via OCR, and accuracy falls to 63--73\% (\cref{tab:dbench}).
Giving the same questions to \drex Max through its image encoder raises its image-only accuracy from \DBMaxTxtImg\% to \DBMaxPixImg\% and its overall accuracy from \DBMaxTxt\% to \DBMaxPix\% (\DBMaxPixK{} of the 1{,}071 questions answered correctly). On the 62 questions that combine text and an image, every model answers 98.4--100\% of the questions correctly, since the text alone is usually sufficient to determine the correct answer.
This gain comes from the image-encoder stage (\cref{sec:maxtrain}).

\paragraph{Comparison with other models.}
When the images are given to \drex Max through its image encoder, its accuracy is statistically indistinguishable from that of d3-turbo and GPT-6 Sol, marginally higher than that of Jev 1.14 ($p{=}0.048$ in an exact McNemar test, a difference that comes from the image questions, since the two models are tied on text-only questions), and marginally lower than that of Claude Opus 5.5 ($p{=}0.046$, see \cref{tab:dbenchpaired} in the Appendix).
\drex Max is clearly ahead of v2.0 and v1.5, by 6.2 and 7.2 points overall, and by 5.1 and 6.4 points on text questions alone, respectively.
Its ECE (1.40\%) is the second lowest among the decision models, after d3-turbo's (1.18\%), and it produces answers in a median of 0.66\,s end-to-end vs.\ 2.1\,s for GPT-6 Sol and 4.3\,s for Claude Opus 5.5.
By category (see \cref{fig:dbcats} in the Appendix, \cref{app:dbench}), Max is weakest on \textsc{Product management}, \textsc{Documents \& meetings}, and \textsc{AI agents \& evals}. \textsc{Product management} and \textsc{Documents \& meetings} are also among the three hardest categories for every other model. Max also trails on \textsc{Trust \& safety}, where it answers 91.7\% correctly vs.\ 95.0--98.3\% for Jev 1.14, d3-turbo, and the frontier models. On the \textsc{Design} category, whose decisions are mostly about images, the accuracy of Max improves, largely due to the image encoder: 96.7\% vs.\ 26.7\% from the text rendering and 33.3--46.7\% for v2.0, v1.5, and Jev 1.14 (\cref{tab:dbcat} in the Appendix).

\begin{figure}[t]
\centering
\resizebox{\textwidth}{!}{\input{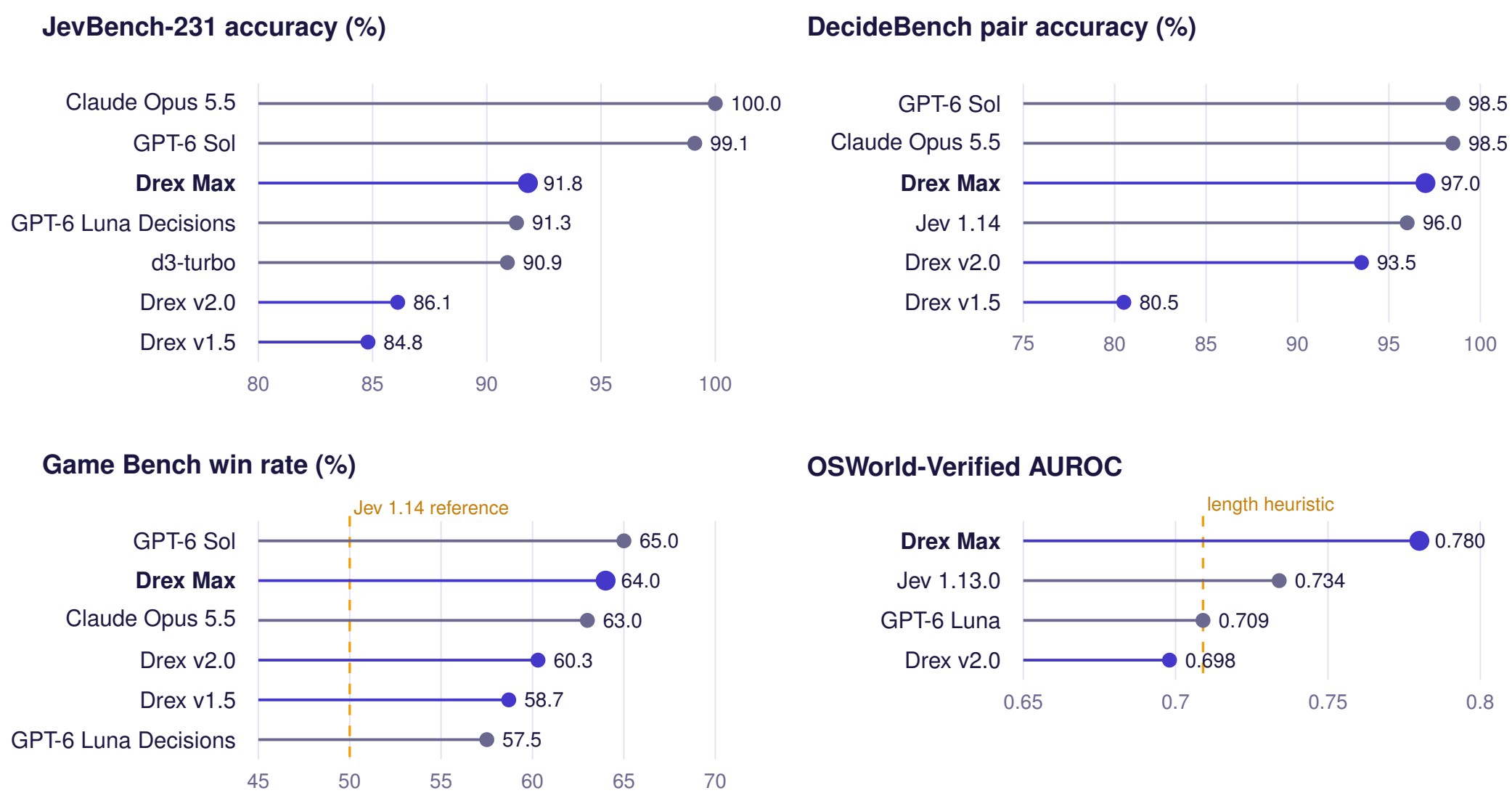}}
\caption{\textbf{Other decision benchmarks.} One dot per model; \drex models are colored \textcolor{drexaccent}{purple}, and axes do not start at zero. The dashed line indicates the Jev 1.14 reference win rate of 50\% in the game arena, and it indicates the ``shorter transcript $\Rightarrow$ success'' heuristic (AUROC 0.709) on OSWorld-Verified. The game-arena win rates (bottom-left panel) come from a separate evaluation that was run when \drex Max was launched (\cref{tab:gamebench} in the Appendix) and have no confidence intervals. The results in the other three panels come from our own evaluations.}
\label{fig:otherbench}
\end{figure}

\paragraph{Other decision benchmarks.}
\drex Max is highly competitive with the strongest decision models on JevBench-231 and outperforms v2.0 and v1.5 ($p<0.01$). It reaches the lowest ECE on DecideBench (1.1\%), has the highest win rate among decision models in the launch game arena, and is the only model that clearly outperforms a transcript-length heuristic when judging finished OSWorld episodes, with the lowest Brier score (\cref{fig:otherbench}; \cref{sec:otherbench,sec:agentic}).

\subsection{Artificial Analysis Decision Index}
\label{sec:aa}

Artificial Analysis maintains its own decision model leaderboard called the Artificial Analysis Decision Index~\citep{artificialanalysis}, where the overall score is the computed as the average of four evaluations with equal weight:
\begin{itemize}
  \item \textbf{AA-Omniscience-Decision}: 500 private questions with five options each. Its score is the average of the accuracy and the \emph{non-hallucination rate}, which is defined as the number of abstentions\footnote{An abstention is when a model refuses to provide an answer.} divided by the number of incorrect responses.
  \item \textbf{HLE-MultiChoice}: 513 text-only multiple-choice questions from Humanity's Last Exam~\citep{rv_hle}, scored by accuracy.
  \item \textbf{When2Call}: 3,652 text-only questions that test whether a model should call a tool. Each question has four options: answer directly, call a tool, ask a follow-up question, or state that it cannot answer. The direct answer is never correct. Scored by accuracy.
  \item \textbf{WildGuardTest}: 1,725 text-only items from user--LLM conversations, each with three binary labels: whether the prompt is harmful, whether the response is harmful, and whether the response is a refusal. Scored by the F1 score averaged over the three tasks.
\end{itemize}
The \drex models were added to the leaderboard on 29 October 2026, and we did not run any of these evaluations (\cref{tab:aa}).

\drex Max scores 63.7 on the Artificial Analysis Decision Index~\citep{aa2026decisionindex}, 1.0--1.6 points behind d1-ultra~\citep{liquidAI2026d1-ultra}, SubQ Snap 1.1, and Elephant 9B~\citep{baseten2026mammoth}, which score 65.3, 65.0, and 64.7, respectively.

Artificial Analysis reports no confidence intervals, so we do not rank \drex Max against these three models, because the difference in score is at most 1.6 points.
\drex Max is 6.6 points above Jev 1.14, 7.5 above Decider v1.2, and 13.0 above \drex v1.5.
More than half of the 13.0-point gain of \drex Max over v1.5 comes from its higher non-hallucination rate on AA-Omniscience-Decision. On this evaluation, the non-hallucination rate rises from 0.6\% to 71.3\%, while its accuracy falls from 37.6\% to 27.6\%. Its improved performance on When2Call (84.8 to 94.0) and WildGuardTest (86.9 to 95.5) each raise its overall score by about two points.
\drex Max achieves the highest WildGuardTest F1 of any model listed (95.5) and a When2Call accuracy above that of every language model, but its accuracy is far below that of the best language models on HLE-MultiChoice (16.0 vs.\ 70.2 for Claude Opus 5.5) and on AA-Omniscience-Decision, where the accuracy (27.6 vs.\ 93.4) makes up half of the score. Thus, the language models score higher overall (79.0 for Claude Opus 5.5 and 71.5 for GPT-6.1 Sol at their best settings) at 44 to 1{,}100 times the cost per task.

\begin{table}[t]
\centering
\footnotesize
\setlength{\tabcolsep}{3pt}
\newcolumntype{R}[1]{>{\raggedleft\arraybackslash}b{#1}}
\begin{tabular}{@{}>{\raggedright\arraybackslash}b{2.70cm} R{0.8cm} R{0.8cm} R{1.55cm} R{1.1cm} R{1.25cm} R{1.25cm} R{1.4cm} R{1.40cm}@{}}
\toprule
 & & \multicolumn{2}{c}{AA-Omniscience} & & & & & \\
\cmidrule(lr){3-4}
Model & Index & Accuracy & Non-halluc\-ination rate & \mbox{HLE-MC} & When2Call & WildGuard & Cost per 1M requests & Time per request (ms) \\
\midrule
Claude Opus 5.5 (Max) & 79.0 & 93.4 & 30.3 & 70.2 & 90.5 & 93.5 & \$51{,}295 & 3571 \\
GPT-6.1 Sol (Low) & 71.5 & 93.0 & 22.9 & 55.4 & 80.6 & 92.0 & \$2{,}048 & 2888 \\
\midrule
d1-ultra & 65.3 & 31.6 & 84.5 & 15.0 & 92.6 & 95.4 & \$32 & 93 \\
SubQ Snap 1.1 & 65.0 & 20.2 & 85.7 & 17.5 & 94.6 & 94.9 & \$30 & 190 \\
Elephant 9B & 64.7 & 11.6 & 93.2 & 18.3 & 94.8 & 93.3 & \$8 & 165 \\
\drex Max & 63.7 & 27.6 & 71.3 & 16.0 & 94.0 & 95.5 & \$46 & 353 \\
\drex v2.0 & 57.9 & 27.4 & 60.3 & 13.6 & 86.2 & 87.9 & \$41 & 252 \\
Jev 1.14 & 57.1 & 33.8 & 37.2 & 18.9 & 80.8 & 93.1 & \$42 & 78 \\
Decider v1.2 & 56.2 & 39.4 & 16.5 & 14.6 & 87.7 & 94.4 & \$15 & 206 \\
\drex v1.5 & 50.7 & 37.6 & 0.6 & 12.1 & 84.8 & 86.9 & \$35 & 186 \\
GPT-6 Luna (Decisions) & 48.5 & 40.6 & 10.1 & 11.7 & 74.1 & 82.8 & \$79 & 168 \\
\bottomrule
\end{tabular}

\caption{\textbf{Artificial Analysis Decision Index} (retrieved 29 October 2026). The table lists two language models at their best setting, the three leading decision models, the \drex models, and the Jev, Decider and GPT-6 Luna Decisions models, out of a total of 35 entries on the leaderboard (an entry is one model with one setting, so some language models appear more than once). The columns under AA-Omniscience list the accuracy and non-hallucination rate (share of wrong answers that were abstentions) of each model for AA-Omniscience-Decision; the numbers in the HLE-MC and When2Call columns are accuracies, and those under WildGuard are average F1 scores on WildGuardTest. Cost is in US dollars per million requests and time is the average per request, both as reported by Artificial Analysis.}
\label{tab:aa}
\end{table}

\subsection{Calibration results}
\label{sec:calib-results}

\paragraph{Protocol.}
To measure the calibration of each model, we evaluate the model on Decision Index 0.4.1 and use every \texttt{choice} decision with gold labels, which yields about 230k decisions in 108k requests per model (retrieval decisions scored by nDCG have no single gold option and are excluded).
We additionally aim to compare the effect of task-specific temperature tuning on calibration.
To do so, the requests are partitioned into two equal subsets uniformly at random with a fixed seed; the temperature $T^\star$ is fitted with \cref{eq:temp} on one subset and we evaluate calibration on the other, so each model is evaluated on about 115k decisions.
Each model is evaluated at its deployed temperature $T_{\mathrm{dep}}$ (\cref{sec:calibration}) and at $T^\star$.
ECE is defined in \cref{eq:ece} in the Appendix; intervals are 95\% percentile bootstraps over requests.

\begin{table}[t]
\centering
\caption{\textbf{Calibration on Decision Index 0.4.1 decisions} (held-out half, $\approx$115k decisions per model). Accuracy does not depend on the temperature. \emph{Deployed temperature}: the model as released ($T_{\mathrm{dep}}$, \cref{sec:calibration}); \emph{Decision Index-tuned temperature}: temperature $T^\star$ fitted on the other half of Decision Index. Mean confidence is the average probability of the top option. ECE is expected calibration error, and NLL is negative log loss. Lower ECE, NLL, and Brier score are better. The last two columns give the accuracy on the $k\%$ of decisions on which the model is most confident.}
\label{tab:calib}
\small
\setlength{\tabcolsep}{3.0pt}
\resizebox{\textwidth}{!}{%
\begin{tabular}{@{}lccccccccc@{}}
\toprule
 & & \multicolumn{3}{c}{Deployed temperature} & \multicolumn{3}{c}{Decision Index-tuned temperature} & \multicolumn{2}{c}{Accuracy on top-$k$\% (\%)} \\
\cmidrule(lr){3-5}\cmidrule(lr){6-8}\cmidrule(lr){9-10}
Model & Accuracy (\%) & ECE (\%) [95\% CI] & NLL & Brier & $T^\star$ & ECE (\%) & NLL & $k{=}80$ & $k{=}50$ \\
\midrule
\drex Max & 87.9 & 3.61 [3.42, 3.83] & 0.360 & 0.172 & 1.35 & 1.35 & 0.345 & 96.1 & 99.3 \\
\drex v2.0 & 86.0 & 1.37 [1.20, 1.56] & 0.382 & 0.194 & 0.91 & 0.79 & 0.381 & 94.4 & 98.9 \\
\drex v1.5 & 85.3 & 4.54 [4.27, 4.77] & 0.429 & 0.207 & 1.48 & 1.17 & 0.398 & 94.0 & 99.3 \\
GPT-6 Luna Decisions & 76.4 & 4.87 [4.65, 5.15] & 0.760 & 0.326 & 1.91 & 6.57 & 0.679 & 85.1 & 91.8 \\
\bottomrule
\end{tabular}}
\end{table}

\begin{figure}[t]
\centering
\includegraphics[width=\textwidth]{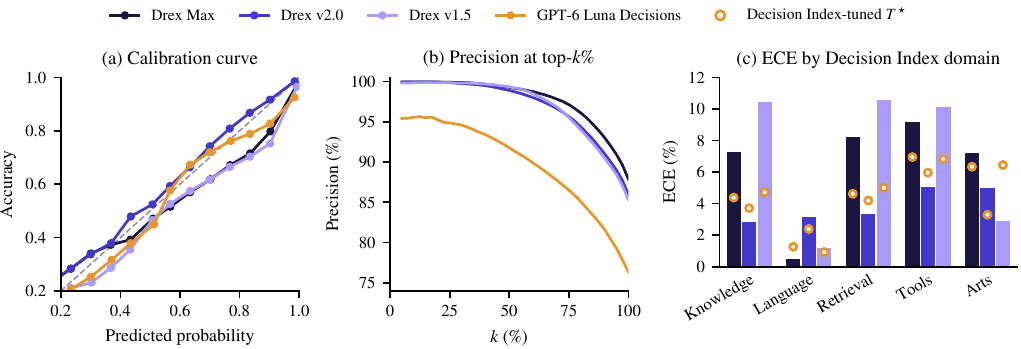}
\caption{\textbf{Calibration of the deployed models} on the evaluation subset of Decision Index 0.4.1. \textbf{(a)} Calibration curves: the accuracy of the decisions in each bin (of predicted probabilities) versus the average predicted probability of the decisions that bin, showing only bins with at least 30 decisions; a perfectly calibrated model lies on the diagonal. \textbf{(b)} Precision@top-$k$\%: the accuracy on the $k$\% of decisions on which the model is most confident, for $k = 25, 50, 75, 100$. \textbf{(c)} Expected calibration error (ECE) in each Decision Index domain for the three \drex models, at the deployed temperature (bars) and at the temperature $T^\star$ fitted on a subset of Decision Index (circles). GPT-6 Luna Decisions is not shown in (c) because we do not have its per-domain breakdown.}
\label{fig:calibration}
\end{figure}

\paragraph{Calibration by model.}
Among the deployed models, \drex v2.0 is the best calibrated, with an ECE of \CalTwoECE\% at \CalTwoAcc\% accuracy.
Its mean confidence is slightly below its accuracy, so the model is slightly under-confident.
Fitting the temperature on a subset of Decision Index gives $T^\star = \CalTwoT$; because $T^\star < 1$, dividing the logits by $T^\star$ results in sharpened output distributions (i.e., more peaked), which raises the confidence of the model and reduces its ECE on the evaluation subset to \CalTwoECET\% (\cref{tab:calib}, \cref{fig:calibration}a).
An earlier measurement on Decision Index 0.2.1, which contains a different set of benchmarks, gave a similarly low ECE for v2.0 of \CalVtwoECE\% [\CalVtwoECELo, \CalVtwoECEHi] on \CalVtwoN{} decisions, with the same tuned temperature $T^\star = \CalVtwoT$.

\drex Max is the most accurate model (\CalMaxAcc\%) and is mildly over-confident: at its deployed temperature, its mean confidence is \CalMaxConf\% and its ECE is \CalMaxECE\%.
Fitting the temperature on the Decision Index subset gives $T^\star = \CalMaxT$; because $T^\star > 1$, dividing the logits by $T^\star$ makes the output distributions flatter (closer to uniform), which lowers the confidence of the model and reduces its ECE on the evaluation subset to \CalMaxECET\% and its NLL from 0.360 to 0.345.

\drex v1.5 is over-confident by a similar margin (ECE is \CalOneECE\%), the tuned temperature is $T^\star{=}\CalOneT$ which causes the ECE to decrease to \CalOneECET\%.
GPT-6 Luna Decisions is the least well-calibrated model, with an ECE of \CalLunaECE\%.
Using the Decision Index-tuned temperature, $T^\star = \CalLunaT$, reduces its NLL from 0.760 to 0.679 but \emph{increases} its ECE on the evaluation subset to \CalLunaECET\%.
Because $T^\star$ is tuned to minimize NLL rather than ECE, and a single temperature rescales the confidence of every decision in the same direction, it cannot correct a model that is over-confident on some decisions and under-confident on others.
The ranking of models by accuracy (Max $>$ v2.0 $>$ v1.5 $>$ Luna) is identical to the rankings by NLL and by Brier score, as NLL and Brier score are proper scoring rules, which reward both accuracy and calibration.
\drex Max's deployed temperature, $T{=}1.009$, is close to one; the temperature value fitted on the subset of Decision Index $T^\star = \CalMaxT$ suggests that its calibration set differs from the Decision Index distribution, and that a deployment in a new environment would benefit from task-specific temperature tuning.

\paragraph{Calibration by domain and option count.}
Per-domain and per-option-count breakdowns (\cref{tab:calibarea,tab:calibk} in the Appendix) show that the ECE computed over all decisions understates the miscalibration of each \drex model within individual domains.
For \drex Max and v1.5, the Language domain, which contains 56\% of the decisions, is almost perfectly calibrated (ECE of 0.5\% and 1.2\%), so it lowers the overall ECE even though the other four domains have ECE of 7.2--9.2\% for Max and 2.9--10.6\% for v1.5.
For \drex v2.0, every domain has a higher ECE (2.8--5.0\%) than the overall ECE of 1.37\%, because v2.0 is under-confident in the Language domain and over-confident in the other domains, and these deviations of opposite direction partly cancel when all decisions are aggregated.
Max's mean confidence exceeds its accuracy by 5--9 percentage points in Knowledge, Retrieval, Tools, and Arts (per-domain ECE 7.2--9.2\%), and the Decision Index-tuned temperature reduces these ECE values to 4.4--7.0\% while Max remains over-confident; v2.0's per-domain ECE is 2.8--5.0\% at its deployed temperature.
We also measure calibration with respect to the number of options in the input. We find that Max is over-confident on questions with 6--20 options (ECE 8.4\% at 71.7\% accuracy), where v2.0 is well-calibrated (ECE 2.0\%).
A single temperature fitted on one distribution or domain of decisions therefore does not guarantee good calibration on another. Thus, we recommend that, for any application of \drex Max that contains a component that depends on whether Max's predicted probability exceeds a threshold, and that serves a narrow type of request, the temperature parameter should be tuned on labeled decisions of that type.

Because every output is a normalized distribution over the supplied options, the model's predicted probability in each decision can be read directly from its output. An application can therefore flag the low-confidence decisions of a decision model for human review without training a separate model to predict which decisions are likely to be wrong. The request together with the predicted distribution serves as the audit record of each decision.
Keeping only the 80\% of decisions on which \drex Max is most confident raises its precision from \CalMaxAcc\% (precision@top-100\%, i.e., accuracy on all decisions) to \CalMaxSelEighty\% (precision@top-80\%), and keeping the most confident 50\% raises it further to \CalMaxSelHalf\% (precision@top-50\%; \cref{fig:calibration}b). The corresponding precisions for v2.0 are \CalTwoSelEighty\% and \CalTwoSelHalf\%.

\subsection{Option-order robustness}
\label{sec:order}

\paragraph{Protocol.}
We drew a sample of requests from the 37 benchmarks of the public index of Decision Index 0.4.1, stratified by benchmark: we sampled from each benchmark separately, so that no benchmark contributes more than 40 requests.
From each benchmark, we kept only the requests in which every question is a \texttt{choice} question with at least one gold label that is a string, the state has at most 12{,}000 characters, and there are at most 16 questions.
We then shuffled these requests with a fixed random seed and took the first 40 (or all of them, if the benchmark has fewer), which gives 1{,}175 requests (\cref{app:order}).
We gave each request to four models, \drex Max, \drex v2.0, \drex v1.5, and Jev 1.14.0, through public APIs.
Each request was given three times: with the original ordering of the options, with the reversed ordering, and with a random permutation of the options of every question.
Of the 1{,}175 sampled requests, \OrdReq{} were answered by all four models under all three orderings.
We excluded the remaining requests for every model, for one of two reasons.
First, in some requests, the options are structured objects with several fields rather than strings (for example, a musical chord or a preference profile), and the public APIs reject such requests because they fail the schema validation of the API.
The evaluation code of the Decision Index converts these options into strings before it passes them to a model, but we gave the requests to the APIs unchanged.
Second, some requests were rejected by the Jev API.
This leaves \OrdN{} decisions (\OrdNTwo{} binary, \OrdNMid{} with 3--5 options, \OrdNSix{} with six or more).
Before we compare the outputs of a model under two orderings, we map the probabilities that it returns for a reordered request back to the original ordering of the options.
\paragraph{Metrics.}
We say a decision \emph{flips} under a reordering of its options if its most probable option under that reordering differs from its most probable option under the original ordering.
The \emph{flip rate} of a model is the fraction of its decisions that flip, $\Pr[\arg\max p \neq \arg\max p^{\pi}]$, where $p$ and $p^{\pi}$ are, respectively, the probability distributions that the model returns under the original ordering and under the reordering $\pi$, after mapping $p^{\pi}$ back to the original ordering (\cref{app:metrics}).
We also report the accuracy of each model under each ordering, and the mean total variation distance $\TV(p, p^{\pi}) = \tfrac12 \lVert p - p^{\pi} \rVert_1$, which measures how much the whole distribution changes, and not only the most probable option.
\begin{table}[t]
\centering
\caption{\textbf{Option-order robustness} on \OrdN{} decisions. The table gives the accuracy (\%) of each model under the original ordering of the options, under the reversed ordering, and under a random permutation. It also gives the accuracy of an ensemble over the three orderings: for each decision, we average the three probability distributions that the model returns under the three orderings, option by option, and take the option with the highest average probability. This accuracy is therefore not the mean of the three other accuracies. The flip rate (\%) is the fraction of decisions whose top option differs from the top option under the original ordering, with a 95\% bootstrap confidence interval for the random permutation. ``Unstable'' is the fraction of decisions whose top option is not the same under all three orderings. The mean total variation distance is computed between the distributions under the original ordering and under the random permutation. $K$ is the number of options of a decision.}
\label{tab:order}
\small
\setlength{\tabcolsep}{2.6pt}
\resizebox{\textwidth}{!}{%
\begin{tabular}{@{}l cccc ccc c ccc@{}}
\toprule
 & \multicolumn{4}{c}{Accuracy (\%)} & \multicolumn{3}{c}{Top-1 flip rate (\%)} & & \multicolumn{3}{c}{Flip under permutation by $K$ (\%)} \\
\cmidrule(lr){2-5}\cmidrule(lr){6-8}\cmidrule(lr){10-12}
Model & \begin{tabular}[b]{@{}c@{}}Original\\ordering\end{tabular} & \begin{tabular}[b]{@{}c@{}}Reversed\\ordering\end{tabular} & \begin{tabular}[b]{@{}c@{}}Permuted\\ordering\end{tabular} & \begin{tabular}[b]{@{}c@{}}Ensemble of\\3 orderings\end{tabular} & \begin{tabular}[b]{@{}c@{}}Reversed\\ordering\end{tabular} & \begin{tabular}[b]{@{}c@{}}Permuted ordering\\{}[95\% CI]\end{tabular} & Unstable & \begin{tabular}[b]{@{}c@{}}Mean total\\variation\\distance\end{tabular} & $K{=}2$ & \hspace{1.0em}$3{\le}K{\le}5$ & $K\ge6$ \\
\midrule
\drex Max & 83.6 & 83.3 & 83.4 & 83.5 & 6.2 & 4.3 [3.3, 5.3] & 7.4 & 0.037 & 1.0 & 7.2 & 13.3 \\
\drex v2.0 & 81.0 & 79.8 & 80.6 & 81.1 & 8.0 & 5.9 [4.8, 7.1] & 9.8 & 0.050 & 2.0 & 9.5 & 16.2 \\
\drex v1.5 & 78.8 & 78.3 & 79.6 & 79.6 & 8.4 & 5.3 [4.2, 6.5] & 9.2 & 0.047 & 2.0 & 7.8 & 15.6 \\
Jev 1.14 & 79.3 & 78.9 & 79.0 & 79.5 & 5.7 & 4.3 [3.3, 5.4] & 7.0 & 0.037 & 2.0 & 5.2 & 14.5 \\
\bottomrule
\end{tabular}}
\end{table}

\begin{figure}[t]
\centering
\includegraphics[width=\textwidth]{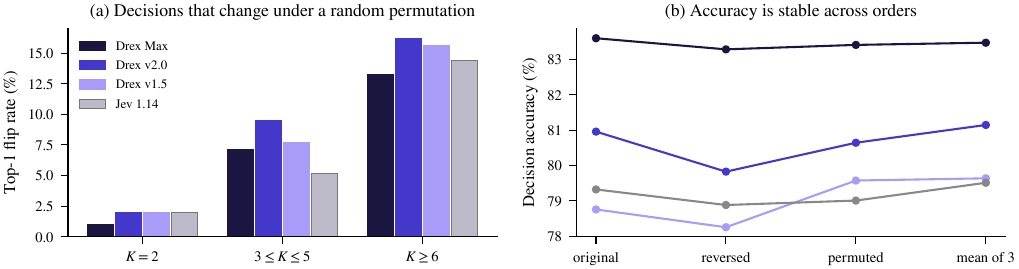}
\caption{\textbf{Option order.} \textbf{(a)} The flip rate under a random permutation of the options, that is, the fraction of decisions whose top option changes, for decisions with $K{=}2$, $3 \le K \le 5$, and $K \ge 6$ options. \textbf{(b)} The accuracy of each model under the original, reversed, and permuted orderings of the options, and the accuracy of the ensemble of the three orderings, which averages the three probability distributions that the model returns for a decision, option by option, and takes the option with the highest average probability (\cref{tab:order}). Under the reversed and permuted orderings and the ensemble, the accuracy of every model differs from its accuracy under the original ordering by at most 1.2 percentage points.}
\label{fig:order}
\end{figure}
\paragraph{Flip rate and accuracy.}
We find that random permutations of the option set change the most probable option in \OrdMaxFlip\% of \drex Max's decisions, \OrdVtwoFlip\% of v2.0's, \OrdVoneFlip\% of v1.5's, and \OrdJevFlip\% of Jev's (\cref{tab:order}, \cref{fig:order}). Reversing the option set, which moves every option, has a greater effect on the output of every model (5.7--8.4\%).
Despite its readout head, whose logits are tied to positions (\cref{sec:readout}), \drex Max is as stable as Jev and more stable than the pointer models, whose hybrid backbones cannot isolate options.
Accuracy, however, changes little; Max scores \OrdMaxAccO\% under the original ordering and \OrdMaxAccP\% under the permutation, and the flips are nearly balanced between gains and losses.

\paragraph{Order instability and confidence.}
Decisions for which the most probable option changes across the three orderings are far less accurate than those that do not: 37.6\% vs.\ 87.2\% for \drex Max, and 37.7--46.8\% vs.\ 81.8--84.7\% for the other models (\cref{tab:ordertta} in the Appendix).
Order instability is thus a strong signal that a decision is wrong.
Since \drex can be used in applications where it should produce abstentions for some decisions, we compare two rules for determining \emph{when} to abstain.
The first rule abstains on the decisions that are unstable under reordering (i.e., decisions for which the most probable option is different under each of the three orderings), and the second rule abstains on the $n$ decisions with the lowest confidence where $n$ is the number of abstentions produced by the first rule.
Both rules yield almost the same accuracy on the non-abstention decisions: 87.2\% vs.\ 87.1\% for Max, and for every model, the two accuracies differ by at most 0.4 percentage points (\cref{tab:ordertta} in the Appendix).
We also define the \emph{coverage} of a rule as the fraction of decisions on which it does not abstain.
At a coverage of 80--85\%, abstaining on the decisions that are either unstable under reordering or have low confidence raises accuracy by at most 0.6 percentage points over abstaining on low confidence alone.
In this sample, the decisions that are changed by option reordering are largely those to which the model already assigns low confidence, so a user who scrutinizes the model's output confidence gains little from further scrutiny of the model's sensitivity to option reordering.
This observation is an association, not a causal account of every flip, and the set of decisions with six or more options, which has the highest flip rate, contains only \OrdNSix{} decisions. Further experiments to replicate this result on larger, independent workloads would be valuable.
We also consider an ensemble of orderings, which averages the three probability distributions that the model returns for a decision, option by option, and predicts the option with the highest average probability.
The ensemble of the three orderings (\cref{tab:order}) changes accuracy by at most 0.8 percentage points but lowers log loss for every model, and significantly so for Max (a reduction of 0.014 nats, 95\% confidence interval $[0.005, 0.024]$) and Jev (a reduction of 0.044 nats, 95\% confidence interval $[0.018, 0.075]$), and lowers ECE for three of the four models.

\paragraph{Option count.}
Binary decisions almost never flip (\OrdMaxFlipTwo\% for Max, 2.0\% for the others), but with six or more options, 13--16\% of decisions change under a permutation for every model, including Jev.
These are the decisions to which the model already assigns low confidence (\cref{tab:calibk} in the Appendix), so most flips change the top option among options with nearly equal probabilities.
The practical consequence for a user is the same as for the small numerical differences in the output probabilities that arise at inference time when a request is processed in a different batch (\cref{sec:servinggate}).
If the confidence of a decision on a question with many options is close to some behaviorally-significant threshold, the decision may change, and this is already indicated by its low confidence.
Order robustness is learned in every model we tested.
Option isolation (\cref{sec:attn}) is an attention mask that would make the pointer head robust to option order by construction, but it can only be applied to attention-only backbone models.
The backbones of \drex v1.5, v2.0, and Max are hybrids with recurrent Gated DeltaNet layers, which cannot apply such a mask, so we have not tested option isolation on any of our released models.

\subsection{Vision benchmarks}
\label{sec:vision}

On the 11 public vision benchmarks of the Decision Index (\cref{tab:vision} in the Appendix), \drex Max has its highest accuracy on documents (InfographicVQA 98.8\%, key-information extraction from receipts and forms 95.5\%, and questions about charts in CharXiv 86.1\%) and on selecting the correct element of a web page for a web agent (Mind2Web 88.6\%).
It has its lowest accuracy on benchmarks whose images require multi-step reasoning or cultural context: MMMU-Pro (53.6\%), R-Bench-M (48.9\%), and Hateful Memes (70.6\%, which corresponds to a chance-corrected skill of only 41.1, since Hateful Memes is a binary task).
We also compare Max with GPT-6 Luna Decisions on the same sample of 80 items from each benchmark for both models.
Max is more accurate than Luna on all 11 benchmarks, and its normalized accuracy of the Decision Index vision leaderboard is 13.2 points higher (77.1 vs.\ 63.9), computed on this sample.
On Decision Bench, Max answers 93.4\% of the text-only questions and 98.3\% of the 60 image-only questions correctly when it receives the images via its image encoder (\cref{sec:dbench}).
Together with the vision benchmarks, this suggests that the accuracy of Max on image inputs is close to its accuracy on text inputs for extraction and grounding, and lower for visual reasoning.
However, these comparisons are made across different benchmarks, not between the same questions given once as images and once as text.

\subsection{Evaluation of training checkpoints}
\label{sec:dynamics}

\begin{table}[t]
\centering
\caption{\textbf{Evaluation of \drex v2.0 training checkpoints.} Each row gives the results of one saved checkpoint of \drex v2.0, evaluated on all benchmarks of Decision Index 0.2.1, on JevBench-231, and in the game arena. For the Decision Index, we use its official evaluation code and report the chance-corrected skill, both overall and in each of its five domains. For JevBench-231, we report accuracy. The game arena consists of eight games from OpenSpiel in which the model plays against Jev 1.13.0, choosing each move from the set of legal moves (\cref{app:evaldetails}). Its score is $(\text{wins} + \nicefrac{1}{2}\,\text{draws}) / \text{games}$, which we report with a 95\% bootstrap confidence interval, with the number of games shown in parentheses.}
\label{tab:traj}
\small
\setlength{\tabcolsep}{3.2pt}
\resizebox{\textwidth}{!}{%
\begin{tabular}{@{}lcccccccc@{}}
\toprule
 & & \multicolumn{5}{c}{Domain (skill)} & & \\
\cmidrule(lr){3-7}
Checkpoint & \shortstack{Decision Index\\0.2.1 (skill)} & Knowledge & Language & Retrieval & Tools & Arts & \shortstack{JevBench\\accuracy (\%)} & \shortstack{Game arena score\\vs.\ Jev 1.13.0 (\%)} \\
\midrule
Stage 1 SFT, step 3.6k & 58.27 & 44.9 & 64.6 & 61.1 & 73.0 & 43.9 & 82.3 & 57.0 [51.6, 62.3] (256) \\
Stage 1 SFT, step 6.8k & 60.02 & 47.5 & 65.3 & 61.2 & 75.5 & 48.0 & 84.4 & 60.9 [55.5, 66.2] (256) \\
Stage 1 SFT, step 9.0k (end) & 60.01 & 47.4 & 65.6 & 61.1 & 75.5 & 47.9 & 84.0 & 57.6 [50.3, 64.6] (144) \\
Stage 2 specialization, step 175 & 60.38 & 47.2 & 65.8 & 62.7 & 75.2 & 48.4 & 87.0 & 60.5 [55.1, 65.6] (256) \\
Stage 2 specialization, step 250 & 60.65 & 47.7 & 65.9 & 62.9 & 75.5 & 48.9 & 85.7 & 60.2 [54.9, 65.2] (256) \\
Stage 2 specialization, step 325 & 60.56 & 47.7 & 65.8 & 62.8 & 75.5 & 48.6 & 86.1 & 58.0 [52.3, 63.5] (256) \\
Stage 3 RLAF (= Drex v2.0) & 60.78 & 47.7 & 65.9 & 62.9 & 76.0 & 49.1 & 85.7 & 60.9 [55.5, 66.2] (256) \\
\bottomrule
\end{tabular}}
\end{table}

\begin{figure}[t]
\centering
\includegraphics[width=\textwidth]{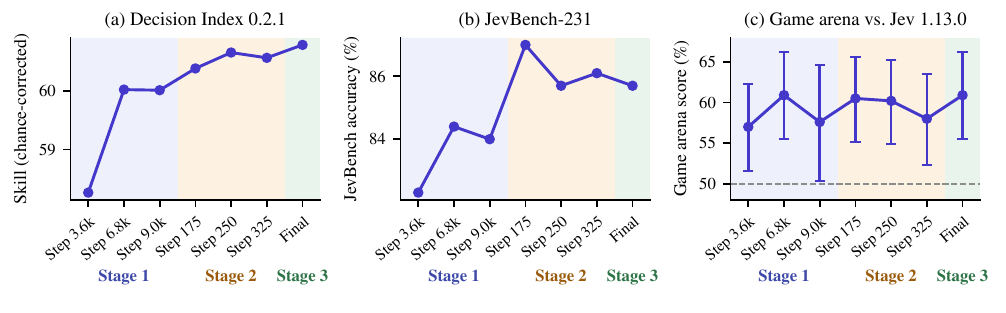}
\caption{\textbf{\drex v2.0 across training stages.} Stage~1 (S1) is supervised fine-tuning, Stage~2 (S2) is the specialization stage (\cref{sec:specialist}), and Stage~3 (S3) is reinforcement learning from adjusted feedback (RLAF). Most of the improvement in the Decision Index 0.2.1 skill score occurs in Stage~1. Stage~2 mainly improves the skill in the Retrieval domain and the accuracy on JevBench-231, and Stage~3 gives small further gains in the skill in the Tools and Arts domains. In the game arena against Jev 1.13.0 (\cref{tab:traj}), the 95\% confidence intervals of the score overlap at every checkpoint.}
\label{fig:trajectory}
\end{figure}

We evaluated seven checkpoints that were saved during the training of \drex v2.0 (\cref{tab:traj,fig:trajectory}).
We evaluated each checkpoint on all benchmarks of Decision Index 0.2.1, for which we report the chance-corrected skill score, overall and in each of its five domains, and on JevBench-231, for which we report accuracy.
We also evaluated each checkpoint in a game arena from OpenSpiel~\citep{openspiel}, where the model plays eight games against Jev 1.13.0, and it chooses each move by answering a single \texttt{choice} question whose options are the set of legal moves.
The score of a model in the game arena is $(\text{wins} + \nicefrac{1}{2}\,\text{draws}) / \text{games}$.
We find that most of the improvement in the Decision Index skill score occurs in \emph{Stage~1}: the score rises from 58.27 at step 3.6k to 60.02 at step 6.8k and then plateaus (60.01 at step 9.0k).
Between steps 3.6k and 6.8k, the skill improves in the Knowledge domain (+2.6), the Arts domain (+4.1), and the Tools domain (+2.5).
\emph{Stage~2}, the specialization stage, raises the Decision Index skill score by 0.55 points by step 325, and by 0.64 points at its best checkpoint, at step 250.
This improvement is almost entirely due to improvement in the Retrieval domain, in which the skill rises by 1.8 points by step 250.
Stage~2 also gives the largest improvement in JevBench accuracy, from 84.0\% to 87.0\% at step 175, although the accuracy then decreases to 85.7\% at step 250.
\emph{Stage~3}, RLAF, raises the Decision Index skill score by a further 0.2 points (+0.5 in the Tools domain and +0.5 in the Arts domain) and produces the released model, whose calibration at its deployed temperature is the best among the \drex models (\cref{sec:calib-results}).
However, we caution against over-interpreting these results.
First, differences of a few tenths of a point between consecutive checkpoints are comparable to the variation between repeated evaluations of the same model (which we measure to up to 0.8 points in the Decision Index score of \drex Max), so only the improvement in Stage~1 is clearly significant.
The 95\% confidence intervals of the game-arena scores also overlap at every checkpoint.
Second, these experiments are observational, not an ablation: Stage~3 starts from the Stage~2 checkpoint, so it demonstrates the effect of each stage appended to the preceding stages, not how each stage contributes to training in isolation.
We did not measure the calibration of the intermediate checkpoints, so the effect of Stage~3 on the calibration of v2.0 remains an open question.

\subsection{Controlled ablations on \drex Max}
\label{sec:maxablate}

We compare variants of \drex Max that share the backbone, the training data, and the training schedule of the released model, but each of which differs from it in one training component (\cref{tab:maxablate}): (1) a model trained without RLAF, (2) a model trained without the consistency terms in the loss function in both SFT and RLAF, and (3) a model trained without the vision training stage.
The models trained without RLAF and without the vision training stage are intermediate checkpoints that were saved during the training of the released model.
The model trained without the consistency terms $\ell_\pi$ and $\ell_s$ was trained separately, with SFT and then RLAF, from the same base model, with $\lambda_\pi = \lambda_s = 0$ and otherwise with the hyperparameters listed in \cref{tab:maxhp} in the Appendix.
We evaluate each variant on all public requests of Decision Index 0.4.1 (\cref{sec:calib-results}) and on the \OrdN{} decisions of our study of option-order robustness (\cref{sec:order}).

\begin{table}[t]
\centering
\caption{\textbf{Controlled comparisons on \drex Max.} Each row corresponds to one variant of \drex Max. For each variant, the table reports four evaluation results and one fitted parameter. The first result is the Decision Index 0.4.1 score on the public benchmarks. For the released model, this score differs from the score of 67.52 on the leaderboard (\cref{tab:di032}) because of the variation between repeated evaluations of the same model. The second and third results are the accuracy and the ECE on the held-out half of the \texttt{choice} decisions of the Decision Index (\cref{sec:calib-results}). The fourth result is the flip rate under a random permutation of the options, on all 1{,}591 decisions of our study of option-order robustness, and on the 173 of those decisions with $K \ge 6$ options (\cref{sec:order}). The fitted parameter is the temperature $T^\star$ that is fitted on the other half of the Decision Index decisions. All variants are evaluated on the same set of decisions.}
\label{tab:maxablate}
\small
\setlength{\tabcolsep}{4pt}
\begin{tabular}{@{}lcccccc@{}}
\toprule
Variant & \begin{tabular}[b]{@{}c@{}}Decision\\Index\end{tabular} & \begin{tabular}[b]{@{}c@{}}Accuracy\\(\%)\end{tabular} & ECE (\%) & $T^\star$ & Flip (\%) & Flip, $K{\ge}6$ (\%) \\
\midrule
SFT only & 64.9 & 86.5 & 4.4 & 1.45 & 5.3 & 14.1 \\
SFT $\to$ RLAF, without $\ell_\pi$, $\ell_s$ & 66.4 & 87.7 & 3.9 & 1.38 & 6.3 & 17.8 \\
SFT $\to$ RLAF (before vision SFT) & 66.9 & 88.0 & 3.5 & 1.34 & 4.2 & 13.0 \\
\textbf{Released: SFT $\to$ RLAF $\to$ vision SFT} & 66.8 & \CalMaxAcc & 3.6 & \CalMaxT & \OrdMaxFlip & \OrdMaxFlipSix \\
\bottomrule
\end{tabular}
\end{table}

\paragraph{Effect of RLAF.}
RLAF has the clearest effect. Relative to SFT alone, it adds 2.0 points to the Decision Index score and 1.5 percentage points of accuracy and lowers ECE by 0.9 percentage points. All three differences are statistically significant.

\paragraph{Effect of the consistency loss terms.}
We find that the consistency terms in the loss function matter for the robustness of the model to the ordering of the options, but not for its accuracy.
Without these terms, the model after RLAF reaches nearly the same Decision Index score (66.4 vs.\ 66.9, which is within the variation between repeated evaluations of the same model) and nearly the same accuracy as the model trained with these terms.
However, its flip rate rises from 4.2\% to 6.3\% (about 67 vs.\ 100 of the 1{,}591 decisions).
On the decisions with $K \ge 6$ options, the flip rate is also higher for the model trained without the consistency terms (from 13.0\% to 17.8\%), but this subset is too small for the difference to be statistically significant.
The model trained without the consistency terms in either stage also has a higher flip rate than the model trained with SFT alone with the terms (6.3\% vs.\ 5.3\%).
This may be due to the judge's preferences for specific orderings being distilled into the target model during RLAF.
However, these two models differ in both training stages, so this comparison does not properly isolate the effect of RLAF.
In addition, RLAF with the consistency terms lowers the flip rate by 1.1 percentage points relative to SFT alone (from 5.3\% to 4.2\%).

\paragraph{Effect of the image-encoder training stage.}
We compare the released model with the model trained with the same pipeline but without the image-encoder training stage.
On text inputs, the two models perform the same within the margin of error: the Decision Index score changes from 66.9 to 66.8, the accuracy from 88.0\% to 87.9\%, and the flip rate from 4.2\% to 4.3\%.
On image inputs, the image-encoder training stage improves the accuracy of the model when it receives the images themselves rather than their text rendering. On the 60 image-only questions of Decision Bench, the accuracy of the model rises from \DBMaxTxtImg\% to \DBMaxPixImg\% (\cref{sec:dbench}).

\section{Deployment and inference optimization}
\label{sec:serving}
In this section, we discuss how the \drex models are served in production settings. Serving a decision
model involves a different set of challenges as compared to serving a language model. Inference for language models is heavily focused on optimizing the decode
loop. On the other hand, decision models do not have a decode loop. They read the request (a state plus one or
more typed questions) in a single prefill pass, and a pointer head returns a
probability distribution over each question's options. The state is encoded once
and shared by all questions. As a result, the entire cost of each request is contained in one compute-heavy pass. This presents a unique set of considerations: short requests
are dominated by kernel-launch overhead, long states strain GPU memory, and the
latency budget is tight since decision models usually sit in the request path
of another system. The techniques in this section discuss how we address these challenges.

\subsection{Latency optimizations}
\label{sec:inferencetechniques}

\begin{table}[t]
\centering
\caption{\textbf{Latency and cost.} The table shows the median latency of a single request and the published price per 1{,}000 decisions of each model in October 2026, when \drex Max was released. We measured these latencies from a test machine in Oregon, which includes 70--215\,ms of network time between the test machine and the servers.}
\label{tab:efficiency}
\small
\setlength{\tabcolsep}{4pt}
\begin{tabular}{@{}lccccccc@{}}
\toprule
 & \drex Max & \drex v2.0 & \drex v1.5 & Jev 1.14 & Luna Dec. & GPT-6 Sol & Opus 5.5 \\
\midrule
Median latency & 183\,ms & 151\,ms & \textbf{145\,ms} & 196\,ms & 326\,ms & $\sim$2.0\,s & $\sim$3.5\,s \\
Cost / 1k decisions & \$0.06 & \$0.06 & \textbf{\$0.04} & \$0.042 & \$0.10 & \$3.33 & \$10.17 \\
\bottomrule
\end{tabular}
\end{table}

A decision model answers a request in a single forward pass, so the latency of a request is dominated by the runtime of this forward pass.
This subsection first describes the two stages of the forward pass of a pointer model and shows which stage dominates the latency.
It then describes three optimizations that reduce this latency: shared prefill, which processes the context once for all questions in a request; CUDA Graphs, which remove the overhead of launching many small GPU kernels; and kernel optimizations, which make individual GPU kernels faster.
\Cref{sec:servinggate} describes how we ensure that these optimizations do not lower accuracy.

\paragraph{Two stages of the forward pass.}
The forward pass of a pointer model on a request can be divided into two stages.
In the \emph{state pass}, the model encodes the context, and the cost of this stage grows with the length of the context.
In the \emph{question pass}, which runs once per question, the model processes only the tokens of that question, which is typically very short and so this pass is cheap.
We distinguish these two passes because a question never attends to the tokens of another question, so the predicted probability for a given question depends only on the context and on the content of that question.
For a given context, only the cost of the first question includes the cost of the state pass. For every subsequent question, whether in the same request or in a later request that retrieves the encoded context from cache, the cost is much lower since the state pass can be skipped.
For example, given a context containing over 100k tokens, once the context has been encoded and cached via the state pass (which is done when processing the first question), each subsequent question about this context is answered in 21.6\,ms, which is less than the time the state pass takes to encode a new context of only 2k tokens (35.6\,ms).
The state pass on new contexts therefore dominates the cost of inference: given a new context of 131{,}072 tokens (the maximum context size for \drex), a request takes about 2.7\,s on a dual H100 configuration, above our 2\,s target for that worst case.

\paragraph{Shared prefill.}
A general-purpose inference engine treats each question as a separate sequence that includes the full context, so a request with many questions encodes its context once per question.
We instead process the context of a batch of questions once, in a state pass, and then run the question pass for each question, which removes the redundant processing of the context.
The speedup grows with the length of the context. For queries with 16 questions, the speedup grows from 6--8$\times$ for a context length of 8k tokens to 11--12$\times$ at 32k tokens. For contexts of at least 1{,}024 tokens, the outputs of this two-stage inference approach are exactly identical to those of a one-stage inference pipeline.  %and the outputs are bit-identical to those without it.
For contexts shorter than 1{,}024 tokens, the shared prefill optimization results in increased latency and small errors in the output probabilities (\cref{rem:independence}), so this optimization is not applied in these cases.

\paragraph{CUDA Graphs.}
For short requests, the GPU is idle for most of the forward pass, because much of the time is spent launching thousands of small kernels.
A CUDA Graph~\citep{rv_cudagraphs} provides a way to mitigate this overhead. A sequence of kernels is ``recorded'' and can later be ``replayed'' via a single launch event. %replaying a ``recorded'' sequence of kernels with a single launch.
For a given range of shapes of the input (i.e., ``input shape bucket''), we record the kernels of the forward pass of the backbone and the pointer head as a single CUDA Graph. We repeat this for all input shape buckets. %Capturing the backbone and the pointer head as one CUDA Graph per shape bucket (a range of input shapes)
Utilizing the CUDA Graph cuts the latency of a single question at 1k tokens by about two thirds.
The benefit disappears for contexts of 6k--8k tokens, where the cost of kernel launches becomes negligible compared to the computation cost of the forward pass. %takes longer than launching its kernels.

\paragraph{Kernel optimizations.}
In order to further reduce latency, we also considered several optimizations of GPU kernels (\cref{app:servemodel}).
Each kernel optimization was evaluated by comparing the model outputs against the outputs from the unoptimized model on real questions. Each kernel optimization is accepted only if the most likely option (i.e., the decision) did not change and the predicted probabilities deviated by no more than a small tolerance (\cref{sec:servinggate}).
A grouped attention kernel looked promising on synthetic benchmarks but was about three times slower on real records, which have short contexts and long question regions.
The above optimization evaluation procedure revealed a packed-prefill bug, an error in encoding packed requests (\cref{sec:packing}) that returned valid responses with wrong answers and would not have been revealed in latency measurements alone.
\Cref{app:servemodel} details the latency measurements of these optimizations for \drex v1.5, v2.0, and Max. \drex Max cannot share the encoded context between questions (\cref{sec:maxcost}), so its inference optimizations aim to increase the speed of individual GPU kernels on Blackwell GPUs (B200 and B300).

\subsection{Accuracy checks for inference optimizations}
\label{sec:servinggate}
\label{sec:latencycost}
\label{sec:servingdeploy}
When experimenting with inference optimizations, we classify each optimization into one of three classes:
\begin{itemize}
    \item \emph{Exact}: The output of the model with the optimization is exactly identical to that of the model without the optimization (e.g.\ parallel tokenization and CUDA Graphs over question sequences).
    \item \emph{Bounded-error}: The optimization causes the outputs to deviate slightly (e.g.\ FP8 and kernel fusion).
    \item \emph{Rejected}: The deviation in the outputs as a result of the optimization is too large, or the optimization causes latency to increase. \Cref{app:servemodel} lists each.
\end{itemize}
In developing \drex, we include all exact optimizations but only a subset of bounded-error optimizations. We choose to accept a bounded-error optimization only if it does not lower accuracy beyond a model-specific threshold.
We score the same questions with and without the change and compute a 95\% paired bootstrap interval on the difference in accuracy.
\begin{table}[t]
\centering
\caption{\textbf{Accuracy cost of bounded-error optimizations} (\drex v1.5). Each row compares one optimization with the unoptimized version, on the same questions for both versions. We note that the baseline model differs across rows. Accuracy differences are in units of percentage points with 95\% paired bootstrap intervals, resampled by document. The last two rows are measured on a 365-question panel, where each decision is worth 0.27 percentage points.}
\label{tab:ledger}
\small
\setlength{\tabcolsep}{3.6pt}
\begin{tabularx}{\textwidth}{@{}>{\raggedright\arraybackslash}Xcccc@{}}
\toprule
Optimization & Answers changed & Accuracy difference & Number of questions & GPU \\
\midrule
Short-context build & 79 / 5{,}966 & $+0.06$ [$-0.05$, $+0.18$] & 28{,}789 & B200 \\
Long-context build & 2 / 127 & $\phantom{+}0.00$ [$-0.41$, $+0.43$] & 2{,}737 (204 docs) & B200 \\
FP8 vs.\ BF16 & --- & $\phantom{+}0.00$ & 1{,}934 & B200 \\
Graph padding + small-batch FP8 & --- & $-0.34$ [$-0.55$, $-0.13$] & 11{,}932 & H100 \\
Kernel fusion (short contexts) & 1 / 365 & --- & 365 & B200 \\
H100 graph policy (self-hosting) & 11 / 365 & --- & 365 & H100 \\
\bottomrule
\end{tabularx}
\end{table}
The rows of \cref{tab:ledger} are the following optimizations.
\begin{itemize}\itemsep0pt
    \item \emph{Short-context build} is a new build of our inference code for requests with contexts of up to 8k tokens. It is compared with the previous release.
    \item \emph{Long-context build} is a new build for requests with contexts of up to 131{,}072 tokens, which run one request per GPU. It is compared with the code in production.
    \item \emph{FP8 vs.\ BF16} compares FP8 matrix multiplications with the same code in BF16.
    \item \emph{Graph padding + small-batch FP8} reduces the overhead of CUDA graphs and of padding the key-value cache, and adds an FP8 matrix-multiplication kernel for small batches. It is compared with the code before these changes, which already used FP8.
    \item \emph{Kernel fusion (short contexts)} fuses GPU kernels, for example normalization with quantization, and tunes kernel settings for short contexts. It is compared with the version without these fused kernels.
    \item \emph{H100 graph policy (self-hosting)} uses CUDA-graph and attention settings tuned for users who run the model on their own H100 GPUs. It is compared with the settings that we use on B200 GPUs, with both versions running on the same H100 GPU.
\end{itemize}
Most optimizations stayed within the threshold and changed only a few answers (\cref{tab:ledger}). For optimizations that led to changes in accuracy greater than the threshold, we weighed the accuracy lost against the speedup before deciding to include the optimization into the final model.
The accuracy of the end-to-end inference stack, compared with the native-precision model, also stays within the threshold.

\subsection{Enterprise models}
\label{sec:enterprise}
Besides the four open-weight \drex models, we serve enterprise models through the hosted \sysone API at \bluehref{https://console.nace.ai}{\texttt{console.nace.ai}}.
These currently include \drex Max v1.1, our strongest general decision model, and \drex for Finance, a variant specialized for banking, accounting, insurance, compliance and financial-reporting decisions.
Unlike the models described in this report, the enterprise models are not released as open weights: they are built on internal proprietary models and are available only through the API, including deployments on customer hardware.
These models use the same request format, question types and readout semantics as the open-weight models, so they can serve as plug-in replacements in applications. %can switch between them by changing only the model identifier.
We do not disclose their architecture, training data or training recipe in this work. Unless stated otherwise, all results in this report are related to the open-weight \drex models.

\section{Discussion and limitations}
\label{sec:limits}
\paragraph{Pointer and readout heads.}
The pointer head and the readout head suit different use-cases.
The pointer head shares one encoding of the context across every question, which makes processing many questions over long documents cheap and the questions are independent of each other by construction, up to floating-point error.
On the other hand, the readout head encodes the context again for each question, with non-causal attention in its global attention layers (\cref{sec:readout}).
\drex Max, which uses the readout head, is more accurate, especially on knowledge-intensive and multi-step decisions.
Because \drex Max also differs from the pointer models in its attention mask, its dense 27B backbone, and its training data, isolating the contribution of each of these components is left to future work.
One such study would compare two decision models that use the same Qwen3.8-27B backbone and the same training data: one with a pointer head whose query and key projections are 256-dimensional (\cref{sec:arch}), and one with a readout head. Furthermore, such a study would compare each model trained with and without the non-causal attention mask.

\paragraph{Closed-book knowledge.}
Every \drex model trails frontier chat models on evaluations that test closed-book knowledge, and the gap is largest on the hardest questions (HLE, GPQA).
An output head can only rank options using the knowledge in the backbone model.
A single forward pass may also be a poor substitute for long chain-of-thought reasoning and tool use.
An approach which may close this gap would be to 
%We have not tested this: such an experiment would
let the model generate a chain of thought, or call tools such as a search engine, before its output head ranks the options. Future studies would train such a model and compare its accuracy on these benchmarks with that of the same model using a single forward pass.

The gap in performance on evaluations that test closed-book knowledge narrows from v1.5 to v2.0 to Max.
On the 37 benchmarks of the public subset of Decision Index 0.4.1, the skill score of the \drex models in the Knowledge domain, as reported on the leaderboard (\cref{tab:di032}), rises from 44.6 for v1.5 to 47.9 for v2.0 and to 57.1 for Max.
However, on these tasks, our largest model still does not match the performance of frontier chat models.

\paragraph{Calibration and robustness to option order.}
\drex Max is slightly over-confident in its predictions: in each of the four domains other than Language, its mean confidence exceeds its accuracy by 5--9 percentage points, and on questions with 6 to 20 options, its ECE is 8.4\% (\cref{sec:calib-results}).
In addition, its deployed temperature ($T_{\mathrm{dep}}{=}1.009$), which was fitted on held-out records from our training distribution, differs from the temperature $T^\star$ that is fitted on held-out Decision Index decisions (\cref{sec:calib-results}).
On questions with six or more options, when the order in which the options are listed in the request is randomly permuted, 13\% of its decisions have a different most probable option (\cref{sec:order}).
Both miscalibration patterns are most apparent in low-confidence decisions: the decisions whose most probable option changes when the options are reordered are largely the decisions to which the model assigns the lowest confidence (\cref{tab:ordertta} in the Appendix).
An application or pipeline that uses the model can therefore reduce the impact of both miscalibration patterns by adding a step to first inspect the output confidence: The pipeline proceeds normally only on decisions whose confidence (the probability of the most probable option) is above a threshold. Otherwise, the pipeline sends the remaining decisions to human review or to a different system for further scrutiny.
The threshold should be tuned on data from the setting in which the model is deployed, since the confidence of the model at a given accuracy differs between domains and between numbers of options.
Our study of option order uses 1{,}591 decisions and excludes requests whose options are structured objects rather than strings (\cref{app:order}).
A larger study would give more precise estimates.
A study that also replaces the option codes of \drex Max (\cref{sec:readout}) with other tokens, for example 1, 2, 3 instead of A, B, C, would also show whether the model's behavior depends on the specific codes, in addition to the order of the options.

\paragraph{Comparability.}
The systems on the Decision Index leaderboard (\cref{tab:di032}) differ from each other in several respects simultaneously: the backbone model, the training data, the training objective, and the inference setup.
Therefore, the difference between the scores of two systems in \cref{tab:di032} should not be attributed to any single design decision.
\Cref{app:limits} discusses four additional limitations of our study: the scope of our evaluation; the latency of the models on long inputs; how we verify that optimizations of the inference code do not adversely affect accuracy (e.g., reduced numerical precision); and how applications should deploy and monitor a decision model responsibly.

\section{Conclusion}
\label{sec:concl}
For each input request, \drex outputs a probability distribution over a set of options that the user defines in the request.
The three pointer models (\drex DLM, v1.5, and v2.0) are built on the open \kev code base.
They encode the context of a request once and share this encoding across all of its questions, they keep the answer to each question independent of the other questions, and they compute the probability of the options of each question with the pointer head (\cref{sec:pointer}).
The models are designed to produce outputs with very low latency.
On \drex v1.5, a new question about a cached state of 114--117k tokens is answered in 21.6\,ms (\cref{sec:serving}).
\drex Max uses a dense multimodal backbone with 27B parameters, removes the causal attention mask in its full-attention layers, and encodes the context again for every question (\cref{sec:max}).
On Decision Index 0.4.1, it scores 67.52 on the public 37 benchmarks and 66.60 on the full score, which includes the private subset (\cref{tab:di032}).
On Decision Bench v4, with images provided to the model via its image encoder, its accuracy is \DBMaxPix\%, which is not significantly different from that of GPT-6 Sol and d3-turbo, and 1.6 percentage points below that of Claude Opus 5.5 (\cref{sec:dbench}).
Its latency on Decision Bench is 0.66\,s, compared with 2.1\,s for GPT-6 Sol and 4.3\,s for Claude Opus 5.5 (\cref{sec:serving}).

Through our experiments, we are able to provide answers to the main research questions:
\textbf{RQ1 (calibration):} The \drex models, which are trained with proper scoring rules, have an ECE of 1.4--4.5\% at their deployed temperatures on the Decision Index decisions, which decreases to 0.8--1.4\% after tuning the temperature on Decision Index examples (\cref{sec:calib-results}).
However, the ECE within individual domains and on questions with many options is higher, so a single temperature parameter is not a substitute for monitoring calibration on each workload.
\textbf{RQ2 (option order):} A random permutation of the order of the options changes the most probable option in 4.3--5.9\% of the decisions of each model, and changes accuracy by at most 1.2 percentage points (\cref{sec:order}).
On the workloads that we evaluated, the decisions that change are largely those to which the model already assigns low confidence.
\textbf{RQ3 (training):} During the training of v2.0, most of the improvement in the Decision Index score was made in the first stage, SFT, and the later stages added smaller gains in specific domains (\cref{sec:dynamics}).
For Max, adding RLAF after SFT raises the Decision Index score by 2.0 points, and the consistency terms in the loss function lower the flip rate from 6.3\% to 4.2\% while leaving the Decision Index score nearly unchanged (\cref{tab:maxablate}).
\textbf{RQ4 (capacity, attention, and modality):} Max differs from the pointer models in the size of its backbone, its attention mask, its output head, and its training data.
It is more accurate than the pointer models, especially in the Knowledge domain of Decision Index.
There are a number of questions that remain open.
All \drex models trail frontier chat models on evaluations that test closed-book knowledge.
Further research might provide ways to mitigate this limitation, perhaps by routing knowledge-intensive queries to other systems.
Additionally, architectural ablations, such as a comparison of the pointer head and the readout head and of the causal and non-causal attention masks on the same backbone, are needed to show how much each component contributes to the gains of Max.

\small
\sloppy
\bibliographystyle{plainnat}
\bibliography{refs}
\normalsize

\appendix

\section{Extended related work}
\label{app:related}

\paragraph{Discriminative scoring.}
Scoring each candidate option by its likelihood under the language model is the standard way to evaluate language models on multiple-choice tasks~\citep{brown2020gpt3,hendrycks2021mmlu}.
This score is biased by the length and exact wording of each option (for example, longer options tend to receive a lower likelihood), and it requires a separate forward pass for each option.
In \emph{multiple-choice prompting}, all options are presented in one prompt, each labeled with a symbol (\texttt{A}, \texttt{B}, \dots), and the model scores the symbol tokens instead of the option texts~\citep{robinson2023mcqa}.
This removes the length bias and requires a single forward pass for all options, but it works only if the model correctly associates each symbol with the option that it labels.
The readout head of \drex Max (\cref{sec:max}) is a trained version of this approach: a bias-free linear layer maps the final hidden state to one logit per option code.
Cross-encoders~\citep{devlin2019bert,nogueira2019passage} score each (query, candidate) pair with a classification head, so they require one forward pass per candidate, and their cost grows linearly with the number of candidates.
Pointer networks~\citep{vinyals2015pointer} output a probability distribution over the positions of their input, and therefore over a set of candidates whose size can differ from one input to the next.
The \drex pointer models combine these two ideas.
As in a cross-encoder, the options are placed in the input together with the question, delimited by special tokens that user text cannot produce (\cref{sec:packing}).
As in a pointer network, a pointer head then selects among the options, using a single forward pass over the state and the question.

\paragraph{Option-order sensitivity.}
Language models answering multiple-choice questions favor particular option positions or symbols, so permuting the options changes their answers~\citep{zheng2024mcq,pezeshkpour2024order}.
Averaging the predictions over permutations of the options reduces this bias, but it multiplies the inference cost by the number of permutations.
The options of a decision model are often actions that an application will take, so its predicted distribution should not depend on the arbitrary order in which the developer lists them.
For a decision model, sensitivity to option order is therefore an error of the model rather than an artifact of the benchmark.
Option isolation, an attention mask under which each option attends only to the state, the instruction, and itself, makes the pointer head permutation-equivariant by construction (\cref{sec:attn}).
It cannot be applied to the hybrid backbones of \drex v1.5, v2.0, and Max, since their recurrent Gated DeltaNet layers cannot apply an arbitrary attention mask.
\Cref{sec:order} measures how robust to option order these models become through training alone.

\paragraph{Calibration and proper scoring.}
Modern neural networks are often over-confident~\citep{guo2017calibration}.
Temperature scaling, which divides the logits by a single constant fitted on held-out data, is a simple and effective fix that is applied after training, and it does not change the ranking of the options~\citep{guo2017calibration}.
After pre-training, the token probabilities of language models are reasonably well calibrated on multiple-choice questions~\citep{kadavath2022know,jiang2021know}.
Confidences that a model states in text (verbalized confidences) and the probabilities of models after preference fine-tuning are often less well calibrated~\citep{tian2023justask}.
A scoring rule is \emph{proper} if its expected value is optimized by reporting the true distribution of the outcome, which makes proper scoring rules natural objectives when the output is a forecast rather than a label~\citep{brier1950,gneiting2007proper}.
The four that we use are the log score, the spherical score, the Brier score, and, for outcomes on an integer scale, the squared Cram\'er distance between the predicted and the observed cumulative distributions~\citep{rv_bellemare2017cramer}.
We use them both as training objectives (\cref{sec:rlaf}) and as evaluation metrics, together with expected calibration error~\citep{naeini2015ece} and precision@top-$k$\%~\citep{geifman2017selective}.

\paragraph{Backbones and attention patterns.}
The \drex decision heads can be attached to backbones of different architectures, and the family uses three kinds (\cref{tab:family}):
\begin{itemize}
  \item \drex DLM uses an autoregressive-to-diffusion conversion with block-bidirectional attention~\citep{fu2025efficientdlm}, which is related to masked diffusion language models~\citep{sahoo2024mdlm,nie2025llada};
  \item \drex v1.5 and \drex Max use hybrids of Gated DeltaNet linear-attention layers and full-attention layers~\citep{yang2025gateddeltanet,mimo2026qwen9b,qwen38};
  \item \drex v2.0 uses a hybrid of the same kind whose feed-forward blocks are sparse mixtures of experts~\citep{shazeer2017moe,fedus2022switch,qwen36moe}.
\end{itemize}
Making part of the attention non-causal in a decoder-only model is similar to prefix language models~\citep{raffel2020t5,dong2019unilm}.
\drex Max removes the causal mask only in the full-attention layers of its hybrid backbone (\cref{sec:max}).
\drex Max processes images with the standard architecture in which an image encoder turns each image into a sequence of embeddings that the language model receives together with the text tokens~\citep{dosovitskiy2021vit,liu2023llava}.
The specialization training stages, which are optional stages that continue training on a narrow family of tasks (\cref{sec:s2e}), use low-rank adaptation~\citep{hu2022lora}.

\paragraph{Decision benchmarks.}
Decision Index~\citep{decisionindex} aggregates decision benchmarks from several domains: knowledge and reasoning (e.g., MMLU-Pro, GPQA, and BBH~\citep{wang2024mmlupro,rein2024gpqa,suzgun2023bbh}), language, retrieval, tool use, and arts.
It reports the score on each benchmark as a skill that is corrected for chance, so that 0 corresponds to uniform guessing (\cref{sec:setup}).
Decision Bench~\citep{decisionbench2026} samples decision questions with a fixed set of options from 36 public datasets of real-world records, such as receipts, documents, and photographs.

\section{Question types and metrics}
\label{app:metrics}

\begin{table}[t]
\centering
\caption{\textbf{The three question types}, with what the user supplies and what the model returns. Every type returns a probability distribution over options supplied by the user.}
\label{tab:types}
\small
\begin{tabular}{@{}lll@{}}
\toprule
Type & User supplies & Model returns \\
\midrule
\texttt{choice} & named options, $K \le 255$ (key order $=$ option order) & $\arg\max$ key, $p_j$ for every key, $c_{\texttt{choice}}$ \\
\texttt{noul}   & a yes/no question; optional definitions of yes/no & $p(\text{yes}) \in [0,1]$ \\
\texttt{score}  & an integer scale, lowest level first & $\hat s$, per-level $p_i$, $c_{\texttt{score}}$ \\
\bottomrule
\end{tabular}
\end{table}

\paragraph{Metrics.}

For a set of decisions with predicted distributions $p^{(n)}$ and gold option indices $y^{(n)}$, let $c^{(n)} = \max_j p^{(n)}_j$ and $a^{(n)} = \mathds{1}[\arg\max_j p^{(n)}_j = y^{(n)}]$.
We report expected calibration error over $B{=}15$ equal-width confidence bins $\{\mathcal{B}_b\}$~\citep{naeini2015ece},
\begin{equation}
  \ECE = \sum_{b=1}^{B} \frac{|\mathcal{B}_b|}{N} \Bigl|\, \tfrac{1}{|\mathcal{B}_b|}\textstyle\sum_{n \in \mathcal{B}_b} c^{(n)} - \tfrac{1}{|\mathcal{B}_b|}\textstyle\sum_{n \in \mathcal{B}_b} a^{(n)} \Bigr|,
  \label{eq:ece}
\end{equation}
negative log-likelihood $-\log p^{(n)}_{y^{(n)}}$, the multi-class Brier score $\lVert p^{(n)} - e_{y^{(n)}} \rVert_2^2$, and \emph{precision@top-$k$\%}, which is the accuracy of the $\lceil \frac{k}{100} N \rceil$ most confident decisions~\citep{geifman2017selective}.

\paragraph{Order-robustness metrics.}
For order robustness, we compare the distribution $p$ under the original order of the options with the distribution $p^{\pi}$ under a permutation $\pi$ of the options, after mapping $p^{\pi}$ back to the original options.
We report the \emph{flip rate} $\Pr[\arg\max p \neq \arg\max p^{\pi}]$, i.e., the fraction of decisions whose top option changes, and the mean total-variation distance $\TV(p, p^{\pi}) = \tfrac12 \lVert p - p^{\pi} \rVert_1$.

\section{Pointer architecture details}
\label{app:archdetails}

\subsection{Packing}
\label{sec:packing}

A request is converted into a single token sequence, which we call \emph{packed}, consisting of a state segment followed by one question block per question~\citep{kev2026}:
\begin{equation}
  \underbrace{\tok{state}\; S}_{\text{segment } 0}
  \;\;
  \underbrace{\tok{q}\, u_1 \textstyle\prod_{j}\bigl(\tok{opt}\, o_{1,j}\, \tok{/opt}\bigr) \tok{decide}}_{\text{segment } 1 \,=\, b_1}
  \;\cdots\;
  \underbrace{\tok{q}\, u_Q \cdots \tok{decide}}_{\text{segment } Q \,=\, b_Q},
  \label{eq:packing}
\end{equation}
where $u_k$ is the instruction of question $k$, $o_{k,j}$ is its $j$-th option, and $\prod_j$ denotes the concatenation over the options of a question.
The five delimiter tokens (\tok{state}, \tok{q}, \tok{opt}, \tok{/opt}, and \tok{decide}) are rarely used special tokens that already exist in the Qwen vocabulary (\texttt{<|fim\_prefix|>}, \texttt{<|fim\_middle|>}, \texttt{<|box\_start|>}, \texttt{<|box\_end|>}, and \texttt{<|fim\_suffix|>}), so no new rows have to be added to the embedding matrix.
Before tokenization, user text is rewritten so that it cannot produce these tokens.
User text therefore cannot imitate a delimiter, and the boundaries between options are always those defined by the request.
Each token carries three IDs.
The segment ID $\seg(i)$ is $0$ for the tokens of the state and $k$ for the tokens of question block $b_k$.
The position IDs of every question block start immediately after the state, so a question block has the same position IDs regardless of how many questions precede it.
The option index identifies the option that the token belongs to.

\subsection{Segment-masked attention}
\label{sec:attn}

\begin{figure}[t]
\centering
\resizebox{\textwidth}{!}{\input{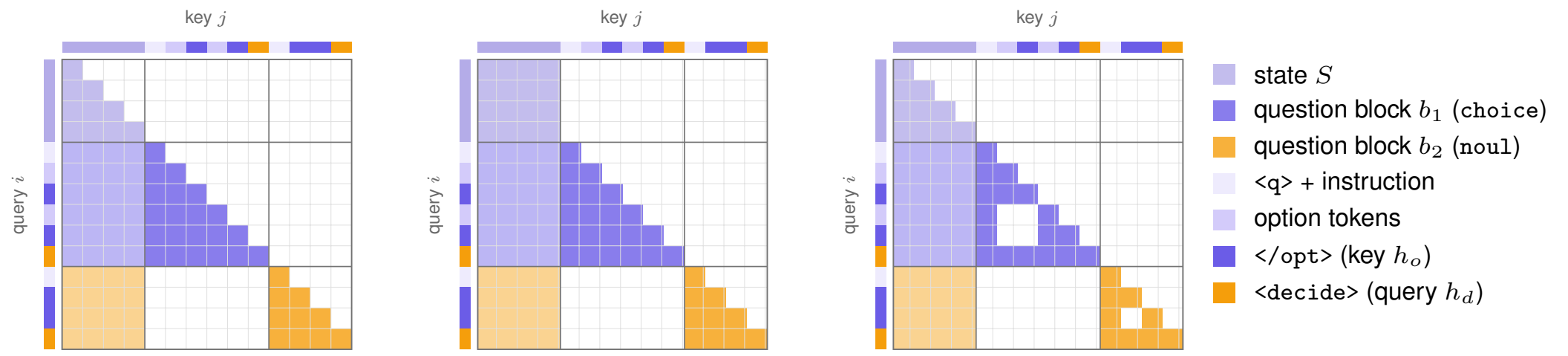}}
\caption{\textbf{Attention masks} for a request with a four-token state, a three-way \texttt{choice} question ($b_1$), and a \texttt{noul} question ($b_2$).
Filled cells mark the query--key pairs that are allowed to attend, and lighter cells mark tokens in the question block that attend to the shared state.
(a)~Block-causal mask of \cref{eq:mask}. (b)~\drex DLM adds bidirectional attention within the state.
(c)~Optional option isolation, in which \tok{decide} still attends to every option.
Hybrid backbones implement (a) by processing each question as a separate causal sequence $[S; b_k]$ that starts from the cached state.
The masks are generated by the code that implements them in the released models.}
\label{fig:masks}
\end{figure}

On backbones that consist only of attention layers, the packed sequence of \cref{eq:packing} is encoded under the \emph{block-causal} mask, under which token $i$ may attend to token $j$ ($M_{ij} = 1$) only if $j$ does not come after $i$ and $j$ belongs either to the state or to the same segment as $i$:
\begin{equation}
  M_{ij} = 1 \iff j \le i \;\wedge\; \bigl(\seg(j) = 0 \;\vee\; \seg(j) = \seg(i)\bigr),
  \label{eq:mask}
\end{equation}
This mask is shown in \cref{fig:masks}a.
Under this mask, the state never attends to a question, and no question attends to another question.
The pointer models use two variants of this scheme, described in the next two paragraphs.

\paragraph{Bidirectional state (\drex DLM).}
Efficient-DLM is trained with bidirectional attention inside blocks~\citep{fu2025efficientdlm}, so \drex DLM enables the \texttt{state\_bidir} option, which additionally sets $M_{ij}=1$ whenever both tokens belong to the state, i.e., $\seg(i)=\seg(j)=0$ (\cref{fig:masks}b).
Every token of the state then attends to every other token of the state.
The state still never attends to a question block, so the cached keys and values of the state do not depend on the questions and can be reused exactly for every question.

\paragraph{Per-question sequences (\drex v1.5, v2.0).}
Gated DeltaNet layers~\citep{yang2025gateddeltanet} process the sequence with a recurrent state that summarizes all previous tokens, so they cannot apply an arbitrary attention mask.
On hybrid backbones, each question $k$ is therefore processed as its own causal sequence $[S;\, b_k]$, which we call a \emph{question sequence}.
Each question sequence starts from the cached result of processing the state $S$, i.e., the key--value cache of the full-attention layers and the recurrent state of the DeltaNet layers.
Each question sequence contains exactly the tokens that \cref{eq:mask} allows question $k$ to attend to, at the same position IDs.
On attention-only backbones, the packed form and the per-question form therefore produce the same outputs up to floating-point error.

\paragraph{Option isolation.}
On attention-only backbones, the mask can additionally restrict each option span (the tokens from \tok{opt} to \tok{/opt}) to attend only to the state, the instruction of its question, and itself (\cref{fig:masks}c).
The \tok{decide} token still attends to every option span, and all option spans of a question start at the same position ID.
Under this mask, the hidden states of an option do not depend on the other options or on their order, so reordering the options reorders the probabilities of the pointer head in the same way: the pointer head is permutation-equivariant by construction.
Hybrid backbones cannot apply this mask, so they must learn robustness to option order from the training data instead (\cref{sec:objective}).
\Cref{sec:order} measures how robust to option order they become.

\begin{remark}[Question independence]
\label{rem:independence}
Under \cref{eq:mask} (or its bidirectional-state or per-question variants), with position IDs of the question block restarting after the state, the output distribution $p_k$ of question $k$ depends only on the state $S$ and question block $b_k$, since every attention or recurrent path into segment $k$ passes only through segments $0$ and $k$.
Adding, removing, or reordering other questions therefore leaves $p_k$ unchanged up to floating-point error.
\drex Max satisfies the same property trivially, since each question is encoded as its own sequence (\cref{sec:max}).
\end{remark}
This property follows directly from the mask.
It ensures that batched inference, replaying a subset of the questions of a request (e.g., for an audit), and caching individual questions produce the same outputs, up to floating-point error, as processing the full request.
The property assumes that every feed-forward or MoE layer processes each token independently of the other tokens.
This holds at inference for dense, hybrid, and MoE backbones, since MoE routing at inference does not drop tokens when an expert exceeds its capacity.

\subsection{Inference and cost}
\label{sec:inference}

\begin{algorithm}[t]
\caption{Inference for a \sysone request with a \drex pointer model}
\label{alg:serve}
\small
\begin{algorithmic}[1]
\Require state $S$; questions $\{(\tau_k, u_k, O_k)\}_{k=1}^{Q}$; backbone $f_\theta$; head $(W_q, W_k)$; temperature $T$
\State $x_S \gets \textsc{Tokenize}(\tok{state}\, S)$ \Comment{delimiters in user text are rewritten}
\State $\mathcal{C} \gets f_\theta(x_S)$ \Comment{encode once; KV cache and/or recurrent state}
\For{$k = 1,\dots,Q$ \textbf{in parallel}}
  \State $x_k \gets \textsc{Tokenize}\bigl(\tok{q}\, u_k \textstyle\prod_j (\tok{opt}\, o_{k,j}\, \tok{/opt})\, \tok{decide}\bigr)$
  \State $H_k \gets f_\theta(x_k \mid \mathcal{C})$ \Comment{reads the state, never another question block}
  \State $\mathbf{q} \gets W_q H_k[\tok{decide}]$;\quad $\mathbf{k}_j \gets W_k H_k[\tok{/opt}_j]$ for $j = 1..K_k$
  \State $p_k \gets \softmax\bigl(\{\mathbf{k}_j^\top \mathbf{q} / \sqrt{d_p}\}_j \,/\, T\bigr)$
  \State $a_k \gets \textsc{Format}(\tau_k, p_k)$ \Comment{$\arg\max$ and $c_{\texttt{choice}}$; $p(\text{yes})$; or $\hat s$ and $c_{\texttt{score}}$}
\EndFor
\State \Return $\{a_k\}_{k=1}^{Q}$
\end{algorithmic}
\end{algorithm}

\Cref{alg:serve} processes the state once and then processes each question block once (one prefill per block), starting from the cached state.
With $|S|$ state tokens and question blocks of $|b_k|$ tokens, the cost of the full-attention layers is $O\bigl(|S|^2 + \sum_k |b_k|(|S| + |b_k|)\bigr)$, and the cost of the DeltaNet layers is linear in $|S| + \sum_k |b_k|$.
Processing the state again for every question would instead cost $O\bigl(\sum_k (|S|+|b_k|)^2\bigr)$ in the full-attention layers.
The model does not generate tokens autoregressively: a request requires only two sequential forward passes, one over the state and one over all question blocks in parallel, regardless of the number of options.
On attention-only backbones, the packed sequence of \cref{eq:packing} allows the state and all question blocks to be processed in a single forward pass.
Our inference system is designed around this split between one long pass over the state and many short passes over the questions (\cref{sec:serving}).

\section{\drex Max: cost and training configuration}
\label{app:maxhp}
\label{sec:maxtrain}

\subsection{Cost and caching}
\label{sec:maxcost}

\begin{remark}[No shared state cache]
\label{prop:nocache}
Let $\mathcal{F}$ be the set of full-attention layers of \drex Max (\cref{sec:max}), and let $|\mathcal{I}|$ denote the number of image tokens.
Under \cref{eq:maxmask}, the hidden state of every state token at every layer $\ell \geq \min \mathcal{F}$, i.e., from the first full-attention layer onward, depends on the question and option tokens that follow it.
Consequently, the representation of $S$ cannot be computed once and reused across questions with different instructions or options.
Scoring $Q$ questions over a state of $|S|$ tokens therefore costs $O\bigl(\sum_k (|S| + |\mathcal{I}| + |b_k|)^2\bigr)$ in the full-attention layers and $O\bigl(\sum_k (|S| + |\mathcal{I}| + |b_k|)\bigr)$ in the DeltaNet layers.
\end{remark}
\noindent\emph{Proof sketch.} In layer $\min\mathcal{F}$, the attention output at a state position $i$ is a convex combination of the value vectors at all $n$ positions of the sequence, including the positions in $b_k$, with weights that depend on $b_k$ through the keys.
Every later layer, causal or not, takes this output as input at position $i$.
Hence two questions with different $b_k$ yield different state representations from layer $\min\mathcal{F}$ onward, and only the first $\min\mathcal{F}-1$ layers of the state can be shared.
The cost follows from encoding one sequence of length $|S| + |\mathcal{I}| + |b_k|$ per question.

The pointer models process the state once per request (\cref{sec:inference}), whereas \drex Max processes it once per question.
For short states this difference is small: \drex Max answers a question in a median of 183\,ms (\cref{tab:efficiency}).
For many questions about a long document, the pointer models are cheaper.

\begin{table}[t]
\centering
\caption{\textbf{\drex Max training configuration} of the three training stages (\cref{sec:maxtrain}). $\lambda_\pi$ and $\lambda_s$ weight the consistency terms, and $\beta$, $\eta$, and $\rho$ are the coefficients of \cref{eq:reward,eq:sdrl}.}
\label{tab:maxhp}
\footnotesize
\setlength{\tabcolsep}{4pt}
\begin{tabularx}{\textwidth}{@{}>{\raggedright\arraybackslash}p{3.6cm}>{\raggedright\arraybackslash}X>{\raggedright\arraybackslash}X>{\raggedright\arraybackslash}X@{}}
\toprule
 & SFT (full parameters) & RLAF (low-rank) & SFT of the image encoder \\
\midrule
Trainable parameters & all backbone weights and the readout head & LoRA adapters (0.4B) and the readout head & image encoder and projector (0.6B) \\
Records / questions & 1.15M / 2.9M & 180k / 420k & 95k / 140k \\
Steps / batch (records) & 9{,}000 / 256 & 1{,}200 / 128 & 1{,}500 / 64 \\
Peak learning rate & $1{\times}10^{-5}$ (backbone), $1{\times}10^{-3}$ (readout) & $5{\times}10^{-5}$ (adapters), $1{\times}10^{-4}$ (readout) & $2{\times}10^{-5}$ \\
Schedule & linear warm-up (3\%), cosine decay & constant after warm-up & linear warm-up (3\%), cosine decay \\
$\lambda_\pi$ / $\lambda_s$ (consistency) & 0.05 / 1.0 & 0.05 / 1.0 & --- \\
$\beta$ / $\eta$ / $\rho$ (\cref{eq:sdrl}) & --- & 0.1 / 0.05 / 0.2 & --- \\
Judge & --- & Qwen3.8-27B at high reasoning effort; $q^{J}$ from the normalized probabilities of the option codes after the trace & --- \\
LoRA rank / $\alpha$ & --- & 64 / 128 & --- \\
Max sequence length & 16{,}384 & 16{,}384 & 16{,}384 (incl.\ image tokens) \\
\bottomrule
\end{tabularx}
\end{table}

\section{Model family}
\label{sec:family}

\begin{table}[t]
\centering
\caption{\textbf{The Drex family.} All four models accept the same request format and return the same output format. DLM, v1.5, and v2.0 each have their own pointer head with 256-dimensional query and key projections (\cref{sec:arch}), and \drex Max uses the readout head of \cref{sec:max}. Index values are chance-corrected Decision Index skill (\cref{sec:setup}): the public index and the full score on the 0.4.1 leaderboard and, for \drex DLM, the index of version 0.2, which is not comparable with 0.4.1.}
\label{tab:family}
\footnotesize
\setlength{\tabcolsep}{3pt}
\resizebox{\linewidth}{!}{%
\begin{tabular}{@{}lcccc@{}}
\toprule
 & \textbf{\drex DLM} & \textbf{\drex v1.5} & \textbf{\drex v2.0} & \textbf{\drex Max} \\
\midrule
Backbone & Efficient-DLM-8B & MiMo-V2.6-Distill-Qwen-9B & Qwen3.6-35B-A3B & Qwen3.8-27B \\
Layer types & block-bidir.\ diffusion & 3:1 DeltaNet : full attn. & 3:1 hybrid, MoE MLP & 3:1 DeltaNet : full attn. \\
Layers / hidden size $d$ & --- & 32 / 4096 & 40 / 2048 & 64 / 5120 \\
Experts (routed / total) & --- & --- & 8 / 256 & --- (dense) \\
Parameters (active / total) & 8B / 8B & 9B / 9B & 3B / 35B & 27B / 27B \\
Decision head & 256-d pointer & 256-d pointer & 256-d pointer & 255-way readout head \\
Head parameters & --- & 2.1M & 1.0M & 1.3M \\
Inputs & text & text & text & text, images \\
State attention & bidirectional & causal (question sequences) & causal (question sequences) & non-causal full-attn.\ layers \\
State shared across questions & yes & yes & yes & no (\cref{sec:maxcost}) \\
Context (open-source default / max) & 16k / 32k & 16k / 131k & 16k / 131k & 16k / 262k \\
Weights (BF16 / Q8\_0 GGUF) & $\sim$16\,GB / 8.7\,GB & $\sim$18\,GB / 9.5\,GB & $\sim$70\,GB / 36.9\,GB & $\sim$49\,GB / 27.3\,GB \\
Inference engines & Py, llama.cpp, Ollama & Py, llama.cpp, Ollama, API & Py, \kev, vLLM, llama.cpp, Ollama, API & Py, llama.cpp, Ollama, API \\
License & CC BY-NC 4.0 & Open RAIL-M & Open RAIL-M & Open RAIL-M (modified) \\
\midrule
Decision Index 0.4.1, public / full & --- (52.31 on 0.2) & 58.08 / 45.78 & 61.25 / 52.76 & 67.52 / 66.60 \\
\bottomrule
\end{tabular}}
\end{table}

\paragraph{\drex DLM.}
The backbone is NVIDIA Efficient-DLM-8B~\citep{fu2025efficientdlm}, an autoregressive language model that was converted into a diffusion language model.
It splits the sequence into blocks of consecutive tokens, with causal attention across blocks and bidirectional attention within each block, and it is trained with position-dependent masking.
We attach the pointer head, enable bidirectional attention within the state (\texttt{state\_bidir}, \cref{sec:attn}), and fine-tune both the head and the backbone~\citep{nace2026drexdlmcard}.
Since its backbone consists only of attention layers, all questions of a request are processed in a single forward pass over the packed sequence (\cref{sec:packing}).
It is released under CC BY-NC 4.0~\citep{rv_ccbync40}, the license of its backbone.

\paragraph{\drex v1.5.}
\drex v1.5 is the smallest model of the family whose weights are released under a license that permits commercial use (Open RAIL-M~\citep{rv_naceopenrailm})~\citep{nace2026drex15card}.
Its backbone, MiMo-V2.6-Distill-Qwen-9B, is a 32-layer Qwen3.5 hybrid with 8.95B parameters and three Gated DeltaNet layers for each full-attention layer.
Xiaomi MiMo trained it by distillation~\citep{mimo2026qwen9b,xiaomi2026mimov26,xiaomi2025mimo,yang2025qwen3}.

\paragraph{\drex v2.0.}
\drex v2.0 is a larger successor of v1.5, and its weights are also openly released~\citep{nace2026drexv2card}.
Its backbone is the language model of Qwen3.6-35B-A3B, a vision-language mixture-of-experts model~\citep{qwen36moe,yang2025qwen3}.
We discard the image encoder and the output layer that predicts the next token (the LM head).
It has 40 layers with the same 3:1 DeltaNet/full-attention pattern as v1.5 and MoE feed-forward blocks that route each token to 8 of 256 experts, so about 3B of 35B parameters are active per token.

\paragraph{\drex Max.}
\drex Max is the largest member of the family and the only one that accepts images as input (\cref{sec:max})~\citep{nace2026drexmaxcard}.
It differs from the other models in three ways: it uses a readout head instead of a pointer head, it removes the causal mask in its full-attention layers, and it processes each question as a separate sequence, so the state is not cached and shared across questions (\cref{sec:maxcost}).
Its request and output formats are the same as those of the other models, so a request written for v1.5 is also valid for Max.

\section{Training data and specialization stages}

The text training mix of \drex v1.5 (\cref{sec:universe}) has 1{,}150{,}460 records.
Its largest domains are Language \& reading (18.3\%) and Tools, routing \& home (15.7\%) (\cref{tab:mix}).

\begin{table}[t]
\centering
\caption{\textbf{Domains of the \drex v1.5 training mix}, with the number of records in each domain and examples of the decision tasks that it contains.}
\label{tab:mix}
\small
\begin{tabular}{@{}lrrl@{}}
\toprule
Domain & Records & Share & Example tasks \\
\midrule
Language \& reading & 210{,}124 & 18.3\% & NLI, paraphrase, sarcasm, contracts \\
Tools, routing \& home & 181{,}004 & 15.7\% & when to call, tool + arguments, side-effect checks \\
Knowledge \& professions & 133{,}288 & 11.6\% & exams, clinical, law, accounting \\
Code, math \& science & 117{,}196 & 10.2\% & program output, SQL, causal graphs \\
Retrieval \& grounding & 112{,}600 & 9.8\% & claim verification, evidence sufficiency, multi-hop \\
Finance \& operations & 99{,}968 & 8.7\% & filing tables, banking intents, chargebacks \\
Policy, legal \& safety & 99{,}900 & 8.7\% & rule selection, allow/deny, phishing, PII \\
Commerce \& support & 61{,}600 & 5.4\% & query--product relevance, return vs.\ exchange \\
Judgment \& uncertainty & 58{,}708 & 5.1\% & deferral, two acceptable answers, rubrics \\
Taste, humor \& media & 52{,}672 & 4.6\% & caption matching, headline edits, next chord \\
Games \& spatial & 23{,}400 & 2.0\% & piece values, legal moves, Tetris placement \\
\midrule
Total & 1{,}150{,}460 & 100\% & \\
\bottomrule
\end{tabular}
\end{table}

\subsection{Specialization stages}
\label{sec:s2e}

Our training code also supports specialization training stages, which continue training on a narrow family of tasks (\cref{sec:objective}).
A specialization stage trains low-rank adapters~\citep{hu2022lora}, starting from the head of the previous checkpoint.
Its training data consist of decisions whose correct answers are computed by a solver and of hard examples mined from existing data, mixed with replayed examples from earlier stages.
A KL penalty keeps the model close to the previous checkpoint, and the final checkpoint of the stage is selected on a held-out split of its training data, not on any benchmark.

\paragraph{\drex DLM and v2.0.}
The model card of \drex DLM~\citep{nace2026drexdlmcard} does provide more details on the training recipe.
\drex v2.0 is trained with the \kev training code~\citep{kev2026} in three stages: supervised fine-tuning of the MoE language model with a pointer head, a specialization stage, and an RLAF stage.
We analyze its checkpoints in \cref{sec:dynamics}.
Its model card~\citep{nace2026drexv2card} released on huggingface includes training recipe..

\section{Evaluation details}
\label{app:evaldetails}

\paragraph{Decision Index versions.}
Version 0.2.1 contains 38 benchmarks across five domains: Knowledge (10), Language (10), Retrieval (6), Tools (5), and Arts (7).
Each domain is weighted in proportion to the square root of its number of benchmarks, except Arts, whose weight is fixed at 10\%.
Version 0.4.1 removes ForecastBench, rebuilds GSM8K, and moves WinoGrande to a different domain.
Its public index averages 37 benchmarks with fixed domain weights: Knowledge 25.8\%, Language 25.8\%, Retrieval 20.0\%, Tools 18.3\%, and Arts 10\%.
These weights are rounded to one decimal, so the printed values sum to 99.9\% although the weights sum to 100\%.
Its \emph{Full} score adds private items, run by the maintainers of the leaderboard, that test either the same skills as the public benchmarks or new domains.
Where a model has both a published public-index score (reported on the 0.4.1 leaderboard) and our rerun of the public suite, the two differ by at most 0.1 point on the Decision Index (61.25 vs.\ 61.24 for v2.0, 58.08 vs.\ 57.98 for v1.5), consistent with the variation between repeated evaluations of the same model.

\paragraph{Other evaluations.}
\emph{JevBench} has 231 public items in three difficulty tiers: easy (48 items), standard (72), and hard (111).
A \texttt{choice} question counts as correct if the option with the highest probability is correct, a \texttt{noul} question if thresholding $p(\text{yes})$ at 0.5 gives the correct answer, and a \texttt{score} question if the level with the highest probability is correct.
\emph{DecideBench v1.1} has 400 decision questions in 200 minimal pairs (\cref{sec:otherbench}) from eight task families, and we report accuracy both per item and per pair.
Our results on it were not submitted to its official leaderboard.
The \emph{game arena} consists of eight games from OpenSpiel~\citep{openspiel}, played with the evaluation code of Kaggle Game Arena.
Each game type is played from 16 fixed openings of two moves (one per player), once with each color, which gives 32 games per game type.
Every legal move is an option, and the model chooses each move with a single forward pass, without searching over future moves.
If the model fails to return a valid move, a random legal move is played instead, and the failure is counted as an error.
The score is $(\text{wins} + \nicefrac{1}{2}\,\text{draws}) / \text{games}$.
In the \emph{OSWorld judging probe}, a decision model receives the text transcript of a finished computer-use episode from OSWorld~\citep{xie2024osworld} and predicts whether the official OSWorld checker gave the episode full credit.
We use 358 episodes from OSWorld-Verified and 105 from OSWorld-V2, taken from publicly released trajectories, and report the AUROC and the Brier score of the predictions, with the verdict of the checker as the label.
\emph{Decision Index vision} consists of the 11 public vision benchmarks of the Decision Index (12{,}210 items), scored with the same chance-corrected skill as the other Decision Index benchmarks.
The \emph{long-context} evaluation asks questions about documents of 8k to 128k tokens.

\section{Additional results}

\subsection{Per-benchmark change from \drex v2.0 to \drex Max}

\begin{figure}[!htbp]
\centering
\includegraphics[width=\textwidth]{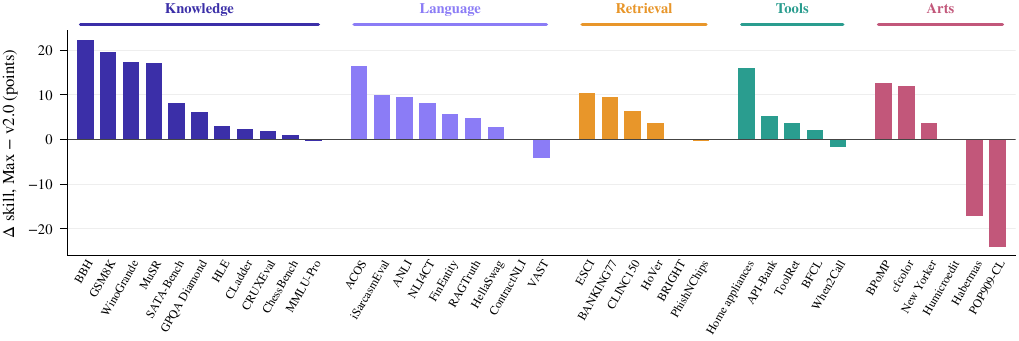}
\caption{\textbf{Per-benchmark change from \drex v2.0 to \drex Max} on Decision Index 0.4.1 (chance-corrected skill, our evaluations), grouped by domain and sorted within each domain. Max scores higher on \DIMaxWins{} of the \DIBenchN{} benchmarks, and its only two large drops are on POP909-CL and Habermas Machine.}
\label{fig:deltamax}
\end{figure}

In our evaluations of Decision Index 0.4.1, \drex Max scores higher than \drex v2.0 on \DIMaxWins{} of the \DIBenchN{} benchmarks of the public index.
Its only two large drops are on POP909-CL and Habermas Machine (\cref{fig:deltamax}; discussed in \cref{sec:di032}).
\Cref{app:bench032} gives the full per-benchmark scores.

\subsection{Decision Bench: paired comparisons and categories}

\begin{figure}[!htbp]
\centering
\begin{minipage}[c]{0.49\textwidth}
\centering
\includegraphics[width=\textwidth]{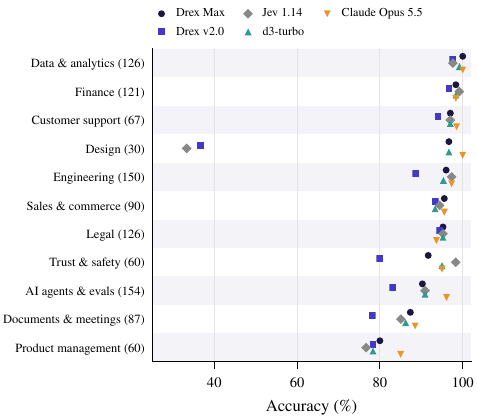}
\end{minipage}\hfill
\begin{minipage}[c]{0.48\textwidth}
\caption{\textbf{Decision Bench v4 by category}, sorted by \drex Max accuracy (images in image encoder); category sizes in parentheses. The frontier models' advantage is concentrated in Product management and AI agents \& evals, and \drex Max's largest deficit is Trust \& safety.}
\label{fig:dbcats}
\end{minipage}
\end{figure}

On all rows of Decision Bench v4, \drex Max, which receives the images with image encoder, is 6.2, 7.2, and 1.8 accuracy points ahead of v2.0, v1.5, and Jev 1.14, respectively.
Its accuracy is not significantly different from that of d3-turbo and GPT-6 Sol, and it is 1.6 points behind Claude Opus 5.5 ($p=0.046$) (\cref{tab:dbenchpaired}; see \cref{sec:dbench}).
\Cref{fig:dbcats} shows the accuracy by category, and \cref{app:dbench} gives per-category and per-task tables.

\begin{table}[!htbp]
\centering
\caption{\textbf{Paired comparisons on Decision Bench v4}: accuracy of \drex Max (images in image encoder) minus that of each other model, in points. The discordant counts are the number of rows that Max answers correctly and the other model incorrectly, followed by the number for the reverse case. The 95\% intervals are bootstrap intervals over rows, and $p$ is from an exact McNemar test.}
\label{tab:dbenchpaired}
\small
\setlength{\tabcolsep}{4pt}
\begin{tabular}{@{}lccccc@{}}
\toprule
 & \multicolumn{3}{c}{All 1{,}071 rows} & \multicolumn{2}{c}{949 text-only rows} \\
\cmidrule(lr){2-4}\cmidrule(lr){5-6}
vs. & $\Delta$ [95\% CI] & Discordant & $p$ & $\Delta$ [95\% CI] & $p$ \\
\midrule
\drex v2.0$^{\dagger}$ & +6.2 [+4.4, +7.9] & 81 / 15 & $<$0.001 & +5.1 [+3.3, +6.8] & $<$0.001 \\
\drex v1.5$^{\dagger}$ & +7.2 [+5.2, +9.2] & 99 / 22 & $<$0.001 & +6.4 [+4.3, +8.5] & $<$0.001 \\
Jev 1.14$^{\dagger}$ & +1.8 [+0.2, +3.5] & 51 / 32 & 0.048 & +0.0 [-1.6, +1.7] & 1.000 \\
d3-turbo & +0.3 [-0.7, +1.3] & 16 / 13 & 0.711 & +0.3 [-0.7, +1.5] & 0.711 \\
GPT-6 Sol & -0.7 [-2.2, +0.9] & 36 / 43 & 0.500 & -0.6 [-2.4, +1.2] & 0.572 \\
Claude Opus 5.5 & -1.6 [-3.1, -0.1] & 24 / 41 & 0.046 & -1.7 [-3.3, +0.0] & 0.060 \\
\bottomrule
\end{tabular}
\end{table}

\subsection{JevBench, DecideBench, and sequential decisions}
\label{sec:otherbench}

\begin{table}[!htbp]
\centering
\caption{\textbf{JevBench-231} (our evaluations). Overall accuracy (\%) with its 95\% Wilson interval, accuracy on each difficulty tier (number of items in parentheses), and the $p$-value of an exact McNemar test vs.\ \drex Max.}
\label{tab:jev}
\small
\setlength{\tabcolsep}{4pt}
\begin{tabular}{@{}lcccccc@{}}
\toprule
Model & Accuracy & 95\% CI & Easy (48) & Standard (72) & Hard (111) & $p$ vs.\ Max \\
\midrule
\drex Max & 91.8 (212/231) & [87.5, 94.7] & 100.0 & 100.0 & 82.9 & --- \\
\drex v2.0 & 86.1 (199/231) & [81.1, 90.0] & 100.0 & 95.8 & 73.9 & 0.002 \\
\drex v1.5 & 84.8 (196/231) & [79.7, 88.9] & 100.0 & 94.4 & 72.1 & $<$0.001 \\
d3-turbo & 90.9 (210/231) & [86.5, 94.0] & 100.0 & 100.0 & 81.1 & 0.791 \\
GPT-6 Luna Decisions & 91.3 (211/231) & [87.0, 94.3] & 100.0 & 98.6 & 82.9 & 1.000 \\
GPT-6 Sol & 99.1 (229/231) & [96.9, 99.8] & 100.0 & 100.0 & 98.2 & $<$0.001 \\
Claude Opus 5.5 & 100.0 (231/231) & [98.4, 100.0] & 100.0 & 100.0 & 100.0 & $<$0.001 \\
\bottomrule
\end{tabular}
\end{table}

\begin{table}[!htbp]
\centering
\begin{minipage}[t]{0.56\textwidth}
\centering
\captionof{table}{\textbf{DecideBench v1.1} (400 items in 200 minimal pairs, i.e., pairs of near-identical questions whose correct answers differ). Item accuracy and pair accuracy (\%; a pair counts as correct only if both of its items are answered correctly), with Wilson intervals, and expected calibration error (ECE) computed from the probabilities that each model outputs.}
\label{tab:decide}
\small
\setlength{\tabcolsep}{3pt}
\begin{tabular}{@{}lccc@{}}
\toprule
Model & Item & Pair & ECE \\
\midrule
\drex Max & 98.5 [96.8, 99.3] & 97.0 [93.6, 98.6] & 1.1 \\
\drex v2.0 & 96.5 [94.2, 97.9] & 93.5 [89.2, 96.2] & 4.3 \\
\drex v1.5 & 90.0 [86.7, 92.6] & 80.5 [74.5, 85.4] & 3.2 \\
Jev & 98.0 [96.1, 99.0] & 96.0 [92.3, 98.0] & 1.8 \\
GPT-6 Sol & 99.2 [97.8, 99.7] & 98.5 [95.7, 99.5] & --- \\
Claude Opus 5.5 & 98.8 [97.1, 99.5] & 98.5 [95.7, 99.5] & --- \\
\bottomrule
\end{tabular}
\end{minipage}\hfill
\begin{minipage}[t]{0.41\textwidth}
\centering
\captionof{table}{\textbf{Game Bench} win rate (\%) over the eight games of the arena, as reported when \drex Max was launched, with Jev 1.14 as the reference model (50\%). We do not have the per-game records of these runs.}
\label{tab:gamebench}
\small
\begin{tabular}{@{}lc@{}}
\toprule
Model & Win rate \\
\midrule
\textbf{\drex Max} & \textbf{64.0} \\
\drex v2.0 & 60.3 \\
\drex v1.5 & 58.7 \\
GPT-6 Luna Decisions & 57.5 \\
Jev 1.14 (reference) & 50.0 \\
\midrule
GPT-6 Sol & 65.0 \\
Claude Opus 5.5 & 63.0 \\
\bottomrule
\end{tabular}
\end{minipage}
\end{table}

\paragraph{JevBench.}
\drex Max answers 212 of the 231 items correctly (91.8\%), which is significantly more than v2.0 and v1.5 (McNemar $p<0.01$), and its accuracy is not significantly different from that of GPT-6 Luna Decisions and d3-turbo (\cref{tab:jev}).
Most of its gain over v2.0 comes from the hard tier, where accuracy rises from 73.9\% to 82.9\% (ten more items answered correctly).
On the standard tier, Max answers three more items correctly than v2.0 (100\% vs.\ 95.8\%).
The easy tier does not separate the two models, since both answer all 48 of its items correctly.
The frontier chat models answer 229 to 231 of the 231 items correctly.
JevBench therefore separates the decision models from the frontier models, but it does not separate the strongest decision models from each other.

\paragraph{DecideBench.}
DecideBench consists of minimal pairs: two near-identical decision questions that differ in a single detail, which changes the correct answer.
A pair counts as correct only if the model answers both of its questions correctly, so pair accuracy measures whether a model changes its answer when that detail changes.
\drex Max reaches 97.0\% pair accuracy vs.\ 93.5\% for v2.0, 80.5\% for v1.5, and 96.0\% for Jev (\cref{tab:decide}).
The frontier models reach 98.5\%.
Max also has the lowest expected calibration error on this benchmark (1.1\%), although an ECE estimated from 400 items is noisy.

\paragraph{Sequential decisions.}
In the game arena (\cref{app:evaldetails}), the model chooses each move by answering a single \texttt{choice} question whose options are all legal moves, without searching over future moves.
We also ran the arena ourselves, with Jev 1.13.0 as the opponent, and scored each game as a win, a draw counting as half a win, or a loss (\cref{app:evaldetails}).
Over 256 games, \drex v1.5 scores 56.8\% [51.4, 62.1] (122 wins, 47 draws, and 87 losses), and v2.0 scores 60.9\% [55.5, 66.2] (132 wins, 48 draws, and 76 losses).
Both bootstrap intervals exclude 50\%.
\Cref{tab:gamebench} reports a separate evaluation that was run when \drex Max was launched, in a different arena in which Jev 1.14 is the reference model with a win rate of 50\%, so its numbers are not comparable with ours.
In that evaluation, \drex Max has the highest win rate among the decision models (64.0\%), within two points of GPT-6 Sol (65.0\%) and Claude Opus 5.5 (63.0\%), and it wins 58\% of its games when it plays directly against v2.0.
We do not have the per-game records of these runs, so we can neither report intervals for these numbers nor reproduce them.
Since each of the eight game types is played 32 times, the 95\% interval of the win rate on a single game type has a half-width of up to about 17 percentage points, so smaller differences on a single game type are within noise.

\subsection{Calibration and order breakdowns}

\begin{table}[!htbp]
\centering
\caption{\textbf{Calibration by Decision Index domain}, on the held-out half of the decisions (\cref{sec:calib-results}). For each model, the table gives the accuracy, the mean confidence (the probability of the top option), the ECE of the released model, and the ECE after rescaling with the single temperature $T^\star$ of \cref{eq:temp}, fitted on held-out data (all in \%).}
\label{tab:calibarea}
\scriptsize
\setlength{\tabcolsep}{2.6pt}
\begin{tabular}{@{}lr cccc cccc cccc@{}}
\toprule
 & & \multicolumn{4}{c}{\drex Max} & \multicolumn{4}{c}{\drex v2.0} & \multicolumn{4}{c}{\drex v1.5} \\
\cmidrule(lr){3-6}\cmidrule(lr){7-10}\cmidrule(lr){11-14}
Domain & $N$ & Acc. & Conf. & ECE & ECE$_{T^\star}$ & Acc. & Conf. & ECE & ECE$_{T^\star}$ & Acc. & Conf. & ECE & ECE$_{T^\star}$ \\
\midrule
Knowledge & 24{,}266 & 75.7 & 82.0 & 7.3 & 4.4 & 70.8 & 72.1 & 2.8 & 3.7 & 67.4 & 77.7 & 10.5 & 4.7 \\
Language & 64{,}364 & 95.9 & 96.4 & 0.5 & 1.3 & 95.4 & 92.2 & 3.2 & 2.4 & 95.8 & 97.0 & 1.2 & 0.9 \\
Retrieval & 12{,}087 & 81.6 & 89.8 & 8.2 & 4.6 & 76.8 & 79.0 & 3.3 & 4.2 & 77.7 & 88.1 & 10.6 & 5.0 \\
Tools & 5{,}816 & 85.9 & 95.0 & 9.2 & 7.0 & 85.7 & 90.7 & 5.0 & 6.0 & 85.0 & 95.1 & 10.1 & 6.8 \\
Arts & 8{,}453 & 71.8 & 77.2 & 7.2 & 6.3 & 71.5 & 66.6 & 5.0 & 3.3 & 68.4 & 69.2 & 2.9 & 6.4 \\
\bottomrule
\end{tabular}
\end{table}

\begin{table}[!htbp]
\centering
\caption{\textbf{Calibration by number of options $K$}, over all decisions of each run, with the probabilities of the released models. Accuracy, mean confidence (the probability of the top option), and ECE (all in \%).}
\label{tab:calibk}
\small
\setlength{\tabcolsep}{3.4pt}
\begin{tabular}{@{}lr ccc ccc ccc@{}}
\toprule
 & & \multicolumn{3}{c}{\drex Max} & \multicolumn{3}{c}{\drex v2.0} & \multicolumn{3}{c}{\drex v1.5} \\
\cmidrule(lr){3-5}\cmidrule(lr){6-8}\cmidrule(lr){9-11}
Options & $N$ & Acc. & Conf. & ECE & Acc. & Conf. & ECE & Acc. & Conf. & ECE \\
\midrule
$K=2$ & 163{,}148 & 92.0 & 95.4 & 3.4 & 90.4 & 89.9 & 0.6 & 90.7 & 94.6 & 3.9 \\
$3\le K\le 5$ & 36{,}953 & 84.0 & 87.1 & 3.2 & 81.5 & 78.2 & 3.4 & 80.3 & 84.2 & 3.8 \\
$6\le K\le 20$ & 15{,}155 & 71.7 & 80.1 & 8.4 & 67.8 & 66.0 & 2.0 & 56.9 & 70.5 & 13.6 \\
$K>20$ & 15{,}061 & 68.6 & 66.3 & 4.0 & 67.4 & 61.6 & 5.8 & 68.2 & 71.1 & 3.5 \\
\bottomrule
\end{tabular}
\end{table}

\begin{table}[!htbp]
\centering
\caption{\textbf{Order instability and order averaging} on the \OrdN{} decisions of \cref{tab:order}. The left columns show the calibration of the original order and of the average of the three orders' distributions (ECE in \%, log loss in nats, Brier), with a bootstrap 95\% interval for the reduction in log loss. The right columns show the share of decisions whose top option is the same in all three orders, the accuracy of stable and unstable decisions, and the accuracy of the same number of most confident decisions. The stratified sample over-weights small benchmarks, so its ECE is not comparable with the ECE in \cref{tab:calib}.}
\label{tab:ordertta}
\footnotesize
\setlength{\tabcolsep}{2.8pt}
\begin{tabular}{@{}lrrrrcrrrrrr@{}}
\toprule
 & \multicolumn{2}{c}{ECE} & \multicolumn{2}{c}{Log loss} & Reduction & \multicolumn{2}{c}{Brier} & Stable & \multicolumn{2}{c}{Accuracy} & Most conf. \\
\cmidrule(lr){2-3}\cmidrule(lr){4-5}\cmidrule(lr){7-8}\cmidrule(lr){10-11}
Model & orig. & avg. & orig. & avg. & (95\% CI) & orig. & avg. & (\%) & stable & unstable & same cov. \\
\midrule
\drex Max & 5.8 & 5.4 & 0.474 & 0.460 & +0.014 [+0.005, +0.024] & 0.237 & 0.232 & 92.6 & 87.2 & 37.6 & 87.1 \\
\drex v2.0 & 3.3 & 3.4 & 0.499 & 0.493 & +0.006 [-0.003, +0.016] & 0.274 & 0.271 & 90.2 & 84.7 & 46.8 & 84.6 \\
\drex v1.5 & 7.1 & 6.0 & 0.579 & 0.568 & +0.011 [-0.001, +0.024] & 0.301 & 0.296 & 90.8 & 82.9 & 37.7 & 83.3 \\
Jev 1.14 & 6.9 & 6.2 & 0.680 & 0.636 & +0.044 [+0.018, +0.075] & 0.300 & 0.296 & 93.0 & 81.8 & 45.9 & 82.0 \\
\bottomrule
\end{tabular}
\end{table}

\subsection{Vision suites}

We evaluate \drex Max on all items of the 11 public vision benchmarks of the Decision Index.
We also compare it with GPT-6 Luna Decisions on the same sample of 80 items per benchmark for both models (\cref{tab:vision}, discussed in \cref{sec:vision}).

\begin{table}[!htbp]
\centering
\caption{\textbf{Decision Index vision suites} (public benchmarks). Left: accuracy and chance-corrected skill of \drex Max on all 12{,}210 items. Right: accuracy of \drex Max and GPT-6 Luna Decisions on the same sample of 80 items per benchmark, with the better of the two in bold. $w$ is the weight of a benchmark in the public vision index; benchmarks without a $w$ have weight 1.}
\label{tab:vision}
\small
\setlength{\tabcolsep}{3.6pt}
\begin{tabular}{@{}llrcc cc@{}}
\toprule
 & & & \multicolumn{2}{c}{Full suite, \drex Max} & \multicolumn{2}{c}{80-item sample} \\
\cmidrule(lr){4-5}\cmidrule(lr){6-7}
Group & Benchmark & Items & Acc. & Skill & \drex Max & Luna Dec. \\
\midrule
Perception & CV-Bench & 2{,}038 & 86.3 & 77.3 & \textbf{85.0} & 73.8 \\
 & BLINK & 961 & 74.1 & 59.9 & \textbf{80.0} & 63.7 \\
 & RealWorldQA & 589 & 84.0 & 73.9 & \textbf{85.0} & 75.0 \\
 & Winoground & 800 & 93.0 & 86.0 & \textbf{93.8} & 83.8 \\
Documents & CharXiv & 532 & 86.1 & 81.5 & \textbf{86.2} & 78.8 \\
 & InfographicVQA & 244 & 98.8 & 98.4 & \textbf{97.5} & 93.8 \\
 & KIE (CORD + FUNSD) & 2{,}465 & 95.5 & 92.6 & \textbf{98.8} & 97.5 \\
Agentic & Mind2Web & 973 & 88.6 & 84.8 & \textbf{88.8} & 81.2 \\
Reasoning & R-Bench-M ($w{=}0.5$) & 665 & 48.9 & 38.6 & \textbf{51.2} & 32.5 \\
 & MMMU-Pro vision ($w{=}0.25$) & 1{,}730 & 53.6 & 46.8 & \textbf{53.8} & 41.2 \\
Safety & Hateful Memes & 1{,}213 & 70.6 & 41.1 & \textbf{75.0} & 71.2 \\
\midrule
\multicolumn{3}{@{}l}{Public vision index (weighted skill)} & \multicolumn{2}{c}{74.5$^{a}$} & \textbf{77.1} & 63.9 \\
\bottomrule
\multicolumn{7}{@{}p{0.95\textwidth}@{}}{\footnotesize $^{a}$Our full-suite run, in which Mind2Web screenshots were re-rendered with a substitute font; the published run reports 73.9.}
\end{tabular}
\end{table}

\subsection{Agentic judgments and knowledge}
\label{sec:agentic}

\begin{table}[!htbp]
\centering
\begin{minipage}[t]{0.50\textwidth}
\centering
\captionof{table}{\textbf{OSWorld judging probe.} Each model predicts, from the transcript of an episode, whether the OSWorld checker gave the episode full credit. We report the AUROC with its 95\% interval and the Brier score (lower is better). A heuristic that predicts success for shorter transcripts reaches an AUROC of 0.709 on OSWorld-Verified and 0.694 on OSWorld-V2.}
\label{tab:osworld}
\small
\setlength{\tabcolsep}{3pt}
\begin{tabular}{@{}lccc@{}}
\toprule
 & \multicolumn{2}{c}{Verified ($n{=}358$)} & V2 ($n{=}105$) \\
\cmidrule(lr){2-3}\cmidrule(lr){4-4}
Model & AUROC & Brier & AUROC \\
\midrule
\textbf{\drex Max} & \textbf{0.780} [0.71, 0.84] & \textbf{0.114} & \textbf{0.737} \\
Jev 1.13.0 & 0.734 [0.66, 0.80] & 0.240 & 0.687 \\
GPT-6 Luna & 0.709 [0.63, 0.78] & 0.148 & 0.653 \\
\drex v2.0 & 0.698 [0.62, 0.77] & 0.133 & 0.636 \\
\bottomrule
\end{tabular}
\end{minipage}\hfill
\begin{minipage}[t]{0.47\textwidth}
\centering
\captionof{table}{\textbf{Comparison with frontier chat models on the same sample}: accuracy (\%) on 100 requests per benchmark, sampled with a fixed seed so that all models answer the same requests. Requests that fail count as wrong.}
\label{tab:frontier}
\small
\setlength{\tabcolsep}{2.6pt}
\begin{tabular}{@{}lccccc@{}}
\toprule
 & Max & v2.0 & Luna & Sol & Opus \\
\midrule
MMLU-Pro & 73 & 74 & 60 & 93 & \textbf{94} \\
GPQA Diam. & 55 & 47 & 50 & \textbf{90} & 89 \\
HLE & 21 & 13 & 12 & 39 & \textbf{56} \\
CRUXEval & 83 & 88 & 65 & \textbf{100} & 95 \\
ARC-Chall. & \textbf{99} & 96 & 98 & \textbf{99} & 98 \\
HellaSwag & \textbf{96} & 95 & 91 & 93 & \textbf{96} \\
\midrule
BFCL$^{c}$ & \textbf{98.0} & 94.7 & 95.3 & 94.7 & 97.3 \\
RAGTruth$^{c}$ & 84.8 & 84.9 & 77.8 & 72.5 & 69.8 \\
When2Call$^{c}$ & 87.0 & 87.3 & 70.7 & 78.3 & 86.7 \\
\bottomrule
\multicolumn{6}{@{}p{0.98\linewidth}@{}}{\footnotesize $^{c}$Fixed-seed tool and hallucination sample (150 BFCL, 300 RAGTruth, 300 When2Call); v1.5 scores 96.0 / 86.0 / 90.7.}
\end{tabular}
\end{minipage}
\end{table}

\paragraph{Computer-use judging.}
None of these models operates a computer itself.
The probe instead asks whether a model can judge, from its transcript, whether a finished episode of a computer-use agent succeeded.
On OSWorld-Verified, \drex Max is the only model whose 95\% interval lies above the AUROC of the heuristic that predicts success for shorter transcripts (0.780 vs.\ 0.709).
Its AUROC is higher than that of v2.0 by 0.082 (bootstrap interval [0.019, 0.143]) and higher than that of Luna by 0.071 ([0.020, 0.119]).
Its lead over Jev (0.046, [$-0.015$, 0.106]) is not significant (\cref{tab:osworld}).
Its Brier score (0.114) is the lowest of all models.
Only \drex Max and v2.0 (0.133) have a lower Brier score than a constant forecast of the base rate of success (0.140).
On OSWorld-V2, the task instruction is not included in the transcript and only 29.5\% of the episodes receive full credit, and all models have a lower AUROC.

\section{Further limitations and broader impact}
\label{app:limits}

\paragraph{Confounds and evaluation scope.}
The systems on the Decision Index leaderboard differ in backbone, training data, training objective, and inference setup at the same time.
Differences between the rows of \cref{tab:di032} and the per-benchmark changes in \cref{fig:deltamax} should therefore not be attributed to any single factor.
The full scores are taken from the leaderboard.
We did not run the private tests ourselves, and we do not have the outputs of the models on individual private test items.
JevBench no longer separates the strongest decision models from each other (\cref{sec:otherbench}).
Decision Bench contains only 60 questions whose state consists of an image alone, which is a small sample for conclusions about image inputs, so our claims about images rest mainly on the vision suites (\cref{sec:vision}).
Benchmarks drawn from public datasets may overlap with the pre-training data of the teacher models despite our filters (\cref{sec:s2e}).

\paragraph{Inference efficiency.}
On \drex v1.5, a request with an uncached state at the maximum length of 131{,}072 tokens takes about 2.7\,s, which exceeds our target of 2\,s.
The cost of \drex Max on a long state grows with the number of questions, since it processes the state once per question (\cref{sec:maxcost}).
We check the accuracy after each change to the inference code separately.
(\cref{tab:ledger}).
Output probabilities can differ slightly when the same request is processed in a batch with different other requests, because the low-precision kernels choose their tiling by batch shape.

\paragraph{Broader impact and responsible use.}
A decision model is designed so that an application can act on its output automatically, without a human in the loop.
By construction, it can only return one of the options that the user supplied.
Its calibration and its robustness to option order, in contrast, are empirical properties: we measure them on benchmark data, and they can degrade when the inputs differ from that data.
Image inputs give attackers additional ways to inject adversarial content, since the model also processes text that is rendered inside an image.
Users should therefore treat a state that contains images as untrusted input, just as they would a state that contains text.
The probabilities are reproducible only up to the numerical variation reported in \cref{sec:serving}.
A decision whose probability is close to its threshold can therefore change between runs, and a decision with many options can change when the options are reordered.
Deployments should monitor confidence and accuracy on live traffic.
They should also keep a margin around decision thresholds, send low-confidence decisions to human review, and log the request, the output distribution, and the model version as the record of each decision.
The model should not be the sole basis for consequential decisions about people.

\section{Per-benchmark results on Decision Index 0.4.1}
\label{app:bench032}

In \cref{tab:fullbench032}, the iSarcasmEval score is the mean over its subtask tracks.

\begin{table}[h]
\centering
\caption{\textbf{Decision Index 0.4.1, all public index benchmarks} (official metric $\times 100$, our evaluations). Best per row in bold.}
\label{tab:fullbench032}
\scriptsize
\setlength{\tabcolsep}{3.2pt}
\renewcommand{\arraystretch}{1.02}
\begin{tabular}{@{}llrrrrr@{}}
\toprule
Benchmark & Metric & Requests & \drex Max & \drex v2.0 & \drex v1.5 & Luna Dec. \\
\midrule
\multicolumn{7}{@{}l}{\emph{Knowledge}} \\
BBH fixed-option tasks & acc. & 5507 & \textbf{93.7} & 78.3 & 85.6 & 66.0 \\
CLadder & acc. & 5000 & 85.7 & 84.5 & \textbf{86.1} & 63.5 \\
CRUXEval & acc. & 570 & \textbf{85.8} & 84.6 & 76.7 & 66.3 \\
ChessBench & acc. & 5000 & \textbf{24.2} & 23.4 & 22.0 & 17.4 \\
GPQA Diamond & acc. & 196 & \textbf{56.1} & 51.5 & 43.9 & 48.5 \\
GSM8K & acc. & 2638 & \textbf{83.2} & 66.8 & 55.0 & 36.3 \\
HLE & acc. & 501 & \textbf{19.0} & 12.8 & 11.4 & 12.2 \\
MMLU-Pro & acc. & 12032 & 72.5 & \textbf{72.8} & 58.6 & 57.8 \\
MuSR & acc. & 752 & \textbf{70.3} & 59.6 & 56.9 & 56.8 \\
SATA-Bench & case exact & 1650 & \textbf{32.2} & 24.2 & 26.9 & 31.0 \\
WinoGrande & acc. & 1267 & \textbf{94.8} & 86.1 & 87.8 & 73.4 \\
\midrule
\multicolumn{7}{@{}l}{\emph{Language}} \\
ACOS & per-review F1 & 1565 & \textbf{39.8} & 23.9 & 5.3 & 15.1 \\
ANLI & macro-F1 & 3200 & \textbf{76.8} & 70.5 & 70.6 & 67.4 \\
ContractNLI & macro-F1 & 123 & 83.2 & \textbf{83.3} & 83.1 & 77.1 \\
FinEntity & macro-F1 & 979 & \textbf{94.7} & 90.8 & 88.0 & 82.7 \\
HellaSwag & acc. & 10042 & \textbf{97.9} & 95.9 & 92.1 & 91.2 \\
NLI4CT & macro-F1 & 5500 & \textbf{86.0} & 81.9 & 80.2 & 78.4 \\
RAGTruth response-level hallucination & F1 on hallucinated class & 2700 & \textbf{84.8} & 82.5 & 82.7 & 77.0 \\
VAST & macro-F1 & 3006 & 75.6 & \textbf{78.4} & 77.2 & 58.0 \\
iSarcasmEval & track mean & 4600 & \textbf{71.4} & 63.7 & 55.1 & 37.2 \\
\midrule
\multicolumn{7}{@{}l}{\emph{Retrieval}} \\
Amazon ESCI & macro-F1 & 5000 & \textbf{62.9} & 54.6 & 55.8 & 54.7 \\
BANKING77 & macro-F1 & 3080 & \textbf{90.0} & 80.5 & 85.5 & 77.5 \\
BRIGHT & nDCG@10 & 220 & 47.1 & \textbf{47.2} & 43.0 & 40.1 \\
CLINC150+OOS & macro-F1 & 5500 & \textbf{95.3} & 89.1 & 91.7 & 88.9 \\
HoVer claim verification & acc. & 4000 & \textbf{87.6} & 85.8 & 86.2 & 71.4 \\
PhishNChips phishing decisions & acc. & 2000 & 74.8 & \textbf{75.0} & 66.6 & 51.5 \\
\midrule
\multicolumn{7}{@{}l}{\emph{Tools}} \\
API-Bank & acc. & 508 & \textbf{88.4} & 83.3 & 85.4 & 84.5 \\
BFCL & case exact & 1694 & \textbf{98.2} & 96.6 & 94.9 & 93.9 \\
Home appliance simulator & case exact & 88 & \textbf{70.5} & 54.5 & 38.6 & 43.2 \\
ToolRet & nDCG@10 & 685 & \textbf{70.6} & 67.3 & 70.2 & 58.8 \\
When2Call MCQ & acc. & 3652 & 85.2 & 86.5 & \textbf{89.7} & 70.2 \\
\midrule
\multicolumn{7}{@{}l}{\emph{Arts}} \\
BPoMP & acc. & 5000 & \textbf{95.4} & 89.1 & 81.5 & 90.8 \\
Habermas Machine & acc. & 1676 & 50.5 & \textbf{62.4} & 59.6 & 42.7 \\
Humicroedit & acc. & 2628 & \textbf{63.5} & \textbf{63.5} & 62.7 & 62.0 \\
New Yorker caption matching & acc. & 528 & \textbf{80.7} & 77.6 & 73.7 & 68.2 \\
POP909-CL & acc. & 2000 & 55.6 & \textbf{79.5} & 76.3 & 18.4 \\
cfcolor & acc. & 5000 & \textbf{66.6} & 60.6 & 59.5 & 61.1 \\
\bottomrule
\end{tabular}
\end{table}

\section{Decision Bench v4 by category and task}
\label{app:dbench}

\begin{table}[H]
\centering
\caption{\textbf{Decision Bench v4 accuracy by category} (\%). \drex Max receives the images, whereas Max (text) receives the same textual version of the images as the text-only models.}
\label{tab:dbcat}
\scriptsize
\setlength{\tabcolsep}{3pt}
\begin{tabular}{@{}lrcccccccc@{}}
\toprule
Category & $N$ & \drex Max & Max (text) & \drex v2.0 & \drex v1.5 & Jev 1.14 & d3-turbo & GPT-6 Sol & Opus 5.5 \\
\midrule
AI agents \& evals & 154 & 90.3 & 90.3 & 83.1 & 82.5 & 90.9 & 90.9 & 92.2 & \textbf{96.1} \\
Customer support & 67 & 97.0 & 97.0 & 94.0 & 95.5 & 97.0 & 97.0 & \textbf{98.5} & \textbf{98.5} \\
Data \& analytics & 126 & \textbf{100.0} & \textbf{100.0} & 97.6 & 97.6 & 97.6 & 99.2 & 99.2 & \textbf{100.0} \\
Design & 30 & 96.7 & 26.7 & 36.7 & 46.7 & 33.3 & 96.7 & \textbf{100.0} & \textbf{100.0} \\
Documents \& meetings & 87 & 87.4 & 87.4 & 78.2 & 80.5 & 85.1 & 86.2 & \textbf{89.7} & 88.5 \\
Engineering & 150 & 96.0 & 96.0 & 88.7 & 85.3 & \textbf{97.3} & 95.3 & 94.7 & \textbf{97.3} \\
Finance & 121 & 98.3 & 98.3 & 96.7 & \textbf{99.2} & \textbf{99.2} & 98.3 & 98.3 & 98.3 \\
Legal & 126 & \textbf{95.2} & \textbf{95.2} & 94.4 & 88.9 & \textbf{95.2} & \textbf{95.2} & 92.1 & 93.7 \\
Product management & 60 & 80.0 & 80.0 & 78.3 & 76.7 & 76.7 & 78.3 & \textbf{85.0} & \textbf{85.0} \\
Sales \& commerce & 90 & 95.6 & 95.6 & 93.3 & 86.7 & 94.4 & 93.3 & \textbf{96.7} & 95.6 \\
Trust \& safety & 60 & 91.7 & 91.7 & 80.0 & 80.0 & \textbf{98.3} & 95.0 & 96.7 & 95.0 \\
\bottomrule
\end{tabular}
\end{table}

\section{Order-robustness protocol}
\label{app:order}

\paragraph{Sampling.}
From the 37 benchmarks of the public index of Decision Index 0.4.1, we kept the requests in which every question is a \texttt{choice} question and at least one gold label is a string.
Of these, we kept only the requests with at most 16 questions and a state of at most 12{,}000 characters when serialized as JSON.
Within each benchmark, we shuffled the requests with the benchmark ID as the seed and kept the first 40 (or all of them, if a benchmark has fewer), which gives 1{,}175 requests.
\paragraph{Variants.}
For every question with at least two options, we built two variants: one with the options in reverse order and one with a random permutation of the options, seeded by the request ID and the question key.
The instruction, the state, and the option texts are unchanged, and only the order of the options in the \texttt{criteria} field of the request changes.
\paragraph{Requests.}
On 26 October 2026, we sent each variant of each request to \texttt{drex-max}, \texttt{drex-v2.0}, and \texttt{drex-v1.5} at \texttt{drex.nace.ai} and to \texttt{jev-1.14.0} at \texttt{api.typesafe.ai}, with up to eight concurrent requests per model.
Each model answered each variant once: a request that failed with a transient error was resent, with increasing delays, until it received an answer.
\paragraph{Exclusions.}
The public APIs reject requests whose options are structured objects rather than strings (e.g., chord or preference options with several fields), since such requests fail their schema validation (HTTP 422).
The evaluation code of the Decision Index converts such options into strings before sending them, but we sent the requests unchanged, so these requests are excluded.
Jev also rejected some requests, which we excluded as well.
Any request not answered in all three orders by all four models is excluded for every model, leaving \OrdReq{} requests and \OrdN{} decisions.
\paragraph{Scoring.}
We renormalize the predicted probabilities over the supplied options, map them back to the original order of the options, and compare them as in \cref{sec:problem}.
Bootstrap intervals are computed by resampling decisions, with 2{,}000 resamples.

\section{Document search}
\label{sec:search}

In this section, we present an application of \drex Max within our enterprise document search framework, which answers natural-language questions over a workspace of a customer's documents, spreadsheets, and tables.
To answer a search query, the framework must decide which search strategy to use, which files to access, and whether the text it has already retrieved is enough to answer the query without accessing more files.
\drex Max makes these decisions, and a separate language model, which we call the \emph{answering model}, writes the final answer from the content that the framework selects.
Each decision is a single \texttt{noul} question (\cref{sec:problem}): \drex Max returns the probability that the answer is yes, and the framework compares this probability with a threshold that we fitted for that type of decision.
\drex Max makes four types of decisions, and we describe each type in its own paragraph below:
\begin{itemize}[leftmargin=1.3em, itemsep=1pt, topsep=2pt]
  \item \emph{Query routing} decides whether a query is answered by a single retrieval pass or by a multi-step search agent.
  \item \emph{File relevance} decides which of the retrieved files are most likely to contain the information that the query asks for.
  \item \emph{Table relevance} decides which of the retrieved tables the framework can skip.
  \item \emph{Passage sufficiency} decides whether the passages already retrieved from a file contain the complete answer.
\end{itemize}
\Cref{tab:search} summarizes the four types.

\begin{table}[t]
\centering
\caption{\textbf{The four types of decisions that \drex Max makes in our document search framework.} Each row describes one type of decision. For each item in the second column, the framework asks \drex Max one \texttt{noul} question, and $p$ denotes the probability that \drex Max assigns to the answer yes. The third column states what an answer of yes means, and the fourth column states how the framework uses $p$.}
\label{tab:search}
\footnotesize
\renewcommand{\arraystretch}{1.5}
\setlength{\tabcolsep}{3pt}
\begin{tabularx}{\textwidth}{@{}>{\raggedright\arraybackslash}p{1.9cm}>{\raggedright\arraybackslash}X>{\raggedright\arraybackslash}X>{\raggedright\arraybackslash}X@{}}
\toprule
Decision type & Description of input items & Interpretation of positive response & Use of $p$ by the framework \\
\midrule
Query routing & Each search query for which the caller lets the framework choose the search strategy. & Answering the query requires a lookup of one specific fact, rather than an investigation that combines facts across many documents or table rows. & If $1 - p \ge 0.80$, the framework passes the query to the search agent; otherwise, it answers the query with fact search. \\
File relevance & Each file on the shortlist, which holds the 12 files ranked highest by the retrieval system. & The file states the information that the query asks for. & The framework sorts the shortlist by $p$ plus a small bonus for a high retrieval rank. \\
Table relevance & Each table on the shortlist. & The rows of the table contain the information that the query asks for, or can be used to compute it. & If $p$ is below a fitted threshold, the framework removes the table from the shortlist, unless the table has the highest $p$ on the shortlist. \\
Passage sufficiency & Each file on the shortlist. & The passages and graph facts that retrieval returned for the file contain the complete answer. & If $p$ is above a high threshold for any file, the framework fetches no files, and the answering model writes the answer from the passages of those files. \\
\bottomrule
\end{tabularx}
\end{table}

\paragraph{Routing search queries.}
The search framework can answer a query with one of two strategies.
In \emph{fact search}, the framework retrieves the files most relevant to the query, fetches their relevant content, and passes this content to the answering model in a single pass.
In the second strategy, the framework passes the query to a \emph{search agent}, which is a language model that works in an interactive loop, as in ReAct: at each step, the agent issues a command against the workspace, such as listing a folder, reading a document, or running an SQL query over a table, observes the output, and chooses its next command, until it can answer.
The search agent can answer queries that fact search cannot, but at a much higher cost in time and computation.
The caller of the search API chooses the strategy by setting a \emph{mode}: The caller can specify the strategy directly by setting the mode to either \texttt{fact} or \texttt{agent}. If the provided mode is set to \texttt{auto}, the framework automatically chooses the strategy by asking \drex Max.
In this case, the state that \drex receives contains only the query, the caller's optional context, and the folder to which the search is restricted, and never the content of the workspace.
\drex is asked whether answering the query requires a \emph{lookup} or an \emph{investigation}.
A lookup finds one specific fact, such as a value, a date, or a clause, that are mentioned in one or few passages.
An investigation must list, count, rank, or compare facts across many documents or table rows, check these facts against each other for consistency, or chain several dependent steps.
The framework passes the query to the search agent when the probability of an investigation is at least 0.80. Otherwise, the query is passed to fact search.
On a held-out split of 662 labeled search queries, this results in 93.8\% of search queries being routed correctly (AUROC 0.966).

\paragraph{Screening retrieved files and tables for relevance.}
Fact search begins with the framework's retrieval system, which ranks the files of the workspace for the query by combining keyword (BM25) search, embedding search over chunks of file text, and, in workspaces that have one, a knowledge graph built from the workspace's documents.
The 12 highest-ranked files form the \emph{shortlist}, which is the set of files whose content fact search may pass to the answering model.
The framework asks \drex about each file on the shortlist in a separate request.
The state of each request contains the query and a description of one file: its path, the pages that matched the query, and the beginning of each text chunk of the file that matched.
\drex is asked whether this file is the record that states the information the query asks for, such as the document, statement, or table that contains the requested value.
The framework then sorts the shortlist by \drex's probability plus a small bonus that is larger for files that retrieval ranked higher, so that files with nearly equal probabilities keep the order that retrieval gave them.

The framework also uses these probabilities to remove tables from the shortlist.
Tables, such as spreadsheets and CSV files, are not passed to the answering model as text.
To use a table, the framework asks a language model to write SQL queries over it, executes these queries, and passes their results to the answering model.
This process takes tens of seconds per table, whereas fetching the text of a document takes under a second.
The framework therefore removes a table from the shortlist when \drex's probability for it is below a fitted threshold, unless the table has the highest probability on the shortlist.
We fitted this threshold on 4{,}826 screening decisions, each of which is a pair of a query and one file on its shortlist, from searches over two evaluation workspaces in which every query is annotated with the files that contain its answer.
On these decisions, this simple thresholding approach correctly retains 99.1\% of the files that contain the gold answer and removes 12\% of the files that do not.

\paragraph{Checking whether retrieved passages already answer the query.}
For each shortlisted file, the retrieval system also returns its \emph{passages}, which are the text chunks of that file that matched the query, and its \emph{graph facts}, which are short statements about entities in the file, such as the due date of an invoice, that the framework extracted from the file when indexing it.
Normally, fact search then fetches the content of each shortlisted file and passes it to the answering model.
Before doing so, the framework asks \drex, separately for each file, whether these passages and graph facts alone contain every required component of the answer, for the entity, period, and document that the query names.
If the probability is above a high threshold for any file, the framework does not fetch the content of any file, and the answering model writes the answer from the passages of the files that passed this check, citing those files.
The threshold is high because the framework does not check such an answer against the files afterwards.
We fitted it on 103 recorded searches with reference answers, for each of which we graded both the answer written from passages and the answer written from the fetched files against the reference.
At this threshold, 37 of the 103 searches are answered from passages, and for none of them the answer from passages was graded worse than the answer from the fetched files.
When we reran these searches, no probability changed by more than 0.005, so the decisions are reproducible.

\paragraph{Fitting the thresholds.}
We fit each threshold in the labeled examples of its decision type described above.
Thresholds are specific to the model: when fitted in the same way, the routing threshold is 0.80 for \drex Max, 0.51 for v2.0, and 0.25 for v1.5. If the threshold for v2.0 were set to be the same as that of Max, the number of correctly routed queries would decrease by 6\%.
In accordance with \cref{sec:calib-results}, a threshold must therefore be changed whenever the model changes.

\section{Inference efficiency by model}
\label{app:servemodel}

This appendix gives the inference measurements behind \cref{sec:inferencetechniques}, in the order in which the models were developed.
Each technique is described once, under the first model that used it.

\subsection{Inference efficiency of \drex v1.5}
\label{sec:servingv15}

Unless stated otherwise, the measurements in this subsection are median latencies of the released \drex v1.5 checkpoint on one NVIDIA B200 GPU, measured in September 2026.
They exclude the time spent in the API gateway, which receives requests and forwards them to the model.

\begin{figure}[t]
\centering
\includegraphics[width=\textwidth]{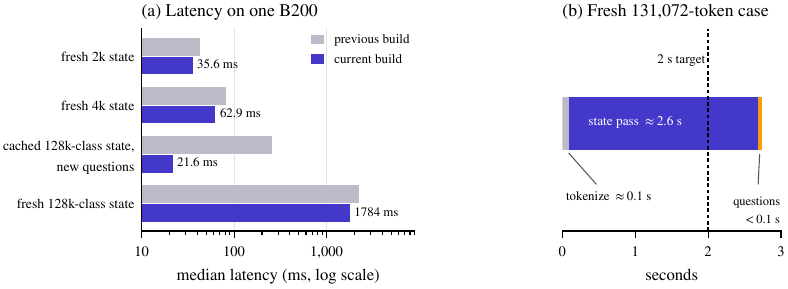}
\caption{\textbf{Inference latency of \drex v1.5.} \textbf{(a)} Median latency of the previous and the current version of our inference code. A new question about a cached state of 114--117k tokens is answered faster than an uncached state of 2k tokens can be processed. \textbf{(b)} For an uncached state at the maximum length of 131{,}072 tokens, almost all of the time is spent in the state pass, which processes the state once per request, and the total exceeds our target of 2\,s.}
\label{fig:serving}
\end{figure}

\paragraph{Execution model.}
\Cref{sec:inferencetechniques} describes the two passes of a request to a pointer model: the \emph{state pass}, which processes the state once and whose cost grows with the length of the state, and the \emph{question pass}, which runs once per question and processes only the tokens of that question.
For \drex v1.5, a new question about a cached state of 114--117k tokens is answered in 21.6\,ms, which is faster than processing an uncached state of 2k tokens (35.6\,ms; \cref{fig:serving}a).
Processing an uncached state remains the main cost: it takes 1.78\,s for a state of 114--117k tokens and about 2.7\,s at the maximum length of 131{,}072 tokens, almost all of it in the state pass (\cref{fig:serving}b).
The latter exceeds our target of 2\,s for this worst case.

\paragraph{Optimizations.}
We run \drex v1.5 on inference code that we wrote for the pointer models, rather than on a general-purpose inference engine such as vLLM~\citep{kwon2023vllm}.
This lets us optimize the state pass and the question pass separately, and compare the outputs after every kernel change with those of the previous code on real questions (\cref{sec:servinggate}).
We batch requests with short states, since the time needed to launch GPU kernels dominates their latency.
We do not batch requests with long states: their computation already saturates the GPU, and batching them doubled the median latency without increasing throughput.
The optimizations are standard techniques, so we list them without their tuning parameters.
Our inference code uses
\begin{itemize}\itemsep0pt
  \item FP8 matrix multiplications in the linear projections, using a custom kernel that is compared bit for bit with the PyTorch matrix multiplication when the inference server starts, and that falls back to the PyTorch kernel for every shape where the two differ.
  \item CUDA graphs, which record a sequence of GPU kernels once and replay it with a single launch, captured for the input shapes that occur in real requests.
  \item Variable-length FlashAttention~\citep{dao2022flashattention}.
  \item An attention kernel in which the sequences of all questions (\cref{sec:attn}) use a single copy of the cached keys and values of the state, instead of one copy per question.
  \item Kernels that take the sequence length as an argument at runtime, instead of being compiled separately for each length. Nearly every request with a long state has a length that has not occurred before, so kernels compiled for each length would have to be compiled while the request waits.
  \item Parallel tokenization of long states, with output identical to sequential tokenization.
  \item Skipping the attention output and the feed-forward computation of the last layer for state tokens, since the pointer head uses only the hidden states at the \tok{decide} and \tok{/opt} tokens.
\end{itemize}
Compared with the previous version of our inference code, these optimizations together reduced the latency of a new question about a cached state of 114--117k tokens from 254\,ms to 21.6\,ms, and the latency of a request with an uncached state of the same length from 2.22\,s to 1.78\,s.
Passing the sequence length at runtime reduced the latency of the first request at a new length from over a second, most of which was spent compiling kernels, to about 160\,ms.
Parallel tokenization reduced the tokenization time of the longest states from 0.3\,s to 0.1\,s.
Skipping the unused computation in the last layer reduced latency by 5\% at the maximum length of 131{,}072 tokens and left all 1{,}200 outputs that we checked unchanged.
When \drex is served on a single H100 GPU, a policy that decides which CUDA graphs to capture for requests of mixed lengths increased throughput over our earlier inference code by 8--49\%, with the gain growing from 1k-token to 8k-token states.
The policy pads the input of a pass only up to shapes that use the same tile of the FP8 matrix-multiplication kernel, and it captures the graphs for short states and for question sequences before the first request arrives.
All of these gains are measured relative to our earlier inference code, not relative to another inference engine.

\subsection{Inference efficiency of \drex v2.0}
\label{sec:servingv2}

Unlike v1.5, Drex v2.0 is served with a general-purpose inference engine, which processes each question as a separate sequence that contains the full state.

\paragraph{Shared prefill.}
Without further changes, a request with many questions would therefore process its state once per question.
Shared prefill processes the prefix that all questions of a request have in common only once, and each question then processes only the rest of its sequence.
We apply it only to requests with more than one question and a state of at least 1{,}024 tokens.
For shorter states, sharing was slower than processing each question separately and changed the output probabilities slightly.
For states of 8k tokens with 16 questions and 8 concurrent clients, shared prefill increased throughput tenfold.
The speedup per request grew with the length of the state, from 6--8 times at 8k tokens to 11--12 times at 32k tokens, and for states of at least 1{,}024 tokens the outputs were bit-identical to those without sharing.

\paragraph{Negative results.}
FlashAttention-4 was slower than the previous attention kernel on short inputs, so v2.0 uses it only for inputs of at least 8{,}192 tokens.
Quantization to FP8 at runtime (online), to MXFP8, or to NVFP4 gave at most a 1.1$\times$ speedup and changed the accuracy, so v2.0 uses none of them.

\subsection{Inference efficiency of \drex Max}
\label{sec:servingmax}

\drex Max cannot share the processed state between questions (\cref{sec:maxcost}), so every question requires processing the full state.
Our inference work on \drex Max therefore focuses on the speed of individual kernels on NVIDIA Blackwell GPUs (B200 and B300).
Every change that alters the output probabilities was checked against the accuracy criterion for \drex Max (\cref{sec:servinggate}).

\paragraph{Recurrent layers.}
The Gated DeltaNet layers of \drex Max originally ran on kernels that we wrote in Triton.
Replacing them with the prefill kernel of FlashInfer for Blackwell GPUs reduced latency by 9--16\% for states of 8k to 128k tokens.
On the 11{,}096 decisions of our accuracy panel (\cref{sec:servinggate}), accuracy changed by $-0.13$ points, and 98.7\% of the top options agreed with those of the same model with the previous kernels, which is within the criterion for \drex Max.
We consider this kernel an interim solution.

\paragraph{MLP quantization.}
Quantizing all 128 MLP projections to NVFP4, instead of 96 of them, reduced the median latency by about 8\%.
The mean accuracy on our panel was essentially unchanged relative to FP8 (87.42 and 87.53).
On the When2Call panel, however, accuracy was 2.00 points higher than with a BF16 reference (95\% interval $+0.60$ to $+3.40$), which exceeds our limit of 1.5 points per dataset (\cref{sec:servinggate}).
We nevertheless released the change after a manual review, and we report the difference here.

\paragraph{Hardware.}
With a single client, \drex Max ran 17--32\% faster on a B300 GPU than on a B200 for states of 4k tokens or more.
With 16 concurrent clients and states of 32k tokens, this advantage fell to 3\%.
On B300, accuracy on When2Call was 1.04 points below the BF16 reference on B200, which is within the limit of 1.5 points per dataset.

\section{Deployment configurations}
\label{app:serve}

\begin{table}[h]
\centering
\caption{Validated inference implementations. $^\ast$Apple silicon requires disabling Metal tensor matmul for the DLM.}
\footnotesize
\begin{tabular}{@{}lllll@{}}
\toprule
Implementation & \drex DLM & \drex v1.5 & \drex v2.0 & \drex Max \\
\midrule
Python reference server & CUDA, Metal$^\ast$ & CUDA & CUDA (\kev) & CUDA \\
llama.cpp fork & \texttt{edlm} branch & \texttt{drex-v1.5} & \texttt{drex-v2.0} & \texttt{drex-max-v1}, text only \\
Ollama fork & \texttt{nace-edlm} branch & \texttt{drex-v1.5} & \texttt{drex-v2.0} & \texttt{drex-max-v1}, text only \\
vLLM pooling + in-process head & --- & --- & supported & --- \\
Hosted API & --- & \texttt{drex-v1.5} & \texttt{drex-v2.0} & \texttt{drex-max} \\
\bottomrule
\end{tabular}
\end{table}

With llama-server, \texttt{-np 2} makes the pointer models encode the state once and share it across questions. The context size \texttt{-c} is divided among the slots. \drex Max reads each question as one prompt, so its batch sizes equal the context.
The open-source packages default to a context limit of 16{,}384 tokens so that they run on ordinary machines (\texttt{KEV\_CONTEXT} / \texttt{SYSTEMONE\_CONTEXT}); the models support up to 131{,}072 tokens for v1.5 and v2.0, 32{,}768 for DLM, and 262{,}144 for \drex Max.
The pointer models can be deployed as a managed API, in the customer's own cloud account, or air-gapped on premises.

\section{Decision Index scoring}
\label{app:di}

For a benchmark $b$ with official metric $m_b$ and chance level $r_b$, the skill is $s_b = 100 \cdot (m_b - r_b) / (1 - r_b)$, clipped to $[0, 100]$.
The chance level is the expected value of the metric under uniform guessing or, for probabilistic tasks, the value of the metric for the baseline stated by the benchmark.
The score of a domain is the weighted mean of the skills of its benchmarks.
On version 0.3, benchmarks that the maintainers designate as \emph{gold} have weight 1.2 and all other benchmarks weight 1.0; we do not reproduce this designation here.
The index is the weighted mean of the domain scores.
On version 0.2.1, the weight of a domain is proportional to $\sqrt{n_{\text{bench}}}$, where $n_{\text{bench}}$ is its number of benchmarks, and the weight of Arts is fixed at 10\%.
On version 0.3, the weights are fixed at 25.8\%, 25.8\%, 20.0\%, 18.3\%, and 10\% for Knowledge, Language, Retrieval, Tools, and Arts; these values are rounded to one decimal and sum to 100\% before rounding.
The \emph{raw} score is the same aggregate computed from the uncorrected metrics $100\, m_b$ instead of the skills.
A request that times out, fails with an error, or returns an output in an unsupported format is scored as wrong.
The fraction of requests that receive a valid answer (\emph{coverage}) is reported alongside the index.

\end{document}